\RequirePackage{fix-cm}
\documentclass[twocolumn]{svjour3}          % twocolumn
\smartqed  % flush right qed marks, e.g. at end of proof

\usepackage{graphicx}
\usepackage{natbib}
\usepackage{filecontents}
\usepackage[colorlinks=true,linkcolor=blue,citecolor=blue]{hyperref}%
\usepackage{mathptmx}
\usepackage{mathrsfs}
\usepackage{mathtools}
\usepackage{amsmath}
\usepackage{amssymb}
\usepackage{amsthm}
\usepackage{algorithm}
\usepackage{algorithmic}
\usepackage{subfigure}
\usepackage{graphicx}
\usepackage{tabularx}
\usepackage{color}
\usepackage{caption}
\usepackage{ragged2e}
\usepackage{epsfig}
\usepackage{algorithm}
\usepackage{multirow}
\usepackage{enumitem}
\usepackage{tensor}
\usepackage{xcolor}
\usepackage{booktabs}
\usepackage{balance}
\usepackage{latexsym}
\usepackage{multirow}
\usepackage{color, colortbl}
\usepackage[normalem]{ulem}
\algsetup{linenosize=\small}
\pdfoutput=1

\theoremstyle{definition}
\definecolor{mygreen}{rgb}{0.000, 0.392, 0.000}
\definecolor{Gray}{gray}{0.9}

\newcommand{\Rone}[1]{\textcolor{black}{\textbf{} #1}}
\newcommand{\Rtwo}[1]{\textcolor{black}{\textbf{} #1}}
\newcommand{\Rthree}[1]{\textcolor{black}{\textbf{} #1}}

\journalname{International Journal of Computer Vision}
\begin{document}

\title{Learning Robust Multi-Scale Representation for Neural Radiance Fields from Unposed Images %\thanks{Grants or other notes
}

\titlerunning{Learning Robust Multi-Scale Representation for Neural Radiance Fields from Unposed Images}        % if too long for running head

\author{Nishant Jain, Suryansh Kumar$\dagger$, Luc Van Gool
}


\institute{Nishant Jain is with IIT Roorkee, India \at
           \email{njain@cs.iitr.ac.in}\\
            Suryansh Kumar is with PVFA Texas A\&M University USA and CVL Lab ETH Z\"urich, Switzerland. \at \email{k.sur46@gmail.com}\\
            Luc Van Gool is with CVL Lab ETH Z\"urich, Switzerland \at \email{vangool@vision.ee.ethz.ch} \\
           $\dagger$ Corresponding Author
           %  \\
%       \emph{Present address:} of F. Author  %  if needed
}

\date{Received: date / Accepted: date}

%, unlike classical Neural Radiance Fields (NeRF) and its similar extensions

\maketitle

\begin{abstract}
We introduce an improved solution to the neural image-based rendering problem in computer vision. Given a set of images taken from a freely moving camera at train time, the proposed approach could synthesize a realistic image of the scene from a novel viewpoint at test time. The key ideas presented in this paper are \textit{(i)} Recovering accurate camera parameters via a robust pipeline from unposed day-to-day images is equally crucial in neural novel view synthesis problem; \textit{(ii)} It is rather more practical to model object's content at different resolutions since dramatic camera motion is highly likely in day-to-day unposed images. To incorporate the key ideas, we leverage the fundamentals of scene rigidity, multi-scale neural scene representation, and single-image depth prediction. Concretely, the proposed approach makes the camera parameters as learnable in a neural fields-based modeling framework. By assuming per view depth prediction is given up to scale, we constrain the relative pose between successive frames. From the relative poses, absolute camera pose estimation is modeled via a graph-neural network-based multiple motion averaging within the multi-scale neural-fields network, leading to a single loss function. Optimizing the introduced loss function provides camera intrinsic, extrinsic, and image rendering from unposed images. We demonstrate, with examples, that for a unified framework to accurately model multiscale neural scene representation from day-to-day acquired unposed multi-view images, it is equally essential to have precise camera-pose estimates within the scene representation framework. Without considering robustness measures in the camera pose estimation pipeline, modeling for multi-scale aliasing artifacts can be counterproductive. We present extensive experiments on several benchmark datasets to demonstrate the suitability of our approach.

\keywords{Neural Radiance Fields \and Motion Averaging \and Multiscale Representation \and Single Image Depth Prediction.}

\end{abstract}

\section{Introduction}\label{sec:intro}
Using neural fields to represent a 3D scene from its multi-view (MV) images has recently become popular for solving novel view synthesis problems. This is primarily due to the \cite{mildenhall2020nerf} work on neural radiance fields popularly known as NeRF. NeRF's idea to scene representation has shown promising results on several computer vision,  graphics, and robotics problems \citep{yu2021pixelnerf,sucar2021imap,zhang2021nerfactor,liu2020neural,martel2021acorn, kaya2022neural, lee2022uncertainty, jain2023enhanced, haghighi2023neural}. Yet, its original design choice has inherent challenges in handling day-to-day  MV images captured from a freely moving camera. For instance, NeRF shows visual artifacts on multiple scale images \citep{barron2021mipnerf}, and its performance degrades even with subtle inaccuracies in camera pose estimates \citep{lin2021barf}. Therefore, to make NeRF and similar approaches more usable for arbitrarily captured MV images, the approach must generalize to more realistic indoor and outdoor scenes with dramatic camera motion.

%\Rone{\sout{captured}}
%must model multiple scale images and estimate correct camera parameters even for dramatic camera motions.

\begin{figure*}[t]
\centering
\subfigure[\label{fig:first_page_teaser}  \textbf{Left}: Qualitative Comparison on Mic dataset]{\includegraphics[width=0.48\linewidth]{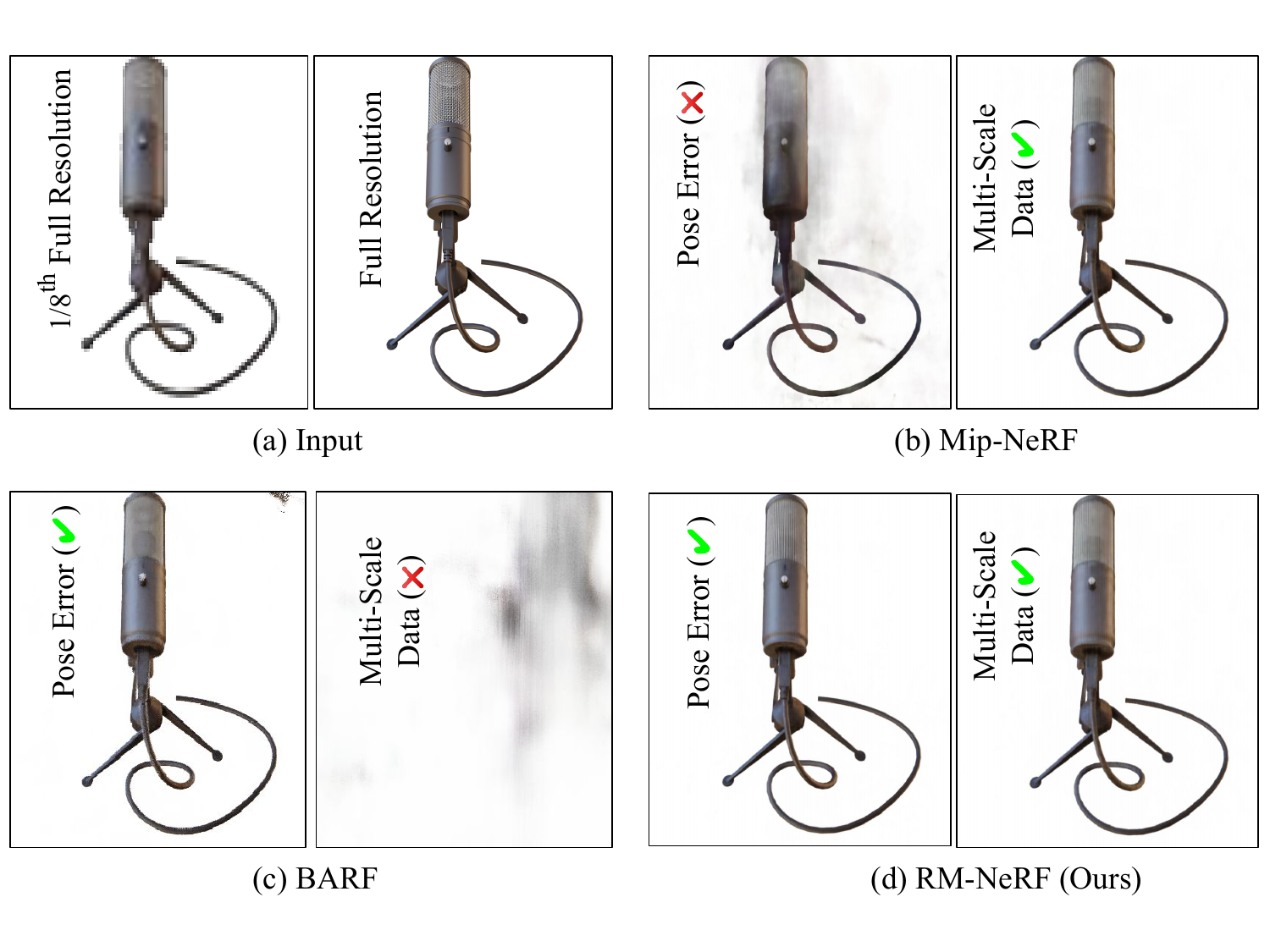}}
~~~\subfigure[\label{fig:mip_cone_illustration}  \textbf{Right}: Intuition on importance of camera pose accuracy. ]{\includegraphics[width=0.48\linewidth]{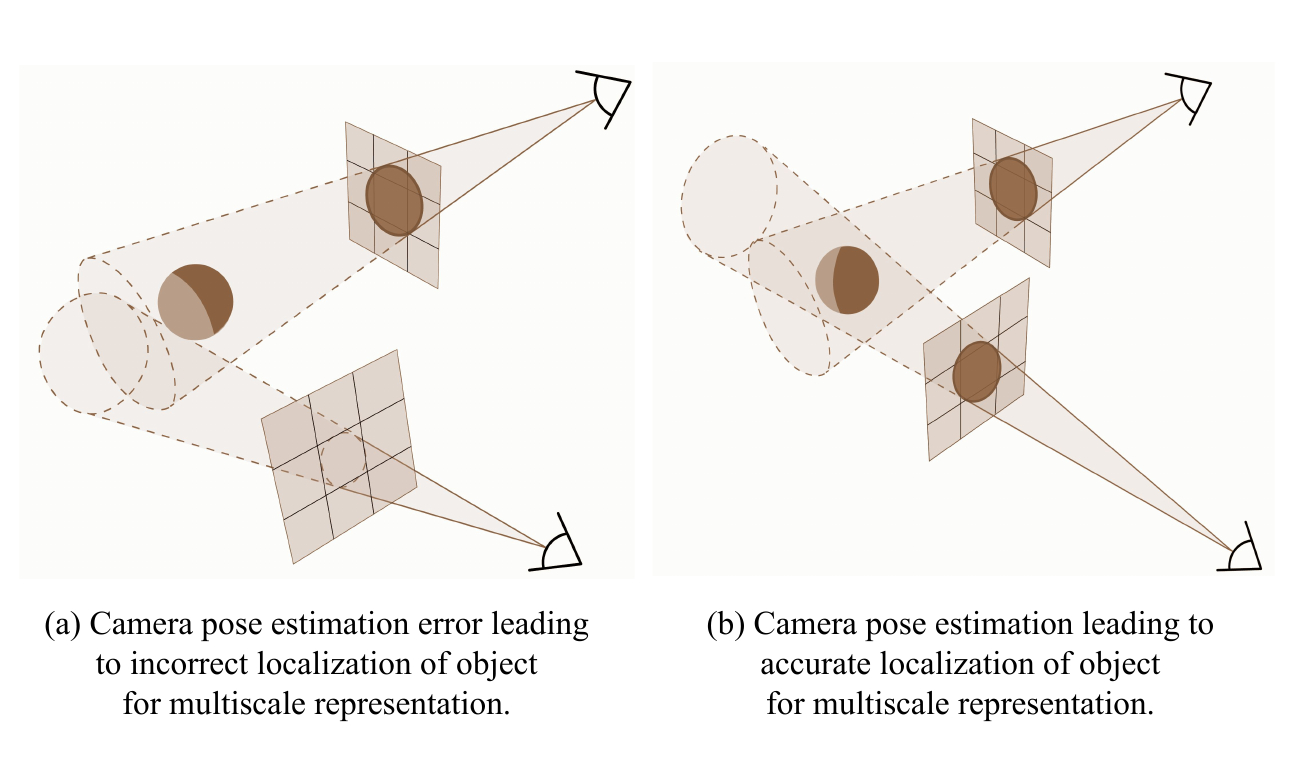}}
\caption{\textbf{Left}: (a) Multi-scaled, multi-view images with camera pose error is fed to related NeRF-based methods \citep{mildenhall2020nerf, barron2021mipnerf, lin2021barf}. (b) \cite{barron2021mipnerf} Mip-NeRF can handle multi-scale imaging effects but fails if the camera poses error persists. (c) \cite{lin2021barf} BARF can handle the camera pose error for same-scale images but fails for multi-scale images. (d) Our approach works well for both cases. \textbf{Right}: (a) Error in the camera pose estimation can lead to incorrect cone casting in the volume space leading to misguided localization of the object for proper modeling. (b) Correct camera poses certify the proper modeling of the object volume for each sampled canonical frustum.}
\label{fig:examples_fig_page_1}
\end{figure*}

While recently proposed Mip-NeRF \citep{barron2021mipnerf} solves the multiscale issues with NeRF, it assumes known camera parameters, i.e., ground-truth camera poses as well as camera intrinsics are given, or estimated via off-the-shelf COLMAP software \citep{schoenberger2016sfm}. On the other hand, recent works such as BARF \citep{lin2021barf}, NeRF--\citep{wang2021nerf}, SC-NeRF \citep{jeong2021self} introduced formulations to simultaneously estimate camera pose yet unsuitable for multiscale unposed images \citep{Jain_2022_BMVC}. Furthermore, available methods in this same vein often ignore the relative camera motion between images, which is a critical prior to absolute camera pose estimation. In both of these independent research directions, a gap exists, i.e., BARF \citep{lin2021barf} and similar methods can jointly solve the camera pose with neural fields representation but cannot address multiscale image issues. On the contrary, Mip-NeRF \citep{barron2021mipnerf} can handle multiscale images but assumes the correct camera pose. Hence, in this work, we introduce a simple and effective approach to fill this gap. By utilizing the fundamentals of scene rigidity, relative camera motion, and scene depth prior, we jointly address the multiscale issues and the challenges in camera parameter estimation within neural fields approaches (see Fig.\ref{fig:first_page_teaser}). Consequently, our self-contained approach performs well for handheld captured MV images.

%\kumar{Here!}
To put the notions intuitively, we show in Fig.\ref{fig:mip_cone_illustration} that the correct intersection of the conical frustum for object localization ---as proposed in Mip-NeRF \citep{barron2021mipnerf}, is possible if both the camera poses are correctly known. One trivial way to solve this is to jointly optimize for object representation and camera pose as done in BARF \citep{lin2021barf} and NeRF--\citep{wang2021nerf} with Mip-NeRF representation idea. As is known that the bundle-adjustment (BA) based joint optimization is complex, sub-optimal, requires good initialization, and can handle only certain types of noise and outlier distribution \citep{chatterjee2017robust}. So, conditioning the multi-scale rendering representation based on BA-type optimization could complicate the approach, hence not an encouraging take on the problem.

To solve the above mentioned challenges, we propose an approach that leverages the fundamentals of scene rigidity and other scene priors that could be estimated from images. Firstly, we estimate the camera motion robustly without having explicit information about the object's 3D position assuming a rigid scene \citep{govindu2001combining}. Secondly, we estimate the geometric prior per frame without using any camera information by relying on single image depth prior \citep{ranftl2021vision}. This helps in overcoming the object's geometry and radiance ambiguity in multi-scale neural radiance fields representation \citep{barron2021mipnerf}. Thirdly, we use relative camera motion prior between the frames with predicted depth to further improve the absolute camera pose solution between frames.

At the heart of the proposed approach lies the idea of disentangling geometry, radiance, and camera parameters in multi-scale neural radiance fields representation for making novel view synthesis more usable and practical. Our approach introduces graph-neural network-based multiple motion averaging with multi-scale feature modeling and per-frame depth prior to solving the problem. For single image depth prior, we rely on \cite{ranftl2021vision} work\footnote{The results can be improved further by using better single image depth prediction model such as \citet{liu2022va}, \citet{liu2023single}, etc.}. In this article, we claim the following contributions.

\smallskip
\noindent
\textbf{Contributions}
\begin{itemize}[leftmargin=*,topsep=0pt, noitemsep]
    \item We propose a novel view synthesis approach to jointly estimate camera parameters and multi-scale scene representation from daily captured multi-view images.  
    
    \item The introduced approach exploits rigid scene assumption to disentangle the camera motion estimation variables from explicit 3D geometry variables. Furthermore, radiance and shape ambiguity is resolved by utilizing the per-image scene depth prior. 
    
    \item The proposed loss function utilizes multi-scale scene representation and per-view scene depth with graph neural network-based multiple motion averaging for robust camera pose parameters estimation leading to improved scene representation.
\end{itemize}

\smallskip
\noindent
\Rone{
%\textbf{Contributions} (in addition to BMVC work)
This article extends our published paper at the British Machine Vision Conference (BMVC), 2022 \citep{Jain_2022_BMVC}. Firstly, the proposed approach extends it to recover accurate scene representation and camera poses starting from entirely random poses. It is referred to as RM-NeRF (w/o pose). Another extension presented is that we further relax the requirement of camera intrinsics as input and recover correct scene representation from randomly initialized camera extrinsic and intrinsic parameters. We refer it as RM-NeRF (E2E). Thus, the proposed approach serves as a unified framework, eliminating the dependency on other third-party modules.
%\begin{itemize}
    %\item We build on top of our previous work and extend it to recover accurate scene reconstruction and camera poses starting from completely random poses (RM-NeRF (w/o pose)).
    %\item We propose another extension which further relaxes the requirement of camera intrinsics and can recover correct scene reconstruction from random extrinsics/intrinsics (RM-NeRF (E2E)). Thus, this can serve as an end to end framework and can eliminate the dependency completely on any third party software. 
    %\item Experimental results show that our newly proposed algorithms, initialized with random intrinsics/extrinsics, are quite effective in presence of high noise in the poses on the multi-scale Blender dataset, whereas RM-NeRF struggles, and are also better than RM-NeRF+COLMAP initialization for a randomly captured scene (collected using our phone by randomly walking around an object). Furthermore, when initialized with COLMAP poses on the tanks and temples dataset, they outperform the RM-NeRF method and are also able to recover quite accurate scene structure when initialized with random poses. 
    %They also outperform RM-NeRF on the challenging in-the-wild scenes of the recently proposed NAVI Dataset where COLMAP struggles to find decent enough initialization. We also analyze their performance along with RM-NeRF on the ScanNet dataset and also for a scene comprising a specular object.
%\end{itemize}
}

{Experimental results show that our proposed extensions initialized with random intrinsic and extrinsic camera parameters, are quite effective. To test this, we presented the RM-NeRF \citep{Jain_2022_BMVC} with very noisy pose initialization in one of the experiment and with COLMAP poses in the next experiment. We observed that the introduced extensions are quite effective and provides commendable results compared to the baseline experiments on day-to-day captured images, which we collected using our phone by randomly walking around an object. Our approach achieves better camera pose estimates and novel view synthesis results than the existing NeRF-based baseline methods when tested on the standard benchmark dataset \citep{mildenhall2020nerf, Knapitsch2017}. Additionally, our approach outperforms RM-NeRF \citep{Jain_2022_BMVC} on tanks and temples dataset \citep{Knapitsch2017} as well on recently proposed NAVI dataset \citep{jampani2023navi} under similar experimental settings. Refer to Sec. \S \ref{sec:result} for more details.
}

%\smallskip

% for modeling neural scene representation.
    % fundamentals to jointly optimize camera pose and rendering loss function. To this end, our approach leverage multi-scale representation \cite{barron2021mipnerf} and introduces graph neural network-based multiple motion averaging to learn the noisy camera motion estimates from the images. As a result, our approach helps in better estimation of the network's model parameters for the multi-scale scene or object representation.
    % \item The proposed method achieves better camera pose estimates and novel view rendering results than the existing NeRF-based baseline approaches when tested on the standard benchmark sequence \cite{mildenhall2020nerf} and other popular real-world sequences \cite{Knapitsch2017}.
% \kumar{Write in a very raw manner what extra we are doing?
% 0. BMVC paper.

% 1. Fully uncalibrated method. $Intrinsic -> NeRF -- Extrinsic -> NoPE NeRF.$

% 2. end-to-end framework. (SIDP work without K, R, t)

% 3. For both pose and novel view rendering.
% }

%\kumar{Here!}

\section{Related Works}
Recently, neural radiance fields (NeRF) based implicit scene representation has gained significant attention in the computer vision and graphics community with several extensions. As a result, discussing all the NeRF-related methods is beyond the scope of the article, and interested readers may refer to  \cite{tewari2021advances} paper for reference. Here, we keep the related work discussion concise and concern ourselves with methods relevant to our proposed approach.

\subsection{Neural Fields for Scene Representation}
NeRF \citep{mildenhall2020nerf} represents a rigid scene as a continuous volumetric field parametrized by a multi-layer perceptron (MLP). It assumes a fully calibrated setting with well-posed input images, i.e., correct camera pose and internal camera calibration matrix is known, and images are captured in a dome setting. Once the experimental setup is prepared, NeRF for each pixel sample points along rays that are traced from the camera's center of projection. Later, these sampled points are transformed using positional encoding to represent each point in a high-dimensional feature vector before being fed to an MLP for density and color estimation for novel view synthesis at test time.

\smallskip
\noindent
\textbf{\textit{(i)} Multiscale NeRF.} \cite{barron2021mipnerf} introduced Mip-NeRF to overcome the limitations with NeRF in rendering multi-resolution images, i.e., MV images are captured at a varying distance from the object. Instead of sampling points along the rays traced from the camera center of projection, Mip-NeRF queries samples along a conical frustum interval region approximated using 3D Gaussian to render the corresponding pixel. Since the image acquisition setup used in NeRF is unrealistic for many practical day-to-day captured videos, Mip-NeRF broadens the scope of NeRF formulation to commonly acquired multi-view and multi-scale image acquisition setups. Yet, the Mip-NeRF assumption on the availability of ground-truth camera pose parameters is uncommon and could substantially restrict its application.

\smallskip
\noindent
\textbf{\textit{(ii)} Uncalibrated NeRF.} Recently, a few methods have appeared to jointly solve camera pose and object's neural representation via NeRF formulation. For example, BARF \citep{lin2021barf} leverages photometric bundle adjustment to estimate the camera poses and recover scene representation jointly. Recently, NeRF$--$\citep{wang2021nerf} introduced an approach for estimating intrinsic and extrinsic camera calibration while training the NeRF model. Nonetheless, these extensions of NeRF work well for the same scale images; accordingly, its usage is limited to a synthetic multi-view dome or hemispherical setup. Not long ago, NoPe-NeRF \citep{bian2022nope} utilized depth maps to estimate camera poses via point cloud alignment and a surface-based photometric loss. As a result, it can reconstruct the scene from randomly initialized poses. Other related work includes iNeRF \citep{yen2020inerf} that solves camera poses given a well-trained NeRF model, and SC-NeRF \citep{jeong2021self}. The method jointly learns the camera parameters and scene representation using a loss function that enforces geometric consistency for a given camera model.

\smallskip
\noindent
\textbf{\textit{(iii)} NeRF extension with other scene priors.}  There have been several attempts to make the NeRF approach either faster or more generalizable by extracting valuable features from the input images. One line of works \citep{yu2021pixelnerf,wang2021ibrnet} involves estimating a feature volume from an image via a generalizable CNN and then feeding the feature vector into the MLP to generalize NeRF idea. Another line of works \citep{chen2021mvsnerf, xu2022point} estimates scene 3D structure prior via MVS-based methods \citep{yao2018mvsnet}. Combining input image features with recovered 3D structure, it learns better scene representation, and such approaches are shown to converge faster.

%We also follow this second line of work and use a feature based point cloud as a prior, following a recent work \cite{xu2022point}, along with the MLP to learn multi-scale representation much faster and render novel views.
% works which first estimate prior 3D structure and features and then use volumetric rendering equation proposed in NeRF to synthesize novel views. Based on these lines, a recent work, PointNeRF\cite{pointnerf},  proposed using the MVS-based networks to first recover a dense point cloud for the scene and then to render a pixel along a given direction, it first estimates the surface point along that direction and $K$ of its nearest neighbours from the estimated structure to predict the color for the pixel by aggregating the color of these neighbours using their volumetric density.

\subsection{Camera Pose Estimation}
Widely used approaches to camera pose estimation from multi-view images are based on image key-points matching and incrementally solve camera pose \citep{agarwal2011building} or use global BA \citep{Triggs:1999:BAM:646271.685629} with five-point \citep{nister2004efficient} or eight-point algorithm \citep{hartley1997defense} \Rone{running at \sout{its} the back-end}. Yet, such methods can provide sub-optimal solutions and may not robustly handle outliers inherent to the unstructured images. To address such an intrinsic challenge with pose estimation, Govindu \citep{govindu2001combining} initiated and later authored a series of robust multiple rotation averaging (MRA) approaches \citep{govindu2016motion, chatterjee2017robust}. The benefit of using MRA is that it uses multiple estimates of noisy relative motion to recover absolute camera pose based on view-graph representation and rotation group structure \citep{govindu2006robustness} \textit{i.e.}, $SO(3)$. Contrary to the robust conventional rotation averaging approaches \citep{chatterjee2017robust, aftab2014generalized, hartley2011l1}, in this work, we adhere to graph neural network-based approaches for robust camera pose estimation via a learned view-graph module, which helps in better removal of the erroneous poses nodes in the graph \citep{Yang_2021_CVPR, gilmer2017neural, purkait2020neurora,li2021pogo}.

Note that part of our work was published as a conference proceeding at the British Machine Vision Conference (BMVC), 2022 \citep{Jain_2022_BMVC}. Nevertheless, this journal version is a substantial extension of the conference paper both in terms of formulation, experimentation and ablation.

\begin{figure*}
    \centering
    \includegraphics[width=0.96\linewidth]{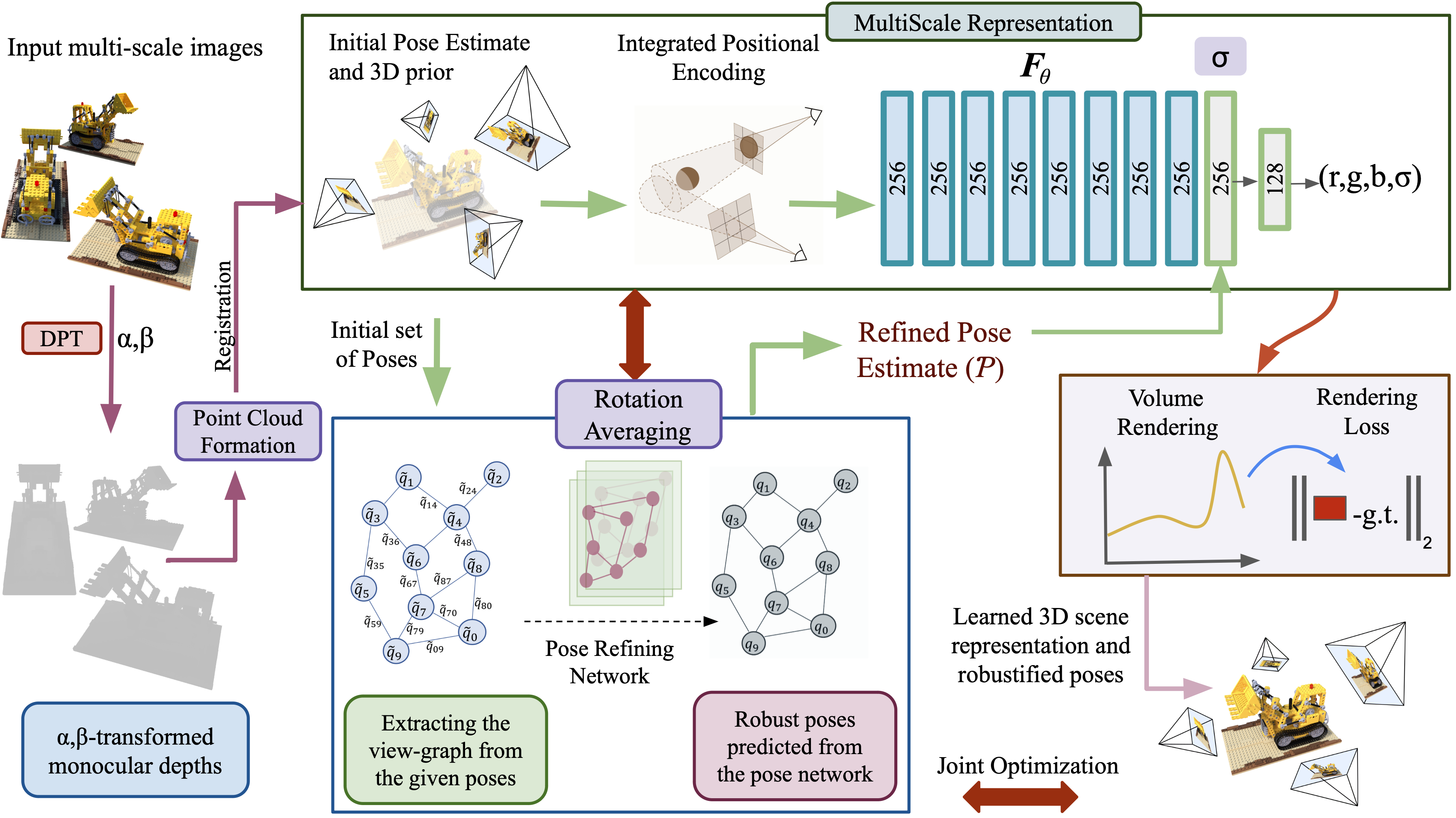}
    \caption{We propose to jointly solve camera poses and learn the multi-scale scene representation. The pipeline consists of a camera pose refining network, single image depth prior to estimate camera pose and the IPE (Integrated Positional Encoding) by casting well-posed conical frustums through the pixels. Later, those are fed to the MLP network for learning the scene representation for novel view synthesis. $\mathcal{P}$ denotes set of camera pose. \Rtwo{Here, DPT symbolizes single image depth prediction model from \cite{ranftl2021vision} work.}}
    \label{fig:pipeline}
\end{figure*}

\section{Problem Statement and Our Approach}\label{sec:mipnerf}
%\kumar{Here!}

Given a set of multi-view images captured from a freely moving handheld camera, the goal is to recover accurate camera pose and learn a better neural scene representation for novel view synthesis. In our problem setting, we predict single image depth prediction (SIDP) prior per frame using off-the-shelf pre-trained model \citep{ranftl2021vision}.

As discussed, a freely moving camera could lead to scene observation at different pixel resolutions, and therefore, we propose to utilize the mipmapping approach to model the scene representation \citep{barron2021mipnerf}. For joint optimization of camera pose with the scene representation parameters, we compose our proposed pipeline with graph-neural network-based robust motion averaging, where the initial pose could be initialized \emph{randomly or via off-the-shelf algorithms}. Still, we are plagued by radiance-geometry ambiguity, so we introduce SIDP per frame to resolve it. Another advantage SIDP brings is that we can use relative camera pose prior per frame to further improve the camera motion estimates and respective scene 3D parameters.

Based on the above discussion, we propose three algorithmic variations of our proposed idea, which is based on the following variations in the experimental initial setup \textit{\textbf{(i)}} The basic version, dubbed as \textbf{RM-NeRF} takes a noisy set of poses estimated using COLMAP \citep{schoenberger2016mvs} with multi-scale MV images as input. It assumes no depth prior per frame while the intrinsic camera matrix is known.
\textit{\textbf{(ii)}} Similar to the first version, we assume the intrinsic camera matrix as well as per view depth prior is known; however, the camera pose is randomly initialized. We call this version of our algorithm as \textbf{RM-NeRF (w/o pose)}. \textit{\textbf{(iii)}} Assuming SIDP per view and randomly initialized camera pose, the third variation \textbf{RM-NeRF (E2E)} estimates camera intrinsic, camera extrinsic, and scene representation from multi-view images. Fig.(\ref{fig:pipeline}) provides the complete pipeline of our algorithm. Depending on our assumptions about the known priors, we utilize the different modules shown in the diagram to optimize the proposed overall loss function.

Next, we discuss the technical details pertaining to our approach pipeline. We begin with a discussion on multiscale representation for NeRF followed by multiple motion averaging. \Rone{ These two concepts form} the basis of our methodology i.e., \textbf{RM-NeRF}.

%\sout{This two concept forms}
\smallskip
\noindent
\textbf{\textit{(a)} Multiscale Representation for NeRF.} 
By leveraging pre-filtering techniques in rendering \citep{amanatides1984ray} \textit{i.e.}, tracing a cone instead of ray, Mip-NeRF \citep{barron2021mipnerf} learns the scene representation by training a single neural network, which can be queried at arbitrary scales. Furthermore, contrary to NeRF, which uses point-based sampling along each pixel ray to form their positional encoding (PE) feature vector, Mip-NeRF uses the volume of each conical frustum along the cone to model the integrated positional encoding (IPE) features. The positional encoding $\gamma(\mathbf{x})$ (as defined in NeRF \citep{mildenhall2020nerf}) of all the point within the conical frustum, \Rthree{having center at $\textbf{o}$ and axis in the direction $\textbf{d}$}, is formulated as
\begin{equation}\label{eqn:ipe}
\begin{aligned}
~~~~\gamma^{*}(\textbf{o},\textbf{d},\textit{$\dot{r}$},\textit{$t_{0}$},\textit{$t_{1}$}) = \frac{\int_{}{}\gamma(\textbf{x})\textbf{F}(\textbf{x},\textbf{o},\textbf{d},\textit{$\dot{r}$},\textit{$t_{0}$},\textit{$t_{1}$})\textit{d}\textbf{x}}{\int_{}{}\textbf{F}(\textbf{x},\textbf{o},\textbf{d},\textit{$\dot{r}$},\textit{$t_{0}$},\textit{$t_{1}$})\textit{d}\textbf{x}},
\end{aligned}
\end{equation}
where, $\mathbf{F}$ is an indicator function regarding whether a point lies inside the frustum in the given range [\textit{$t_{0}$}, \textit{$t_{1}$}] \Rthree{and $\dot{r}$ is the ray corresponding to the axis}. Since Eq.\eqref{eqn:ipe} is computationally intractable with no closed form solution, it is approximated using multivariate Gaussian which provides ``integrated positional encoding'' (IPE) feature, proposed in \cite{barron2021mipnerf}\footnote{For more details and derivations, refer \cite{barron2021mipnerf} work.}.

% Now, casting a cone from a pixel to the scene space is dependent on the camera pose. 
% % From Eq.\eqref{eqn:volume} it is easy to deduce that an incorrect cone back-projection can lead to misrepresentation of the object's feature and can affect the overall results. 
% Hence, it is important to have a reliable camera pose for multi-scale NeRF representation. Before, we delve into our graph neural network based robust motion averaging for multi-scale representation, we review the concept of rigidity in multiple view geometry and its relation to multiple motion averaging.

\smallskip
\noindent
\textbf{\textit{(b)} Scene Rigidity and Multiple Motion Averaging.} 
Assume a pin-hole camera model with intrinsic calibration matrix $\mathbf{K} \in \mathbb{R}^{3 \times 3}$ and extrinsic calibration $\mathbf{R} \in SO(3), \mathbf{t} \in \mathbb{R}^{3 \times 1}$ as the rotation matrix and translation vector, respectively w.r.t assumed reference.  We can relate $i^{th}$ image pixel $x = [u_i, v_i, 1]^{T}$ to its corresponding 3D point $\mathbf{x} = [{x}_i, {y}_i, {z}_i]^{T}$ using the following popular projective geometry relation, \textit{i.e.},
\begin{equation}\label{eq:img_relation}
~~~~\Rone~~{s}[u_i, v_i, 1]^{T} = \mathbf{K} \big[\mathbf{R} ~|~\mathbf{t}\big] \big[{x}_i, {y}_i, {z}_i, 1\big]^{T}
\end{equation}
\Rone{where $s$ is the constant scale factor.}
Eq.\eqref{eq:img_relation} indicate a non-linear interaction between 3D scene point and camera motion. 
% Another way to think about it is that the 3D scene point and camera motion are entangled in observed image projection. 
Yet, the classical epipolar geometry model suggests that if the scene is rigid $x'^{T}\mathbf{E}~x = 0$ must hold \citep{hartleymultiple}, where $x'$ is the image correspondence of $x$ in the next image frame. It is well-studied that $\mathbf{E}$ can be decomposed into $\mathbf{R}~\text{and}~\mathbf{t}$ such that $\mathbf{E} = [\mathbf{t}]_{\times} \mathbf{R}$, where $\mathbf{E} \in \mathbb{R}^{3 \times 3}$ is the essential matrix and $[\mathbf{t}]_{\times} \in \mathbb{R}^{3 \times 3}$ is the skew-symmetric matrix representation of the translation vector. Using this epipolar relation, we can estimate rigid camera motion without making use of any actual 3D observation. Nonetheless, rigid motion solution based on epipolar algebraic relation is not robust to outliers and may provide unreliable results with more multi-view images \citep{chatterjee2017robust}. So to estimate robust camera motion independent of 3D scene point in a computationally efficient way led to the success of robust motion averaging approaches in geometric computer vision \citep{govindu2006robustness, aftab2014generalized, chatterjee2017robust}. 
% As a result, we now can solve for robust camera motion independent of the actual 3D scene point. 
Moreover, given rotations, solving translations generally becomes a linear problem \citep{chatterjee2017robust}. Consequently, solution to motion averaging reduces to rotation averaging problem. 
% Once average rotation is computed, translation is estimated \cite{chatterjee2017robust}. 
% Given several relative rotation $R_{ij} \in SO(3)$, multiple rotation averaging (MRA) aim to recover absolute rotations $R_i \in SO(3)$ using the compatibility criteria $R_{ij}R_{i} = R_{j}$ \cite{hartley2013rotation}. The multiple rotation averaging (MRA) problem is generally formulated in terms of graph representation. In this paper, we propose to use graph neural network to solve camera pose, which is excellent at dealing with outliers and learns noisy patterns from data.

\subsection{\textbf{RM-NeRF}: Formulation and Optimization}
Let $\mathcal{I}$ be the set of multi-view images taken at different distances from the object (see top left: Fig.\ref{fig:pipeline}). {RM-NeRF} aims at simultaneously updating the MLP parameterized multi-scale representation network ($\theta$) and set of camera poses $\mathcal{P}$, given estimated noisy poses $\Tilde{\mathcal{P}}$ and camera intrinsics $\mathbf{K}$. Assuming the favorable distribution model $\Phi()$, we can write the overall goal of RM-NeRF as
\begin{equation}\label{eq:prob}
\theta, \mathcal{P}\sim \Phi({\theta},\mathcal{P} | \mathcal{I},\Tilde{\mathcal{P}},\mathbf{K}).
\end{equation}

The above formulation can further be simplified based on rigid scene assumption. As a result, we can optimize for the camera pose without explicit knowledge of 3d points in the scene space. Accordingly, we simplify the Eq.\eqref{eq:prob} as
\begin{equation}
    \label{eq:prob_independence}
\Phi({\theta},\mathcal{P} |\mathcal{I},\Tilde{\mathcal{P}},\mathbf{K}) = \overbrace{\Phi(\theta |\mathcal{I}, \mathcal{P},\mathbf{K})}^{\textrm{Multiscale MLP}} ~\cdot \overbrace{\Phi(\mathcal{P} | \mathcal{I}, \Tilde{\mathcal{P}},\mathbf{K}).}^{\textrm{Motion averaging}}
\end{equation}
% However, another possiblity can arise if the scene is rigid, we can refine the camera pose  \emph{independent} of 3d object point in the scene space. Accordingly, we can simplify the Eq.\eqref{eq:prob} as follows:
% \begin{equation}
%     \label{eq:prob_independence}
%     \Phi({\theta},\mathcal{P}|\mathcal{I},\Tilde{\mathcal{P}}) = \overbrace{\Phi(\theta|\mathcal{I}, \mathcal{P})}^{\textrm{Multiscale MLP}} ~\cdot \overbrace{\Phi(\mathcal{P}|\mathcal{I}, \Tilde{\mathcal{P}})}^{\textrm{Motion averaging}}
% \end{equation}
% %
% %
% Using the above relation, we refine the camera pose and optimize for multi-scale neural radiance fields rendering and featurization alternatively for solving the problem
% \subsubsection{(a) Graph Representation for MRA}
%\vspace{-9.0mm}
% \noindent
%
Eq.\eqref{eq:prob_independence} allows a separate modeling scheme for camera pose recovery and the 3D scene representation. Next, we describe our motion averaging approach for camera motion estimation, followed by its modification to recover robust camera pose estimates leading to RM-NeRF joint optimization.

\subsubsection{Graph Neural Networks for MRA.}\label{sec:rotavg}
Assume a directed view-graph $\mathcal{G} = (\mathcal{V, E})$ (see Fig.\ref{fig:pipeline} center bottom). A vertex $\mathcal{V}_j \in \mathcal{V}$ in this view graph corresponds to $j^{th}$ camera absolute rotation ${R}_{j}$ and $\mathcal{E}_{ij} \in \mathcal{E}$ corresponds to the relative orientation $\Tilde{R}_{ij}$ between view $i ~\text{and} ~j$ (in Fig.\ref{fig:pipeline} represented as quaternions). Here, we assume noisy relative camera motion for view graph initialization. We aim to recover accurate absolute pose ${R}_{j}$ and jointly model the object representation. Conventionally, in the presence of noise, the camera motion is obtained by solving the following optimization problem to satisfy well-known compatibility criteria for rotation group \citep{hartley2013rotation}, i.e.,
\begin{equation}\label{eq:rotation_avg_classic}
\underset{\{{R}_{j}\} }{\text{argmin}} \sum_{\mathcal{E}_{ij} \in \mathcal{E}} \rho\Big( d(\Tilde{R}_{ij}, {R}_j {R}_i^{-1}) \Big),
\end{equation}
where, $d(.)$ denotes a suitable metric on $SO(3)$ and $\rho(.)$ is the robust loss function defined over that metric.
%  \subsubsection{(b) Graph Neural Networks for MRA}
 Minimizing this cost function $\rho(.)$ in Eq.\eqref{eq:rotation_avg_classic} using conventional method may not be apt for several types of noise distribution observed in the real-world multi-view images. Therefore, we adhere to learn the noise distribution from the input data at train time and infer the noisy pattern to robustly predict absolute rotation. We pre-train graph neural network in a supervised setting to learn the mapping $f$ that takes noisy relative rotation $\Tilde{R}_{ij}$ and predict absolute rotations \textit{i.e.}, $\{R_j^{f}\}:=f(\Tilde{R_{ij}}; \Theta)$, where $\Theta$ is the network parameters. 

\smallskip
\smallskip
We now discuss working of our camera pose network performing multiple rotation averaging (MRA) based on message passing graph neural networks (GNNs). We first discuss the working of Message Passing Networks(MPNN) involving a graph node and its neighbours. 
%Each graph node passes the information describing its current state to all of the adjacent nodes. Later, we describe our pose robustifying pipeline in detail by leveraging these MPNNs.

% We now discuss the message passing GNNs, mentioned in the paper for doing MRA, in detail. Firstly, we formulate the message passing setting and then we describe our complete pipeline for robustifying the rotations. 

\begin{figure}
     \centering
     \includegraphics[scale=0.2]{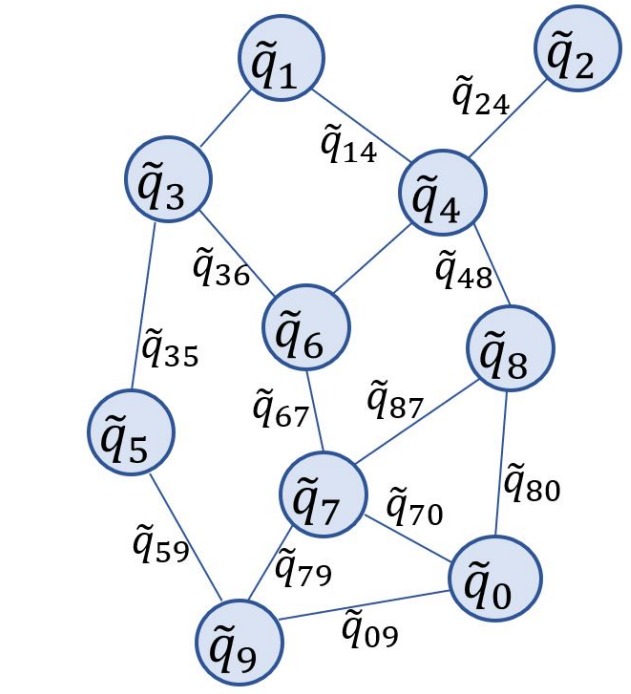}
     \caption{\small  An example view graph extracted form the input poses. The vertex set $\mathcal{V}_{j}$ of this viewgraph correspond to initial absolute orientation of each of the images and the edge set $\mathcal{E}_{ij}$ correspond to the relative orientations between the image pairs.  }
     \label{fig:exampleviewgraph}
 \end{figure}

\smallskip
\noindent
\textbf{\textit{(i)} Message Passing Scheme.} Given a directed view-graph $\mathcal{G}$ (see Fig. \ref{fig:exampleviewgraph}) with $N$ cameras and $M$ pairwise relative orientation, we use the message passing neural network approach to operate on it. Let $m_{j}^{(t)}$ be the message functions that correspond to the message from the neighboring nodes $u\in\mathcal{N}_{j}$. Denoting $\psi^{(t)}$ as the update functions ($T$ layers), and $h_{j}^{(t-1)}$ the state of node $j$ at time step $(t-1)$, the feature node state $h_{j}^{(t)}$ at time $t$ in the graph is updated as:
 \begin{equation}
h_{j}^{(t)} = \psi^{(t)}\big(h_{j}^{(t-1)}, m_{j}^{(t)}\big)
 \end{equation}
 $\psi^{(t)}$ corresponds to concatenation operation followed by a 1D convolution and ReLUs.
 \Rtwo{Intuitively, the node $j$ state at time $t$ is updated via update function $\psi^{(t)}$ based on the current message function value $m_{j}^{(t)}$ and state of the node $j$ at $t-1$ (concatenation). Yet, we want to have a smooth update of the graph node value hence 1D convolution.}
 The message function $m_{j}^{(t)}$ at node $j$ due to all neighbor $\mathcal{N}_{j}$ is expressed as
\begin{equation}
m_{j}^{(t)} = \Omega_{\mathcal{V}_i \in \mathcal{N}_j} h_{i\xrightarrow{}j}^{(t)} %\Psi^{(t)}(h_{v}^{(t-1)},h_{u}^{(t-1)}, e_{uv}).
\end{equation}
 Here, $\Omega(.)$ denotes a differentiable function like the softmax activation function, $h_{i\xrightarrow{}j}^{(t)}:= \Psi^{(t)}(h_{j}^{(t-1)},h_{i}^{(t-1)}, e_{ij})$ is the accumulated message for the edge $\mathcal{E}_{ij}$ at $t$. $\Psi^{(t)}$ is concatenation operations followed by 1D convolution and ReLU activation. In our setup, $\mathcal{N}_j$ is the set of all neighboring cameras connected to $\mathcal{V}_j$ and $e_{ij}$ is the edge feature of the edge $\mathcal{E}_{ij}$. For more details on the messaging passing algorithmic details refer  \cite{gilmer2017neural, purkait2020neurora}
 
%  \kumar{Put some figure corresponding to the view-graph if possible, similar to \cite{purkait2020neurora}}.
 
 %We now discuss the details regarding our GNN-based MRA scheme.
%
%\kumar{Here!}

\smallskip
\smallskip
\noindent
\textbf{\textit{(ii)} Robustifying Poses using GNN.} The GNNs pipeline for estimating robust pose consists of three major steps: 

\smallskip
\noindent
\textit{(1) Cleaning the view-graph.} We first estimate the relative rotations $\Tilde{R}_{ij}$ from the noisy rotations $R_{i}$ due to input data. Next, we apply cycle consistency check to remove the outliers \citep{aftab2014generalized, hartley2013rotation}. Local cyclic graph structure of the view-graph must provide orientation close to identity. Violation of such a local constraint helps in removal of bad camera pose estimates. \Rtwo{In this work, we have proposed three approaches, .i.e, RM-NeRF, RM-NeRF (w/o) pose and RM-NeRF (E2E). For RM-NeRF, we initialized the camera poses in the view-graph using COLMAP \citep{schoenberger2016mvs}. For the rest of our approaches, we initialized the view-graph with random camera poses.}

%
%
% \textit{(a)} Extraction of the relative rotations $\Tilde{R}_{ij}$ from the noisy rotations $R_{i}$, available in the input data, and then applying the cyclic consistency check\cite{aftab2014generalized, hartley2013rotation} to remove the outliers. 

\smallskip
\noindent
\textit{(2) Computing noisy initial solution using the extracted relative rotations}. For this, we build a minimum spanning tree (MST) using all the nodes in the view-graph by fixing the root node to be the node with maximum neighbours (greatest fan-out). Then, we generate an initial absolute rotation $\hat{R_{i}}$ for each node by solving for the motion variables from the root pose value to other nodes along the tree structure.

\smallskip
\noindent
\textit{(3) Refining the initial solution using Graph Neural Networks}. For applying GNNs, we require features corresponding to each node in a view graph. We use the rotation matrix in the initial solution corresponding to every node in the graph as its input feature. Furthermore, we also pass the observed relative rotations $\Tilde{R}_{ij}$ as edge features to the GNN following the formulation described in the previous paragraph. Furthermore, instead of directly passing these relative rotations as edge features, we instead pass the discrepancy between these observed relative rotations and the initial solution resulting in the the edge feature $e_{uv}=\hat{R_{v}}^{-1}\Tilde{R}_{uv}\hat{R}_{u}$, to the GNN.

The resultant input view graph then becomes
$\mathcal{G}$ = $\{\hat{R_{i}}, e_{ij}\}$ leading to a supervised learning problem 
$R_{j}^{f} := f(\{\hat{R}_{i},e_{ij}\};\Theta)$, which is trained using the rotation averaging loss function. Moreover, we know that relative rotation between any 2 nodes in the viewgraph is invariant to any constant angular deviation in form of rotation matrix $R$ to both the nodes and thus, both the solution sets \{$R_{i}R, R_{j}R$\} and \{$R_{i}, R_{j}$\} result in the same discrepancy when using the rotation averaging loss function. To handle this issue involving an unknown global ambiguity in the rotations, we add a regularizer in our objective function to handle such a discrepancy between the absolute rotations. This results in the following objective function 
\begin{equation}
\label{eq:loss_mra}
\begin{aligned}
~~~~\mathcal{L} & =  \sum_{\mathcal{G} \in \mathcal{D}} \sum_{\mathcal{E}_{ij} \in \mathcal{E}}  d_{Q}({q}_{ij}^{f}, q_{ij})  + \beta \sum_{\mathcal{V}_{j} \in \mathcal{V}} d_{Q}(q_{j}^{f}, q_j),
\end{aligned}
\end{equation}
%
%
% Given the formulation and pipeline of our camera pose estimation approach, we now analyse its performance separately from the complete reconstruction pipeline, to demonstrate its effectiveness in resolving errors in the input rotations.
%
%
\Rtwo{where, $\textit{d}\textsubscript{Q} = \text{min}\{\|{p} - {q}\|_2, \|{p} + {q}\|_2\}$ measures distance between two quaternion ${p}, {q}$}. Thus, our overall \textbf{RM-NeRF} loss solves for accurate camera poses and scene representation jointly. Concretely, we combine Eq.\eqref{eq:loss_mra} ($\mathcal{
L}_{mra}$) with the squared error between the true ${C(\mathbf{r})}$ and predicted $\hat{C}(\mathbf{r}) $ pixel colors ($\mathcal{L}_{rgb}$) to define the overall \textbf{RM-NeRF} loss $\mathcal{L}$ as
\begin{equation}
\label{eq:loss_combined}
\begin{aligned}
   \overbrace{\sum_{\mathcal{E}_{ij} \in \mathcal{E}}  d_{Q}({q}_{ij}^{f}, q_{ij}) + \beta \sum_{\mathcal{V}_{j} \in \mathcal{V}} d_{Q}(q_{j}^{f}, q_j)}^{\mathcal{L}_{mra}} +  \overbrace{\sum_{\mathbf{r} \in \mathcal{R}}\|{C(\mathbf{r})} - \hat{C}(\mathbf{r})\|_2^2}^{\mathcal{L}_{rgb}},
\end{aligned}
\end{equation}
Here, $\beta$ is a scalar constant. $q_{ij}$'$\text{s}$ symbolizes corresponding quaternion representation of the rotation matrix defined in Eq.\eqref{eq:loss_mra}. $\mathcal{V}$ denotes the vertex set of the view graph corresponding to the scene being optimized and  
%$\mathcal{D}$ is the entire dataset of view-graphs.
$\mathcal{E}$ denotes the corresponding edge set.

\subsubsection{RM-NeRF Joint Optimization}
\label{sec:biased_opt}
% Eq.\eqref{eq:loss_combined} encapsulates all the variables that we aim to optimize for our approach. However, directly optimizing it for both pose and rendering variables using the gradient descent algorithm will led to inferior minima due to highly non-linear dependence between the pose network parameters and the rendering variable.
%Optimizing eq. \ref{eq:loss_combined} directly using the gradient descent algorithm will establish a dependence between the pose network parameters and the scene structure. This dependence will be caused by the gradient of the rendering cost function w.r.t. pose network parameters.
% Denoting the MLP parameters in rendering network as $\theta$ and camera pose network parameters as $\Theta$, 

% the complete optimization objective is to search for parameters $\theta$ and $\Theta$ jointly such that loss $\mathcal{L}$ defined in Eq.(\ref{eq:loss_combined}) is minimized.
% \begin{equation}
%     (\theta, \Theta) = \arg \min_{\theta, \Theta} \left ( \sum_{\mathbf{r} \in \mathcal{R}}\|{C(\mathbf{r})} - \hat{C}(\mathbf{r})\|_2^2 + %\sum_{\mathcal{G} \in \mathcal{D}} 
%     \sum_{\mathcal{E}_{ij} \in \mathcal{E}}  d_{Q}({q}_{ij}^{f}, q_{ij})
%     & + \beta \sum_{\mathcal{V}_{j} \in \mathcal{V}} d_{Q}(q_{j}^{f}, q_j) \right )
%     \label{eq:complete_opt_objective}
% \end{equation}
Let's denote MLP parameters in rendering network as $\theta$ and camera pose network parameters as $\Theta$. Our objective is to optimize for the parameters $\theta$ and $\Theta$ jointly such that  Eq.(\ref{eq:loss_combined}) loss is as minimum as possible.  
Using gradient based optimization for this search process requires calculating $\nabla_{\theta} \mathcal{L}$ and $\nabla_{\Theta} \mathcal{L}$. As $\mathcal{L}_{mra}$ is independent of rendering
network, we have $\nabla_{\theta} \mathcal{L} = \nabla_{\theta} \mathcal{L}_{rgb}$. This appears to be similar as previous optimization landscape for the rendering network, but here the poses would be changing continuously resulting in different numeric value of the gradient, making the optimization difficult to converge. Now, for the pose network $\nabla_{\Theta}\mathcal{L}$ will have 2 terms: $\nabla_{\Theta}\mathcal{L}_{rgb}$ and $\nabla_{\Theta}\mathcal{L}_{mra}$. The second term is easy to handle given the pose network is able to solve the rotations as shown in \cite{purkait2020neurora}. The first term is something that would entangle the search process for $\theta$ and $\Theta$.
For clarity, let's assume the loss due to predicted color as
$\Phi(\theta, \mathcal{\gamma(P)})$, where $\mathcal{P}$ (with slight abuse of notation) denotes the poses having rotations predicted by the pose network, $\gamma$ denotes the positional encoding \citep{rahaman2019spectral}, then the gradient of $\Phi(\theta, \mathcal{\gamma(P)})$ w.r.t the pose network parameters $\Theta$  can be computed using backpropagation as:
\begin{equation}
~~~~\nabla_{\Theta}\mathcal{L}_{rgb} = \frac{\partial \Phi(\theta, \mathcal{\gamma(P)})}{\partial \Theta} = \frac{\partial \Phi(\theta, \mathcal{\gamma(P)})}{\partial \mathcal{\gamma(P)}} \frac{\partial \gamma(\mathcal{P})}{\partial \mathcal{P}} \frac{\partial \mathcal{P}}{\partial \Theta}\label{eq:joint_op_grad}
\end{equation}
% where $\gamma$ denotes the positional encoding\cite{rahaman2019spectral}.
% The above derivation shows the entangled dependency between the $\theta$ and $\Theta$ and its effect on the overall optimization via gradient decent. Since we don't have an optimal prior information on either $\theta$ or $\Theta$ to estimate a favorable gradient for optimization.
%
% This extra flow of gradient affects the convergence of the overall optimization by entangling the search process for optimal $\theta$ and $\Theta$. 
%
Differentiating this $\gamma$ function might result in updates being favourable to higher frequencies ($k$) as pointed out previously in \cite{lin2021barf}. Accordingly, we modify $\gamma$ function further to
\begin{equation}
~~~~\gamma^*(x,k) = e^{g(k)}\gamma(x),
\label{eq:gamma}
\end{equation}
where, $g(k)=\min(\frac{t-k}{b},0)$, $t$ is annealed from $0$ to maximum number of modes and $b$ is a scalar constant.
The term $\nabla_{\Theta}\mathcal{L}_{rgb}$ shown in Eq.(\ref{eq:joint_op_grad}) results in correlated updates on MLP network and pose network parameters and can result in a highly non-convex optimization.
% However, we don't have an optimal prior information on either $\theta$ or $\Theta$ to estimate a favorable gradient for optimization.
% And therefore, it is difficult to comment on the numerical stability of this gradient term and it may lead the pose network to provide arbitrary poses. 
% This joint optimization of the pose and the multi-scale image rendering can result in a highly non-convex optimization thereby making it difficult to recover a favorable solution. 
%
%landscape and can make it very difficult, especially for the pose network, to reach the optimal solution. 
%
To make optimization stable, we use the following weighted loss function:
\begin{equation}
\label{eq:weighted}
\begin{aligned}
~~~~\mathcal{L} = \lambda \mathcal{L}_{mra} + (1-\lambda) \mathcal{L}_{rgb}
\end{aligned}
\end{equation}
where $\lambda$ is a scalar constant. 
%Fig.(\ref{fig:pipeline}) provides the overall flow-diagram of our approach.
% and update the parameters of both networks simultaneously \textit{i.e.}, the error between predicted and ground truth pixel colors and the robust rotation averaging cost function. 

% To realise the joint optimization, we update the pre-trained pose network for every $N=100$ updates of the MLP rendering network. Next, we describe our joint optimization methodology.
% \textbf{Case 1 : $\lambda=0.5$}. This result in equal weights for both rendering cost function ($\mathcal{L}_{rgb}$) and robust MRA cost function ($\mathcal{L}_{mra}$). This results in  a high degree of entanglement in updating the weights of pose network and scene network throughout the optimization process.  \\
% \textbf{Case 2 : $\lambda=1$}. This results in a disjoint optimization process for the pose network and the scene network as the pose network is only updated with respect to the MRA loss ($\mathcal{L}_{mra}$) and scene network with respect to the color rendering loss ($\mathcal{L}_{rgb}$). We refer to this as the alternating optimization of the pose network and the scene and this results in further simplification of eq. \label{eq:prob} as follows:
% \begin{equation}
%     \label{eq:prob_independence}
%     \Phi({\theta},\mathcal{P}|\mathcal{I},\Tilde{\mathcal{P}}) = \overbrace{\Phi(\theta|\mathcal{I}, \mathcal{P})}^{\textrm{Multiscale MLP}} ~\cdot \overbrace{\Phi(\mathcal{P}|\mathcal{I}, \Tilde{\mathcal{P}})}^{\textrm{Motion averaging}}
% \end{equation}

\subsection{\textbf{RM-NeRF (w/o pose)}: Random initial camera poses}
%\kumar{Here!}
Our approach RM-NeRF as discussed in the previous sub-section works well in practice, given that the initial camera pose estimates are not random. The intuition behind this is that pose-refining based on MRA is used in scenarios where the amount of noise in the initial estimated poses is distributed across the global pose graph, hence we can recover a good overall solution. Yet, RM-NeRF formulation could fail if the pose graph initialization is reasonably erroneous. This puts a hard constraint of providing reasonable camera pose initialization to perform MRA.

To overcome such a practical limitation, we propose an extension to the introduced \textbf{RM-NeRF} formulation. The idea is if we have prior knowledge about the scene's geometry, we could constrain the camera motion per view at the same time and resolve radiance-geometry ambiguity in neural image rendering. To this end, we assume depth per frame is known and predicted using \citet{ranftl2021vision} network\footnote{With the recent progress in single image depth prediction (SIDP) network, it is quite a reasonable assumption.}. Given the depth prediction per view, we constrain scene point alignment via relative camera pose. Yet, solving for camera pose using such a \Rone{constraint} could lead to sub-optimal solution and requires further pose refinement. Nevertheless, it fits our purpose of using the MRA. Accordingly, we use our pose-refining GNN (same as RM-NeRF) in an iterative manner.

\textbf{RM-NeRF (w/o pose)} extension is based on the intuition that refining the camera poses estimated using 3d scene point alignment loss per view at each step via GNN can help avoid sub-optimal solution in the optimization landscape of the overall objective. Thus, we propose a loss function that allows camera poses to be initialized randomly and still be able to recover good camera pose estimates. For each step during optimization, the current poses are updated using the point cloud alignment and then refined using our pose-refining GNN.

%Let us now look at this method in detail, beginning with its formulation.

%With this method, we 

The goal is to model the network parameters ($\theta$) and the correct camera pose set ($\mathcal{P}$) using the input image set ($\mathcal{I}$) and known camera intrinsics ($\mathbf{K}$) involving an intermediate monocular depth prediction per view and reasonable camera pose set $\hat{\mathcal{P}}$ estimation using monocular depth, followed by refinement. For better abstraction, we can define the proposed intuition in terms of following equation.
%
%
%This leads to the following probabilistic formulation:
%
\begin{equation}
\begin{aligned}
\label{eq:prob_no_pose}
    \theta, \mathcal{P} &\sim \Phi({\theta},\mathcal{P} | \mathcal{I}, \mathbf{K}) \\
 & =\Phi({\theta},\mathcal{P} | \mathcal{I}, \mathbf{K}, \mathcal{D})\Phi(\mathcal{D} | \mathcal{I})\\
 & = \Phi({\theta},\mathcal{P} |\mathcal{I}, \mathbf{K}, \hat{\mathcal{P}})\Phi(\hat{\mathcal{P}} | \mathcal{D},\mathbf{K})\Phi(\mathcal{D} | \mathcal{I}).
\end{aligned}
\end{equation}
Here, $\mathcal{D}$ denotes the set of predicted depth map per frame. 
%\kumar{Here!}
% where, $\approx$ symbolizes approximation and 
% \Rthree{We have written this as an approximation because we have added a latent depth variable but without and marginalization and currently just use a monocular based method estimated depth set for this. Again we similarly condition on another initial pose latent variable but without any marginalization and only utilize the current given value. In practice, this approximation sign can be replaced by equality.}
Given that we are feeding the initial estimate $\hat{\mathcal{P}}$ into motion averaging network with parameters $\Theta$ to predict the refined pose $\mathcal{P}$, this leads to the following relation: $\mathcal{P}=f_\Theta(\hat{\mathcal{P}})$.
This allows us to randomly initialize the camera pose set $\hat{\mathcal{P}}$. Thus, given $\mathcal{I}, \mathcal{D}$, and $\hat{\mathcal{P}}$, we perform an iterative optimization by minimizing the chamfer distance $\mathcal{L}_{cd}$ of the scene points between views leading to depth and camera pose refinement.
%via $\mathcal{L}_{cd}$ and refinement via $f_\theta$. 
%\\

Similar to the concurrent work Nope-NeRF \citep{bian2022nope}, we define two learnable parameters $\alpha_i$, $\beta_i$ to transform each monocular depth $D_i$ to a global frame for multi-view consistency. Denoting transformed depth as $D_i^*$, we write 
\begin{equation}
~~~~D_i^* = \alpha_i D_i + \beta_i.
\end{equation}
Such transformation parameters is learnt by aligning the transformed and rendered depth ($\hat{D}$) via an MLP loss
\begin{equation}
~~~~\mathcal{L}_d = ||D^* - \hat{D}||_2.
\end{equation}
\Rthree{Here, $\alpha_i$, $\beta_i$ are scalar parameters. The role of $\alpha$ is to fix the scale in the monocular depth map. Whereas $\beta$ is responsible for handling the additive bias. This is because relative depth should be consistent over different camera viewpoints. A constant scale and bias factor seem sufficient to fix it for each view.}
%This is usually how the monocular depths predicted from some already freezed neural network are aligned with the SFM estimated structure.
%and thus only a constant scale and a bias factor are enough. 
%Next, we discuss the loss to learn the camera poses
Similar to \cite{bian2022nope}, assuming the known transformed depths for $i^{th}$ and $j^{th}$ image along with their relative pose $T_{ij} = {T}_j{T}^{-1}_i$, we unprojected the depth maps to scene point clouds $P_i$ and $P_j$ respectively. Here, $T_i, T_j  \in \mathcal{P}$.  
The camera pose corresponding to each of these images should be such that relative pose $T_{ji}$ aligns $P_i$ to $P_j$. Thus, the Chamfer Distance ($\mathcal{L}_{cd}$) between the $P_j$ and the transformed point cloud $T_{ji}P_i$ becomes a suitable objective function constraint for the camera poses.  Using this objective as an additional loss function, we arrive at the following overall loss function  $\mathcal{L}_{agg}$---across all the training images---to optimize the MLP parameters $\theta$, transformation parameters $\alpha$, $\beta$ and the randomly initialized set of poses $\mathcal{P}$:
\begin{equation}
    \mathcal{L}_{agg} = \left(\mathcal{L}_{mra} + \mathcal{L}_{rgb} + \mathcal{L}_d\right)+ \sum_{i,j}\mathcal{L}_{cd}\left(T_{ji}P_i, P_j\right) 
    \label{eq:ext_agg}
\end{equation}
%
%expected have a more complex cost landscape and a much
%expected have a more complex cost landscape and a much
%\kumar{Here!}
The loss proposed in Eq.\eqref{eq:ext_agg} captures our overall notions. Yet, it is complex and challenging to optimize efficiently compared to RM-NeRF, accounting for the fact that we want to allow for random camera pose initialization. To understand this better, let us look at the gradient descent-based update term of optimization variables in Eq.\eqref{eq:ext_agg}. At any optimization step $t$, the rendering MLP network parameters ($\theta^t$) are updated just w.r.t. the rendering loss $\mathcal{L}_{rgb}$ with the gradient term being $\frac{\partial\mathcal{L}_{rgb}}{\partial \theta}$ (same as RM-NeRF). On the other hand, the overall updates involved in camera pose estimation is intricate. For each step $t$, we first update the initial estimate of each pose $\hat{\mathcal{P}}^t_i$ using chamfer distance:
\begin{equation}
  \hat{\mathcal{P}}^t_i = \hat{\mathcal{P}}^{t-1}_i  - \alpha \sum_{j}\frac{\partial}{\partial \hat{\mathcal{P}}} \mathcal{L}_{cd}(T_jT^{-1}_iP_i, P_j)
  \label{eq:init_pose}
\end{equation}
Given the updated poses $\hat{\mathcal{P}}^t$ and the MLP parameters $\theta^t$, we now update the camera pose network parameters $\Theta^t$ using $\mathcal{L}_{mra}$ and $\mathcal{L}_{rgb}$ similar to \textbf{RM-NeRF} \textit{i.e.},
\begin{equation}
   \Theta^t = \Theta^{t-1} - \beta \frac{\partial}{\partial \Theta}\left(\mathcal{L}_{mra}(f_\Theta(\hat{\mathcal{P}}^t)) + \Phi(\theta^t, \gamma(f_\Theta(\hat{\mathcal{P}}^t))) \right)
   \label{eq:pose_nn}
\end{equation}
% \small
% \noindent
Note, however we use disjoint set of loss functions for updating $\hat{\mathcal{P}}^t_i$, $\Theta^t$ and therefore, do not update $\hat{\mathcal{P}}^t_i$ using $\mathcal{L}_{rgb}$ or $\mathcal{L}_{mra}$ due to the term $\frac{\partial \Theta^t}{\hat{\mathcal{P}}^t_i}$, and deal with them only using $\Theta^t$. Such updates in Eq.\eqref{eq:init_pose} and Eq.\eqref{eq:pose_nn} can be either simultaneously for each step or applied alternatively for some fixed number of steps. We follow the later strategy and update $\hat{\mathcal{P}}^t_i$, $\Theta^t$ alternatively, for a fixed number of steps ($K$).  
\subsection{\textbf{RM-NeRF (E2E)}: Unknown \Rone{Intrinsic} Camera Matrix} 
%\kumar{Here!}\sout{Intrisic}
Even though our \textbf{RM-NeRF (w/o pose)} method overcomes the requirement of good initialization of camera pose-graph variables, it still requires the intrinsic matrix for a given image set, which may not be available for real-world multi-view data. To handle this, we introduce the third extension of our algorithm referred as \textbf{RM-NeRF (E2E)}.

RM-NeRF (E2E) estimates both camera poses and intrinsic camera parameters from the multi-view image set. Additionally, it is able to work well with randomly initialized intrinsic matrix and camera poses, given the $\mathcal{D}$ is provided or predicted via a trained model. The overall loss is similar to \textbf{RM-NeRF (w/o poses)} except now the intrinsic matrices $\mathbf{K}$ is estimated leveraging the following relation among $\theta, \mathcal{P}, \mathbf{K}$.
\begin{equation}
\label{eq:prob_no_intrinsic}
~~~~\theta, \mathcal{P}\sim \Phi({\theta},\mathcal{P},\mathbf{K} |\mathcal{I},\hat{\mathcal{P}})\Phi(\hat{\mathcal{P}} |\mathcal{D})\Phi(\mathcal{D} |\mathcal{I})
\end{equation}
%The overall objective remains same as proposed in the Eq.\eqref{eq:ext_agg}, but now the $\mathbf{K}$ matrix updated alongside poses. 
Note, $\alpha$ and $\beta$ are updated only using the term $\mathcal{L}_d$ and the intrinsic ($\mathbf{K}$), extrinsic ($\mathcal{P}$) and MLP network parameters ($\theta$) are updated using the three terms except $\mathcal{L}_d$ in Eq.\eqref{eq:ext_agg}
\begin{equation}
~~~~\mathbf{K}, \mathcal{P}, \theta = \arg \min_{\mathbf{K},\mathcal{P},\theta} \left(\mathcal{L}_{mra} + \mathcal{L}_{rgb} + \mathcal{L}_{cd}\right)
\end{equation}
where, $\mathcal{L}_{cd}$ is shorthand for $\sum_{i,j}\mathcal{L}_{cd}(T_{ji},P_i,P_j)$.

\smallskip
\smallskip
\noindent
\textbf{Updating $\mathbf{K}$.} The overall objective for the updating intrinsics at step $t$ ($\mathbf{K}^t$) involves minimizing losses $\mathcal{L}_{rgb}$ and $\mathcal{L}_{cd}$: 
\begin{equation}
~~~~\mathbf{K}^t = \mathbf{K}^{t-1} - \frac{\partial}{\partial \mathbf{K}}(\mathcal{L}_{rgb} + \mathcal{L}_{cd})
\end{equation}
The intrinsics are being updated alongside camera poses $\hat{\mathcal{P}}$ w.r.t. loss $\mathcal{L}_{cd}$ and pose-refining GNN parameters $\Theta$ w.r.t. loss $\mathcal{L}_{rgb}$. Such a strategy may lead to suboptimal solution due to the complex nature of optimization. Thus, we only update intrinsics $\mathbf{K}$ using the rendering loss $\mathcal{L}_{rgb}$ alongside the GNN parameters $\Theta$. 
% \smallskip
% \noindent
\subsection{Optimization Implementation Details} 
% Given the highly non-linear nature of the joint optimization process, we adhere to the scene rigidity principle \textit{i.e.}, to intially optimize for poses without conditioning on 3D structure, and then later refine them jointly. 
\smallskip
\textbf{RM-NeRF.} We begin with a disjoint optimization scheme for camera poses and scene representation by fixing $\lambda=1$ for some initial number of epochs. For this case, Eq.\eqref{eq:prob_independence} depicts the modified formulation of the problem statement. 
% Then, we establish a dependency in their search process and increase it with time, exponentially annealing $\lambda$ from 1 to 0.5. 
% This is based on the assumption that initially the scene network will generate poor quality reconstruction even with correct poses. 
% As a result pose network might diverge to completely arbitrary poses. Hence, initially the pose network is trained by giving much more weight to the robust MRA cost function ($\mathcal{L}_{mra}$). 
%This setup is similar to \textit{warm-starting} both the networks and then optimizing them jointly. 
% \begin{equation}
%     \label{eq:prob_independence}
%     \Phi({\theta},\mathcal{P}|\mathcal{I},\Tilde{\mathcal{P}}) = \overbrace{\Phi(\theta|\mathcal{I}, \mathcal{P})}^{\textrm{Multiscale MLP}} ~\cdot \overbrace{\Phi(\mathcal{P}|\mathcal{I}, \Tilde{\mathcal{P}})}^{\textrm{Motion averaging}}
% \end{equation}
After the initial optimization of both the networks via biased weighting strategy,  $\lambda$ is annealed by using an exponential decay, \textit{i.e.}, $\lambda=\lambda_{0}e^{-kt}$ where $\lambda_{0}=1$. This annealing goes till $\lambda=0.5$ and then we fix it at $0.5$ for the remaining optimization process. 

\smallskip
\noindent
\textbf{RM-NeRF (w/o pose)}. Here, we perform initial updates using only the $\mathcal{L}_{cd}$ and $\mathcal{L}_d$ loss function for some number of epochs and then use all the loss function terms (equi-weighted) to update the variables and parameters. Our idea to train the overall model this way is a warm-up step, given that the optimization landscape can be pretty complex.

\smallskip
\noindent
\textbf{RM-NeRF (E2E)}. For this case, we first perform updates only using the $\mathcal{L}_{cd}$, $\mathcal{L}_d$ and $\mathcal{L}_{rgb}$ losses (equi-weighted). This leads to an initial estimate of the camera intrinsics and poses. Furthermore, we include the $\mathcal{L}_{mra}$ while performing updates in an equi-weighted fashion. 

\section{Experimental Setup, Results and Ablations}
\label{sec:result}
%\vspace{-2.0mm}
% Our approach requires optimization of two networks \textit{(i)} Graph Neural Network (GNN) for robust rotation averaging optimization based on message passing strategy \cite{gilmer2017neural, purkait2020neurora}, \textit{(ii)} Multi-layer Perceptron (MLP) network optimization for neural multi-scale scene representation. Our GNN architecture for pose optimization is inspired from Purkait \textit{et al.} \cite{purkait2020neurora} FineNet, whereas the MLP based rendering network is similar to Mip-NeRF \cite{barron2021mipnerf}. Fig.(\ref{fig:pipeline}) provides the overall flow-diagram of our approach. 
% \subsection{Implementational Details and Baselines}
Our overall pipeline involves optimizing parameters for two neural networks: (a) Graph Neural Network (GNN) for robust refinement of camera poses and (b) MLP network to learn the multi-scale NeRF representation for the scene. For the GNN, we follow the architecture of FineNet, proposed by \cite{purkait2020neurora}. For the MLP, we use the same architecture and sampling scheme as the Mip-NeRF paper \citep{barron2021mipnerf} (``coarse" and ``fine" sampling involving 128 samples each to render the color for a given pixel). 

% Our approach requires optimization of two networks \textit{(i)} Graph Neural Network (GNN) for robust rotation averaging optimization based on message passing strategy \citep{gilmer2017neural, purkait2020neurora}, \textit{(ii)} Multi-layer Perceptron (MLP) network optimization for neural multi-scale scene representation. Our GNN architecture for pose optimization is inspired from \cite{purkait2020neurora} FineNet, whereas the MLP based rendering network is similar to Mip-NeRF \citep{barron2021mipnerf}. 
% Fig.(\ref{fig:pipeline}) provides the overall flow-diagram of our approach. 
% For performance comparisons, we defined BARF \cite{lin2021barf} and Mip-NeRF \cite{barron2021mipnerf} as our baselines.

\noindent
Furthermore, we pre-train our pose-refining GNN in a supervised setting using a dataset comprising synthetically generated view graphs, proposed by  \cite{purkait2020neurora}. This dataset contains 1200 view graphs with up to 30\% outliers. Additionally, the dataset comprises noisy rotations where the noise is sampled from a Gaussian distribution with a standard deviation between 5\textdegree and 30\textdegree. We used 20\% of the dataset to test and train the network on the remaining examples. Once trained, our pose refinement network achieves a mean angular error (MAE) of $2.09^\circ$ and a median angular error of $1.1^\circ$ on the test set. We do this pre-training for $250$ epochs using a learning rate of $5\times10^{-5}$ and a weight decay of $10^{-4}$. Also, we drop one-fourth of the edges in the view graph to minimize overfitting at train time.

% To overcome the overfitting problem, we drop a fraction of edges (0.25) in the view-graph input to the pose network.

% The mean and median angular error values resulting from evaluating our trained pose robustifying network on this dataset are $2.09^\circ$ and $1.1^\circ$ respectively. 

% This dataset consists
% of both outliers(fraction varying between 0 and 0.3) and
% noisy rotations where noise is sampled from a normal distribution with standard deviation varying uniformly between 5° and 30°. 
% The training dataset comprises sparse view-graphs with ground-truth absolute poses divided into 80\% training, 10\% validation, and 10\% testing. 
%\kumar{Here!}

The trained pose-refinement network is now updated alongside the scene representation MLP network for each scene. Given refined rotations, a linear optimizer is employed to solve translation \citep{chatterjee2017robust}.
%We estimate the final refined translations using these predicted rotations.
%solving for translations is a linear optimization and thus, we estimate the final refined translations using these predicted rotations.
% Using these refined rotations, we estimate the translations which are expected to be robust as given rotations, solving for translations is a linear optimization process. 
% Later, we use the pre-trained pose network for our overall optimization. We implemented our approach using JAX \citep{jax2018github} and simulated our code on a 32 GB Nvidia V100 GPU computing machine. 
We set the value of hyperparameter $\beta$ (used in Eq.\eqref{eq:loss_mra}) to 0.1 during both the pretraining and pose refinement stages. Similarly, we fix the value of hyperparameter $b$ from Eq.\eqref{eq:gamma} to be 10 for all the experiments. The training of our proposed involves optimizing the MLP network for 100k iterations per scene using the Adam Optimizer \citep{kingma2014adam}. We use a batch size of 4096 and a logarithmically varying learning rate (between $5\times10^{-4}$ and $5\times10^{-5}$).
% The $\beta$ parameter  in  Eq. \eqref{eq:loss_mra} loss function, in the main paper, is set to 0.1 at train time.  
 % Also, the value of parameter $b$ used in Eq. \eqref{eq:gamma} is kept fixed at 10. The MLP model for learning the multi-scale scene representation is trained for 100k iterations for every scene using the Adam optimizer \citep{kingma2014adam} with a batch size of 4096 rays and a learning rate varying logarithmically from $5\times10^{-4}$ to $5\times10^{-5}$. Similar to \cite{barron2021mipnerf}, we use ``coarse" and ``fine" sampling schemes with 128 samples for rendering.
  All of our experiments have been carried out on a 32 GB Nvidia V100 GPU computing machine.
\subsection{Test sets and Results}\label{sec:mutlib}

We evaluate our proposed method under two settings: 

\smallskip
\noindent
\textbf{\textit{(i)}} This setting involves analysis of the introduced approach on a synthetic dataset. This dataset is generated by rendering images of 3D object using a pre-defined and well structured camera poses covering 360\textdegree~view of the object. Thus, ground-truth poses are well defined. For this experiments, we use the Blender dataset provided by \cite{mildenhall2020nerf}\footnote{CC-BY-3.0 license.} and its multi-scaled version \cite{barron2021mipnerf}, consisting of single object scene centered around a single object without any background. Each scene in this dataset comprises of 100 images  with $800\times800$ resolution with is captured by moving the camera along a fixed hemisphere surrounding this object. We simulate realistic scenario for this dataset by (a) using the multi-scaled version representing varying distance of camera from the object and (b) adding noise to the ground truth poses.
% To compare our method with the baselines, we used the Blender dataset provided by the authors of NeRF \citep{mildenhall2020nerf}\footnote{CC-BY-3.0 license.} and its multi-scale version provided by Barron \textit{et al.} \citep{barron2021mipnerf}. It consists of single object scenes comprising synthetic objects and corresponding ground truth poses, with each scene consisting of $M=100$ images with $800\times800$ resolution.
% It consists of single object scenes comprising synthetic objects and corresponding ground truth poses, with each scene consisting of $M=100$ images with $800\times800$ resolution. 
% We have used seven scenes for our evaluation, namely \textrm{Lego, Ship, Drums, Mic, Chair, Ficus} and \textrm{Materials}. 
% To simulate the effect of pose estimation errors in real-world datasets, we add noise to the ground truth poses. 

\smallskip
\noindent
\textbf{\textit{(ii)}} This setting tests our method on the real-world images which are acquired by a freely moving camera with no access to ground-truth camera parameters. The camera parameters have to  \Rone{be} estimated. For this, we use the Tanks and Temples dataset \citep{Knapitsch2017}, comprising various scenes from indoor and outdoor real-world settings. Other than this, we test our approach on the popular ScanNet dataset \citep{dai2017scannet} comprising a diverse set of indoor scenes. Additionally, we show results on two other datasets. This first one captures a box using a regular phone with arbitrary camera trajectory, and the second is a scene taken from a recent work by \cite{yen2022nerf}. 

More details regarding each dataset are provided in the following subsections.

\smallskip
\noindent
\subsubsection{Multi-Scaled Images of Object}
\label{sec:multi_blender}
For evaluating our method in an object-centric synthetic setting comprising a complete 360\textdegree~view, we study the multi-scaled version of the NeRF Blender dataset, proposed by \cite{barron2021mipnerf}. It is generated by resizing each image in the NeRF Blender dataset to three different \Rone,\Rtwo{resolutions} and concatenating them with the original dataset. Resizing these images does not change the ground-truth poses, however the intrinsics for images at each resolution are updated accordingly.
% consists of 400 image scenes generated by scaling every image in the original Blender dataset to 4 different resolutions. These different resolution images are synthesized by concatenation of actual resolution images with downsampled images by a factor of 2, 4, and 8. 
This resizing can be interpreted as changing the camera distance from the center of the object and thus, this dataset is more closely \Rone{aligned} with real-world setting as compared to Blender dataset \citep{mildenhall2020nerf}.
%\sout{algined}
%\sout{resoltuions}
% which consists of images captured from the exact same distance
% This scale can also be interpreted as the distance of the object from the camera. Therefore, it resembles the real-world datasets much more closely than the original Blender dataset, which contains all the images at the same image resolution and nearly similar distances. The ground truth extrinsic poses are the same as the original dataset, but the camera intrinsics are changed according to the image resolution.

%We now make this dataset more closely mimic the real world setting by perturbing the ground truth poses. 
%

We further perturb the ground-truth camera poses to make this dataset close to the real-world setting. Given all the cameras in ground-truth poses lie on a hemisphere, we limit ourselves to only perturbing the rotational parameters. For this, we first sample Gaussian noise $\delta$\textbf{p} $\sim \mathcal{N}(\textbf{0},1e^{-1}\textbf{I})$ in the axis angle space, convert it into rotation matrix representation and multiply it with the rotations of the ground truth poses, hence perturbing ground-truth camera orientation.

% We perturb the poses for every scene by first sampling the noise from a normal distribution 

\smallskip
\noindent
\textbf{Baselines.} We compare the proposed RM-NeRF, RM-NeRF (w/o pose) methods with the Mip-NeRF, NeRF$-$$-$\citep{wang2021nerf}, BARF \citep{lin2021barf} baselines to highlight their effectiveness in dealing with camera pose errors and multi-scale images simultaneously.
To further demonstrate the challenges with this dataset setting, we define three new baselines and evaluate them against our methods. These baselines are designed by combining existing baselines that can deal with pose errors and multi-scale issues separately. This leads to the first baseline (\textbf{Base A}) being a result of directly combining Mip-NeRF and BARF by replacing the positional encoding function to create the Mip-NeRF Integrated Positional Encoding with the pose encoding function used by BARF. Both its components, Mip-NeRF and BARF, when used separately, can either solve multi-scale issues (Mip-NeRF) or pose errors (BARF). On similar lines, we define a second baseline (\textbf{Base B}) which involves feeding BARF estimated poses to the Mip-NeRF multi-scale modeling scheme.
Finally, we define the third baseline (\textbf{Base C}), which combines Mip-NeRF with NeRF$-$$-$ by replacing the positional encoding scheme in NeRF$-$$-$ with the Integrated Positional encoding that uses frustum-based volumetric modeling for each pixel, instead of rays. Other than these baselines, we compare our proposed methods against the recent Point-NeRF \citep{xu2022point} method, which has shown to be quite efficient in convergence leading to high-quality renderings on the original Blender dataset \citep{mildenhall2020nerf}.

% in its axis-angle form. It is then converted to the rotation matrix representation and multiplied with the ground truth poses, disturbing their orientations. 
% We did this purposely to make the dataset resemble real-world settings closely, thus making it challenging to learn the multi-scale scene representation. 
% 
\begin{table*}[t]
\centering
%  \scriptsize
% \resizebox{\columnwidth}{!}
\resizebox{\textwidth}{!}
{\begin{tabular}{ccccccccccccccccc}
        %\hline
        \toprule
        \multicolumn{1}{c|}{}                                        & \multicolumn{2}{c|}{Lego}                         &  \multicolumn{2}{c|}{Ship}     & \multicolumn{2}{c|}{Drums}   & \multicolumn{2}{c|}{Mic}
        & \multicolumn{2}{c|}{Chair}
        & \multicolumn{2}{c|}{Ficus} & \multicolumn{2}{c|}{Materials}  &
        \multicolumn{2}{c}{Hotdog} \\
        
        \multicolumn{1}{c|}{}                                                                          & \multicolumn{1}{c}{{\fontsize{6.5}{4}\selectfont PSNR$\uparrow$}} & \multicolumn{1}{c|}{{\fontsize{6.5}{4}\selectfont LPIPS$\downarrow$}} &  \multicolumn{1}{c}{{\fontsize{6.5}{4}\selectfont PSNR$\uparrow$}} & \multicolumn{1}{c|}{{\fontsize{6.5}{4}\selectfont LPIPS$\downarrow$}} &                     \multicolumn{1}{c}{{\fontsize{6.5}{4}\selectfont PSNR$\uparrow$}} & \multicolumn{1}{c|}{{\fontsize{6.5}{4}\selectfont LPIPS$\downarrow$}} & \multicolumn{1}{c}{{\fontsize{6.5}{4}\selectfont PSNR$\uparrow$}} & \multicolumn{1}{c|}{{\fontsize{6.5}{4}\selectfont LPIPS$\downarrow$}} & 
        \multicolumn{1}{c}{{\fontsize{6.5}{4}\selectfont PSNR$\uparrow$}} & \multicolumn{1}{c|}{{\fontsize{6.5}{4}\selectfont LPIPS$\downarrow$}} & 
        \multicolumn{1}{c}{{\fontsize{6.5}{4}\selectfont PSNR$\uparrow$}} & \multicolumn{1}{c|}{{\fontsize{6.5}{4}\selectfont LPIPS$\downarrow$}}
        & 
        \multicolumn{1}{c}{{\fontsize{6.5}{4}\selectfont PSNR$\uparrow$}} & \multicolumn{1}{c|}{{\fontsize{6.5}{4}\selectfont LPIPS$\downarrow$}}
        &
        \multicolumn{1}{c}{{\fontsize{6.5}{4}\selectfont PSNR$\uparrow$}} & \multicolumn{1}{c}{{\fontsize{6.5}{4}\selectfont LPIPS$\downarrow$}}
        
        \\ \midrule
        % \multicolumn{1}{c|}{\begin{tabular}[c|]{@{}c@{}}NeRF\end{tabular}}          & Passed  & Passed & \multicolumn{1}{c|}{100\%} & Passed & Passed & \multicolumn{1}{c|}{100\%}  & Passed  & Passed & \multicolumn{1}{c|}{100\%}   & Passed  & Passed & \multicolumn{1}{c}{100\%}                      \\ 
        %\cellcolor[HTML]{ffff80}
        %\cellcolor[HTML]{ff9090}
        %\hline
        \multicolumn{1}{c|}{\begin{tabular}[c]{@{}c@{}}Mip-NeRF\end{tabular}} & 21.52   & \multicolumn{1}{c|}{0.06} &  24.54 &  \multicolumn{1}{c|}{ 0.07} & 13.34   & \multicolumn{1}{c|}{0.075}    &  24.71  & \multicolumn{1}{c|}{ 0.05} & 29.1  &  \multicolumn{1}{c|}{0.049}  & 22.47  &  \multicolumn{1}{c|}{0.055} & 19.7  &  \multicolumn{1}{c|}{0.089}& 27.09  &  \multicolumn{1}{c}{0.053}        \\ %\hline
        \multicolumn{1}{c|}{\begin{tabular}[c|]{@{}c@{}}BARF\end{tabular}}          & 10.88   & \multicolumn{1}{c|}{0.55} & 8.81 &  \multicolumn{1}{c|}{0.74}  & 11.56  &  \multicolumn{1}{c|}{0.76}  & 12.35  &  \multicolumn{1}{c|}{0.57}  & 14.35  &  \multicolumn{1}{c|}{0.47}  & 11.88  &  \multicolumn{1}{c|}{0.65}  & 12.28  &  \multicolumn{1}{c|}{0.61}
        & 14.28  &  \multicolumn{1}{c}{0.46}
        \\     
        
        \multicolumn{1}{c|}{\begin{tabular}[c]{@{}c@{}}Base A\end{tabular}} & 11.67   & \multicolumn{1}{c|}{0.49} &  14.28 &  \multicolumn{1}{c|}{0.28} & 13.25  &  \multicolumn{1}{c|}{0.67}    & 12.28  & \multicolumn{1}{c|}{0.41}  & 15.12  &  \multicolumn{1}{c|}{0.20}  & 12.31  &  \multicolumn{1}{c|}{0.25} & 13.31  &  \multicolumn{1}{c|}{0.42}        & 16.17  &  \multicolumn{1}{c}{0.39}  \\ %\hline
        \multicolumn{1}{c|}{\begin{tabular}[c]{@{}c@{}}Base B\end{tabular}} & 12.46 &  \multicolumn{1}{c|}{0.37} & 13.43 &   \multicolumn{1}{c|}{0.31} & 11.32  &  \multicolumn{1}{c|}{0.58}    &  14.26  &  \multicolumn{1}{c|}{ 0.29} & 13.71  &  \multicolumn{1}{c|}{0.42}  & 11.56  &  \multicolumn{1}{c|}{0.52}     & 12.22  &  \multicolumn{1}{c|}{0.47}& 15.87  &  \multicolumn{1}{c}{0.42}      \\ 
        \midrule
         \multicolumn{1}{c|}{\begin{tabular}[c]{@{}c@{}}NeRF--\end{tabular}} & 16.89 &  \multicolumn{1}{c|}{0.094} &  19.89 &   \multicolumn{1}{c|}{0.12} & 15.67  &  \multicolumn{1}{c|}{0.074}    &  18.35  &  \multicolumn{1}{c|}{ 0.08}  & 20.22  &  \multicolumn{1}{c|}{0.098}  & 14.44  &  \multicolumn{1}{c|}{0.13} & 15.77  &  \multicolumn{1}{c|}{0.22} & 18.69  &  \multicolumn{1}{c}{0.20}         \\ %\hline
        %  \multicolumn{1}{c|}{\begin{tabular}[c]{@{}c@{}}Mip-NeRF+COLMAP\end{tabular}} &  \textbf{27.88} &   \multicolumn{1}{c|}{ \textbf{0.044}} &  \textbf{26.97} &    \multicolumn{1}{c|}{ \textbf{0.071}} &  \textbf{25.7}   & \multicolumn{1}{c|}{ \textbf{0.058}} &  \textbf{32.67}  &   \multicolumn{1}{c|}{ \textbf{0.008}} & \textbf{35.3}  &  \multicolumn{1}{c|}{\textbf{0.031}}  & \textbf{29.2}  &  \multicolumn{1}{c}{\textbf{0.035}}         \\
        %  \hline
        % \multicolumn{1}{c|}{\begin{tabular}[c]{@{}c@{}}RM-NeRF(alternating)\end{tabular}} &  \textbf{26.68}   & \multicolumn{1}{c|}{ \textbf{0.045}} &  \textbf{26.76}  & \multicolumn{1}{c|}{ \textbf{0.065}} &  \textbf{25.823}  &   \multicolumn{1}{c|}{ \textbf{0.046}} &  \textbf{32.087}  & \multicolumn{1}{c|}{ \textbf{0.008}} &  \textbf{35.1}  & \multicolumn{1}{c|}{ \textbf{0.032}} &  \textbf{29.12}  & \multicolumn{1}{c}{ \textbf{0.034}} \\ %
        % \multicolumn{1}{c|}{\begin{tabular}[c]{@{}c@{}}RM-NeRF(joint)\end{tabular}} &  22.2 &   \multicolumn{1}{c|}{ 0.058} &  23.34  & \multicolumn{1}{c|}{ 0.074} &  15.62  &   \multicolumn{1}{c|}{ 0.068} &  22.61  &   \multicolumn{1}{c|}{ 0.041} &  25.7  &   \multicolumn{1}{c|}{ 0.068} &  22.7  &   \multicolumn{1}{c}{ 0.061} \\
        \multicolumn{1}{c|}{\begin{tabular}[c]{@{}c@{}}Base C\end{tabular}} & 18.28 &  \multicolumn{1}{c|}{0.089} &  16.32 &   \multicolumn{1}{c|}{0.22} & 17.25  &  \multicolumn{1}{c|}{0.070}    &  19.42  &  \multicolumn{1}{c|}{ 0.073}  & 18.67  &  \multicolumn{1}{c|}{0.114}  & 16.32  &  \multicolumn{1}{c|}{0.12} & 16.58  &  \multicolumn{1}{c|}{0.207} & 17.55  &  \multicolumn{1}{c}{0.223}         \\

        \multicolumn{1}{c|}{\begin{tabular}[c]{@{}c@{}}Point-NeRF\end{tabular}} & 20.12 &  \multicolumn{1}{c|}{0.079} &  22.47 &   \multicolumn{1}{c|}{0.10} & 18.32  &  \multicolumn{1}{c|}{0.066}    &  22.37  &  \multicolumn{1}{c|}{ 0.061}  & 26.67  &  \multicolumn{1}{c|}{0.077}  & 20.23  &  \multicolumn{1}{c|}{0.107} & 17.23  &  \multicolumn{1}{c|}{0.112} & 24.45  &  \multicolumn{1}{c}{0.092}         \\ 

        \midrule
        \multicolumn{17}{c}{Ours} \\
        \midrule
        % \multicolumn{1}{c|}{\begin{tabular}[c]{@{}c@{}} \textit{naive}\end{tabular}} & \cellcolor[HTML]{ff9090} \textbf{ 27.01}    &  \multicolumn{1}{c|}{\cellcolor[HTML]{ff9090} \textbf{0.044}} & \cellcolor[HTML]{ff9090} \textbf{ 26.59}  & \multicolumn{1}{c|}{\cellcolor[HTML]{ff9090} \textbf{ 0.067}} & \cellcolor[HTML]{ff9090} \textbf{ 26.07}   & \multicolumn{1}{c|}{\cellcolor[HTML]{ff9090} \textbf{0.043}} & \cellcolor[HTML]{ff9090} \textbf{32.8}  &   \multicolumn{1}{c|}{\cellcolor[HTML]{ff9090} \textbf{0.008}} &  \cellcolor[HTML]{ff9090}\textbf{35.23}  &   \multicolumn{1}{c|}{\cellcolor[HTML]{ff9090} \textbf{0.031}} &  \cellcolor[HTML]{ff9090}\textbf{29.28}  &   \multicolumn{1}{c|}{\cellcolor[HTML]{ff9090}\textbf{0.032}} & \cellcolor[HTML]{ff9090}\textbf{24.8}  &  \multicolumn{1}{c|}{\cellcolor[HTML]{ff9090}\textbf{0.061}}
        % & \cellcolor[HTML]{ff9090}\textbf{32.5}  &  \multicolumn{1}{c}{\cellcolor[HTML]{ff9090}\textbf{0.028}} \\

        \multicolumn{1}{c|}{\begin{tabular}[c]{@{}c@{}} \textit{RM-NeRF}\end{tabular}} &  { 27.01}    &  \multicolumn{1}{c|}{ {0.044}} &  { 26.59}  & \multicolumn{1}{c|}{ { 0.067}} &  { 26.07}   & \multicolumn{1}{c|}{ {0.043}} &  {32.8}  &   \multicolumn{1}{c|}{ {0.008}} &  {35.23}  &   \multicolumn{1}{c|}{ {0.031}} &  {29.28}  &   \multicolumn{1}{c|}{{0.032}} & {24.8}  &  \multicolumn{1}{c|}{{0.061}}
        & {32.5}  &  \multicolumn{1}{c}{{0.028}} \\

        \multicolumn{1}{c|}{\begin{tabular}[c]{@{}c@{}}RM-NeRF {(w/o pose)}\end{tabular}} & 26.34 &  \multicolumn{1}{c|}{0.048} &  26.02 &   \multicolumn{1}{c|}{0.081} & 25.23  &  \multicolumn{1}{c|}{0.051}    &  32.1  &  \multicolumn{1}{c|}{ 0.009}  & 34.67  &  \multicolumn{1}{c|}{0.042}  & 29.04  &  \multicolumn{1}{c|}{0.044} & 23.2  &  \multicolumn{1}{c|}{0.072} & 31.7  &  \multicolumn{1}{c}{0.030}         \\

         \midrule
        \multicolumn{17}{c}{With Same Initialization as RM-NeRF} \\
        \midrule
        % \multicolumn{1}{c|}{\begin{tabular}[c]{@{}c@{}}E2E\end{tabular}} & 18.28 &  \multicolumn{1}{c|}{0.089} &  16.32 &   \multicolumn{1}{c|}{0.22} & 17.25  &  \multicolumn{1}{c|}{0.070}    &  19.42  &  \multicolumn{1}{c|}{ 0.073}  & 18.67  &  \multicolumn{1}{c|}{0.114}  & 16.32  &  \multicolumn{1}{c|}{0.12} & 16.58  &  \multicolumn{1}{c|}{0.207} & 17.55  &  \multicolumn{1}{c}{0.223}         \\

\multicolumn{1}{c|}{\begin{tabular}[c]{@{}c@{}}RM-NeRF {(w/o pose)}\end{tabular}} & \textbf{27.91} &  \multicolumn{1}{c|}{\textbf{0.033}} &  \textbf{27.23} &   \multicolumn{1}{c|}{\textbf{0.056}} & \textbf{26.37}  &  \multicolumn{1}{c|}{\textbf{0.046}}    &  \textbf{32.9}  &  \multicolumn{1}{c|}{ \textbf{0.008}}  & \textbf{35.98}  &  \multicolumn{1}{c|}{\textbf{0.029}}  & \textbf{30.12}  &  \multicolumn{1}{c|}{\textbf{0.031}} & \textbf{25.8}  &  \multicolumn{1}{c|}{\textbf{0.052}} & \textbf{32.8}  &  \multicolumn{1}{c}{\textbf{0.021}}         \\

        \bottomrule
       
    \end{tabular}
}
\vspace{1.00mm}
\caption{\footnotesize 
\footnotesize Comparison of our proposed methods  (RM-NeRF and RM-NeRF (w/o pose)) with the baselines, on the Multi-scale Blender dataset using synthetically perturbed poses. Our method shows significantly better performance as compared to the existing BARF \citep{lin2021barf}, Mip-NeRF \citep{barron2021mipnerf} NeRF-- \citep{wang2021nerf} baselines and the newly defined Base A, Base B, Base C baselines. Furthermore, these results reflect that our method can efficiently handle multi-scale issues and pose errors simultaneously. 
}\label{table:1}
\end{table*}

\begin{table*}[t]
\centering
%  \scriptsize
% \resizebox{\columnwidth}{!}
\resizebox{\textwidth}{!}
{\begin{tabular}{ccccccccccccccccc}
        %\hline
        \toprule
        \multicolumn{1}{c|}{}                                        & \multicolumn{2}{c|}{Lego}                         &  \multicolumn{2}{c|}{Ship}     & \multicolumn{2}{c|}{Drums}   & \multicolumn{2}{c|}{Mic}
        & \multicolumn{2}{c|}{Chair}
        & \multicolumn{2}{c|}{Ficus} & \multicolumn{2}{c|}{Materials}  &
        \multicolumn{2}{c}{Hotdog} \\
        
        \multicolumn{1}{c|}{}                                                                          & \multicolumn{1}{c}{{\fontsize{6.5}{4}\selectfont PSNR$\uparrow$}} & \multicolumn{1}{c|}{{\fontsize{6.5}{4}\selectfont LPIPS$\downarrow$}} &  \multicolumn{1}{c}{{\fontsize{6.5}{4}\selectfont PSNR$\uparrow$}} & \multicolumn{1}{c|}{{\fontsize{6.5}{4}\selectfont LPIPS$\downarrow$}} &                     \multicolumn{1}{c}{{\fontsize{6.5}{4}\selectfont PSNR$\uparrow$}} & \multicolumn{1}{c|}{{\fontsize{6.5}{4}\selectfont LPIPS$\downarrow$}} & \multicolumn{1}{c}{{\fontsize{6.5}{4}\selectfont PSNR$\uparrow$}} & \multicolumn{1}{c|}{{\fontsize{6.5}{4}\selectfont LPIPS$\downarrow$}} & 
        \multicolumn{1}{c}{{\fontsize{6.5}{4}\selectfont PSNR$\uparrow$}} & \multicolumn{1}{c|}{{\fontsize{6.5}{4}\selectfont LPIPS$\downarrow$}} & 
        \multicolumn{1}{c}{{\fontsize{6.5}{4}\selectfont PSNR$\uparrow$}} & \multicolumn{1}{c|}{{\fontsize{6.5}{4}\selectfont LPIPS$\downarrow$}}
        & 
        \multicolumn{1}{c}{{\fontsize{6.5}{4}\selectfont PSNR$\uparrow$}} & \multicolumn{1}{c|}{{\fontsize{6.5}{4}\selectfont LPIPS$\downarrow$}}
        &
        \multicolumn{1}{c}{{\fontsize{6.5}{4}\selectfont PSNR$\uparrow$}} & \multicolumn{1}{c}{{\fontsize{6.5}{4}\selectfont LPIPS$\downarrow$}}
        
        \\ \midrule
        % \multicolumn{1}{c|}{\begin{tabular}[c|]{@{}c@{}}NeRF\end{tabular}}          & Passed  & Passed & \multicolumn{1}{c|}{100\%} & Passed & Passed & \multicolumn{1}{c|}{100\%}  & Passed  & Passed & \multicolumn{1}{c|}{100\%}   & Passed  & Passed & \multicolumn{1}{c}{100\%}                      \\ 
        %\cellcolor[HTML]{ffff80}
        %\cellcolor[HTML]{ff9090}
        %\hline
        \multicolumn{1}{c|}{\begin{tabular}[c]{@{}c@{}}Mip-NeRF\end{tabular}} & 19.37   & \multicolumn{1}{c|}{0.09} &  22.13 &  \multicolumn{1}{c|}{ 0.10} & 12.28   & \multicolumn{1}{c|}{0.089}    &  22.99  & \multicolumn{1}{c|}{ 0.09} & 28.2  &  \multicolumn{1}{c|}{0.066}  & 21.36  &  \multicolumn{1}{c|}{0.065} & 17.9  &  \multicolumn{1}{c|}{0.095}& 26.23 &  \multicolumn{1}{c}{0.073}        \\ %\hline
        \multicolumn{1}{c|}{\begin{tabular}[c|]{@{}c@{}}BARF\end{tabular}}          & 9.98   & \multicolumn{1}{c|}{0.65} & 8.831 &  \multicolumn{1}{c|}{0.84}  & 10.34  &  \multicolumn{1}{c|}{0.91}  & 11.23  &  \multicolumn{1}{c|}{0.77}  & 13.12  &  \multicolumn{1}{c|}{0.71}  & 10.67  &  \multicolumn{1}{c|}{0.76}  & 11.88  &  \multicolumn{1}{c|}{0.77}
        & 12.98  &  \multicolumn{1}{c}{0.56}
        \\     
        
        % \multicolumn{1}{c|}{\begin{tabular}[c]{@{}c@{}}Base A\end{tabular}} & 11.67   & \multicolumn{1}{c|}{0.49} &  14.28 &  \multicolumn{1}{c|}{0.28} & 13.25  &  \multicolumn{1}{c|}{0.67}    & 12.28  & \multicolumn{1}{c|}{0.41}  & 15.12  &  \multicolumn{1}{c|}{0.20}  & 12.31  &  \multicolumn{1}{c|}{0.25} & 13.31  &  \multicolumn{1}{c|}{0.42}        & 16.17  &  \multicolumn{1}{c}{0.39}  \\ %\hline
        % \multicolumn{1}{c|}{\begin{tabular}[c]{@{}c@{}}Base B\end{tabular}} & 12.46 &  \multicolumn{1}{c|}{0.37} & 13.43 &   \multicolumn{1}{c|}{0.31} & 11.32  &  \multicolumn{1}{c|}{0.58}    &  14.26  &  \multicolumn{1}{c|}{ 0.29} & 13.71  &  \multicolumn{1}{c|}{0.42}  & 11.56  &  \multicolumn{1}{c|}{0.52}     & 12.22  &  \multicolumn{1}{c|}{0.47}& 15.87  &  \multicolumn{1}{c}{0.42}      \\ 
        % \midrule
         \multicolumn{1}{c|}{\begin{tabular}[c]{@{}c@{}}NeRF--\end{tabular}} & 16.89 &  \multicolumn{1}{c|}{0.094} &  19.89 &   \multicolumn{1}{c|}{0.12} & 15.67  &  \multicolumn{1}{c|}{0.074}    &  18.35  &  \multicolumn{1}{c|}{ 0.08}  & 20.22  &  \multicolumn{1}{c|}{0.098}  & 14.44  &  \multicolumn{1}{c|}{0.13} & 15.77  &  \multicolumn{1}{c|}{0.22} & 18.69  &  \multicolumn{1}{c}{0.20}         \\ %\hline

        \midrule
        \multicolumn{17}{c}{Ours} \\
        \midrule
        % \multicolumn{1}{c|}{\begin{tabular}[c]{@{}c@{}} \textit{naive}\end{tabular}} & \cellcolor[HTML]{ff9090} \textbf{ 27.01}    &  \multicolumn{1}{c|}{\cellcolor[HTML]{ff9090} \textbf{0.044}} & \cellcolor[HTML]{ff9090} \textbf{ 26.59}  & \multicolumn{1}{c|}{\cellcolor[HTML]{ff9090} \textbf{ 0.067}} & \cellcolor[HTML]{ff9090} \textbf{ 26.07}   & \multicolumn{1}{c|}{\cellcolor[HTML]{ff9090} \textbf{0.043}} & \cellcolor[HTML]{ff9090} \textbf{32.8}  &   \multicolumn{1}{c|}{\cellcolor[HTML]{ff9090} \textbf{0.008}} &  \cellcolor[HTML]{ff9090}\textbf{35.23}  &   \multicolumn{1}{c|}{\cellcolor[HTML]{ff9090} \textbf{0.031}} &  \cellcolor[HTML]{ff9090}\textbf{29.28}  &   \multicolumn{1}{c|}{\cellcolor[HTML]{ff9090}\textbf{0.032}} & \cellcolor[HTML]{ff9090}\textbf{24.8}  &  \multicolumn{1}{c|}{\cellcolor[HTML]{ff9090}\textbf{0.061}}
        % & \cellcolor[HTML]{ff9090}\textbf{32.5}  &  \multicolumn{1}{c}{\cellcolor[HTML]{ff9090}\textbf{0.028}} \\

        \multicolumn{1}{c|}{\begin{tabular}[c]{@{}c@{}} \textit{RM-NeRF}\end{tabular}} &  { 26.10}    &  \multicolumn{1}{c|}{ {0.071}} &  { 26.12}  & \multicolumn{1}{c|}{ { 0.081}} &  { 25.24}   & \multicolumn{1}{c|}{ {0.083}} &  {31.6}  &   \multicolumn{1}{c|}{ {0.012}} &  {34.11}  &   \multicolumn{1}{c|}{ {0.053}} &  {28.43}  &   \multicolumn{1}{c|}{{0.042}} & {24.1}  &  \multicolumn{1}{c|}{{0.073}}
        & {31.3}  &  \multicolumn{1}{c}{{0.052}} \\

        \multicolumn{1}{c|}{\begin{tabular}[c]{@{}c@{}}RM-NeRF {(w/o pose)}\end{tabular}} & 26.34 &  \multicolumn{1}{c|}{0.048} &  26.02 &   \multicolumn{1}{c|}{0.081} & 25.23  &  \multicolumn{1}{c|}{0.051}    &  32.1  &  \multicolumn{1}{c|}{ 0.009}  & 34.67  &  \multicolumn{1}{c|}{0.042}  & 29.04  &  \multicolumn{1}{c|}{0.044} & 23.2  &  \multicolumn{1}{c|}{0.072} & 31.7  &  \multicolumn{1}{c}{0.030}         \\

        % \multicolumn{1}{c|}{\begin{tabular}[c]{@{}c@{}}E2E\end{tabular}} & 18.28 &  \multicolumn{1}{c|}{0.089} &  16.32 &   \multicolumn{1}{c|}{0.22} & 17.25  &  \multicolumn{1}{c|}{0.070}    &  19.42  &  \multicolumn{1}{c|}{ 0.073}  & 18.67  &  \multicolumn{1}{c|}{0.114}  & 16.32  &  \multicolumn{1}{c|}{0.12} & 16.58  &  \multicolumn{1}{c|}{0.207} & 17.55  &  \multicolumn{1}{c}{0.223}         \\
                 \midrule
        \multicolumn{17}{c}{With Same Initialization  of RM-NeRF (w/o pose) as RM-NeRF} \\
        \midrule
        % \multicolumn{1}{c|}{\begin{tabular}[c]{@{}c@{}}E2E\end{tabular}} & 18.28 &  \multicolumn{1}{c|}{0.089} &  16.32 &   \multicolumn{1}{c|}{0.22} & 17.25  &  \multicolumn{1}{c|}{0.070}    &  19.42  &  \multicolumn{1}{c|}{ 0.073}  & 18.67  &  \multicolumn{1}{c|}{0.114}  & 16.32  &  \multicolumn{1}{c|}{0.12} & 16.58  &  \multicolumn{1}{c|}{0.207} & 17.55  &  \multicolumn{1}{c}{0.223}         \\

\multicolumn{1}{c|}{\begin{tabular}[c]{@{}c@{}}RM-NeRF {(w/o pose)}\end{tabular}} & \textbf{27.23} &  \multicolumn{1}{c|}{\textbf{0.039}} &  \textbf{26.95} &   \multicolumn{1}{c|}{\textbf{0.068}} & \textbf{26.06}  &  \multicolumn{1}{c|}{\textbf{0.051}}    &  \textbf{32.3}  &  \multicolumn{1}{c|}{ \textbf{0.010}}  & \textbf{34.88}  &  \multicolumn{1}{c|}{\textbf{0.034}}  & \textbf{29.34}  &  \multicolumn{1}{c|}{\textbf{0.043}} & \textbf{25.1}  &  \multicolumn{1}{c|}{\textbf{0.061}} & \textbf{32.1}  &  \multicolumn{1}{c}{\textbf{0.033}}         \\

        \bottomrule
       
    \end{tabular}
}
\vspace{1.00mm}
\caption{\Rthree{\footnotesize 
\footnotesize Comparison of our proposed methods  (RM-NeRF and RM-NeRF (w/o pose)) with the baselines, on the Multi-scale Blender dataset using synthetically perturbed poses, this time adding noise to both rotational and translational parameters. Again, our methods shows significantly better performance as compared the other methods thereby verifying their effectiveness in simultaneous presence of rotational, translational errors and multi-scale issues.}
}\label{table:2_rot_trans_blender}
\end{table*}

\begin{table*}[t]
\centering
%  \scriptsize
%\resizebox{\columnwidth}{!}
\resizebox{\textwidth}{!}
{\begin{tabular}{ccccccccccccccccc}
        \toprule
        \multicolumn{1}{c|}{}                                        & \multicolumn{2}{c|}{Lego}                         &  \multicolumn{2}{c|}{Ship}     & \multicolumn{2}{c|}{Drums}   & \multicolumn{2}{c|}{Mic}
        & \multicolumn{2}{c|}{Chair}
        & \multicolumn{2}{c|}{Ficus} & \multicolumn{2}{c|}{Materials}  &
        \multicolumn{2}{c}{Hotdog} \\
        
        \multicolumn{1}{c|}{}                                                                          & \multicolumn{1}{c}{{\fontsize{6.5}{4}\selectfont PSNR$\uparrow$}} & \multicolumn{1}{c|}{{\fontsize{6.5}{4}\selectfont LPIPS$\downarrow$}} &  \multicolumn{1}{c}{{\fontsize{6.5}{4}\selectfont PSNR$\uparrow$}} & \multicolumn{1}{c|}{{\fontsize{6.5}{4}\selectfont LPIPS$\downarrow$}} &                     \multicolumn{1}{c}{{\fontsize{6.5}{4}\selectfont PSNR$\uparrow$}} & \multicolumn{1}{c|}{{\fontsize{6.5}{4}\selectfont LPIPS$\downarrow$}} & \multicolumn{1}{c}{{\fontsize{6.5}{4}\selectfont PSNR$\uparrow$}} & \multicolumn{1}{c|}{{\fontsize{6.5}{4}\selectfont LPIPS$\downarrow$}} & 
        \multicolumn{1}{c}{{\fontsize{6.5}{4}\selectfont PSNR$\uparrow$}} & \multicolumn{1}{c|}{{\fontsize{6.5}{4}\selectfont LPIPS$\downarrow$}} & 
        \multicolumn{1}{c}{{\fontsize{6.5}{4}\selectfont PSNR$\uparrow$}} & \multicolumn{1}{c|}{{\fontsize{6.5}{4}\selectfont LPIPS$\downarrow$}}
        & 
        \multicolumn{1}{c}{{\fontsize{6.5}{4}\selectfont PSNR$\uparrow$}} & \multicolumn{1}{c|}{{\fontsize{6.5}{4}\selectfont LPIPS$\downarrow$}}
        &
        \multicolumn{1}{c}{{\fontsize{6.5}{4}\selectfont PSNR$\uparrow$}} & \multicolumn{1}{c}{{\fontsize{6.5}{4}\selectfont LPIPS$\downarrow$}}
        
        \\ \midrule
        \multicolumn{1}{c|}{\begin{tabular}[c]{@{}c@{}} \textit{RM-NeRF}\end{tabular}} &  { 23.21}    &  \multicolumn{1}{c|}{ {0.093}} &  { 22.24}  & \multicolumn{1}{c|}{ { 0.12}} &  { 23.22}   & \multicolumn{1}{c|}{ {0.091}} &  {29.26}  &   \multicolumn{1}{c|}{ {0.041}} &  {30.34}  &   \multicolumn{1}{c|}{ {0.065}} &  {25.21}  &   \multicolumn{1}{c|}{{0.088}} & {20.2}  &  \multicolumn{1}{c|}{{0.14}}
        & {27.65}  &  \multicolumn{1}{c}{{0.11}} \\

        \multicolumn{1}{c|}{\begin{tabular}[c]{@{}c@{}}RM-NeRF {(w/o pose)}\end{tabular}} & \textbf{26.34} &  \multicolumn{1}{c|}{\textbf{0.048}} &  \textbf{26.02} &   \multicolumn{1}{c|}{\textbf{0.081}} & \textbf{25.23}  &  \multicolumn{1}{c|}{\textbf{0.051}}    &  \textbf{32.1}  &  \multicolumn{1}{c|}{\textbf{ 0.009}}  & \textbf{34.67}  &  \multicolumn{1}{c|}{\textbf{0.042}}  & \textbf{29.04}  &  \multicolumn{1}{c|}{\textbf{0.044}} & \textbf{23.2}  &  \multicolumn{1}{c|}{\textbf{0.072}} & 31.7  &  \multicolumn{1}{c}{\textbf{0.030}}         \\
        \bottomrule
    \end{tabular}
}
\vspace{1.00mm}
\caption{\Rone{\footnotesize 
\footnotesize \textbf{Increased Noise.} Comparison of our proposed method RM-NeRF (w/o pose) with the previously existing RM-NeRF on the Mulit-scale Blender dataset, but with increased noise as compared to table \ref{table:1}. Here, the basic RM-NeRF algorithm struggles to perform whereas the RM-NeRF has quite decent performance. }
}\label{table:noise}
\end{table*}

\begin{table*}[t]
\centering
%  \scriptsize
%\resizebox{\columnwidth}{!}
\resizebox{\textwidth}{!}
{\begin{tabular}{ccccccccccccccccc}
        %\hline
        \toprule
        \multicolumn{1}{c|}{}                                        & \multicolumn{2}{c|}{Truck}                         &  \multicolumn{2}{c|}{M60}     & \multicolumn{2}{c|}{Train}   & \multicolumn{2}{c|}{Family}
        & \multicolumn{2}{c|}{Ignatius}
        & \multicolumn{2}{c|}{Horse} & \multicolumn{2}{c|}{Museum}  &
        \multicolumn{2}{c}{Francis} \\
        
        \multicolumn{1}{c|}{}                                                                          & \multicolumn{1}{c}{{\fontsize{6.5}{4}\selectfont PSNR$\uparrow$}} & \multicolumn{1}{c|}{{\fontsize{6.5}{4}\selectfont LPIPS$\downarrow$}} &  \multicolumn{1}{c}{{\fontsize{6.5}{4}\selectfont PSNR$\uparrow$}} & \multicolumn{1}{c|}{{\fontsize{6.5}{4}\selectfont LPIPS$\downarrow$}} &                     \multicolumn{1}{c}{{\fontsize{6.5}{4}\selectfont PSNR$\uparrow$}} & \multicolumn{1}{c|}{{\fontsize{6.5}{4}\selectfont LPIPS$\downarrow$}} & \multicolumn{1}{c}{{\fontsize{6.5}{4}\selectfont PSNR$\uparrow$}} & \multicolumn{1}{c|}{{\fontsize{6.5}{4}\selectfont LPIPS$\downarrow$}} & 
        \multicolumn{1}{c}{{\fontsize{6.5}{4}\selectfont PSNR$\uparrow$}} & \multicolumn{1}{c|}{{\fontsize{6.5}{4}\selectfont LPIPS$\downarrow$}} & 
        \multicolumn{1}{c}{{\fontsize{6.5}{4}\selectfont PSNR$\uparrow$}} & \multicolumn{1}{c|}{{\fontsize{6.5}{4}\selectfont LPIPS$\downarrow$}}
        & 
        \multicolumn{1}{c}{{\fontsize{6.5}{4}\selectfont PSNR$\uparrow$}} & \multicolumn{1}{c|}{{\fontsize{6.5}{4}\selectfont LPIPS$\downarrow$}}
        &
        \multicolumn{1}{c}{{\fontsize{6.5}{4}\selectfont PSNR$\uparrow$}} & \multicolumn{1}{c}{{\fontsize{6.5}{4}\selectfont LPIPS$\downarrow$}}
        
        \\ \midrule
         \multicolumn{17}{c}{COLMAP poses and intrinsics} \\
        
        \midrule
        
        \multicolumn{1}{c|}{\begin{tabular}[c]{@{}c@{}}Mip-NeRF\end{tabular}} & 25.37 &  \multicolumn{1}{c|}{0.34} &  26.42 &   \multicolumn{1}{c|}{0.34} & 24.97  &  \multicolumn{1}{c|}{0.48}    &  25.98  &  \multicolumn{1}{c|}{ 0.42}  & 24.22  &  \multicolumn{1}{c|}{0.46}  & 27.45  &  \multicolumn{1}{c|}{0.26} & 26.71  &  \multicolumn{1}{c|}{0.43} & 29.57  &  \multicolumn{1}{c}{0.40}         \\
        \multicolumn{1}{c|}{\begin{tabular}[c]{@{}c@{}}Point-NeRF\end{tabular}} & 25.29 &  \multicolumn{1}{c|}{0.34} &  26.54 &   \multicolumn{1}{c|}{0.33} & 24.84  &  \multicolumn{1}{c|}{0.49}    &  25.76  &  \multicolumn{1}{c|}{ 0.44}  & 24.13  &  \multicolumn{1}{c|}{0.47}  & 26.88  &  \multicolumn{1}{c|}{0.29} & 26.77  &  \multicolumn{1}{c|}{0.43} & 28.88  &  \multicolumn{1}{c}{0.43}         \\
        
        \multicolumn{1}{c|}{\begin{tabular}[c]{@{}c@{}}RM-NeRF \end{tabular}} & \textbf{26.12} &  \multicolumn{1}{c|}{\textbf{0.32}} &  \textbf{27.60} &   \multicolumn{1}{c|}{\textbf{0.32}} & \textbf{25.78}  &  \multicolumn{1}{c|}{\textbf{0.42}}    &  \textbf{27.12}  &  \multicolumn{1}{c|}{ \textbf{0.29}}  & \textbf{25.29}  &  \multicolumn{1}{c|}{\textbf{0.37}}  & \textbf{27.89}  &  \multicolumn{1}{c|}{\textbf{0.25}} & \textbf{27.32}  &  \multicolumn{1}{c|}{\textbf{0.41}} & \textbf{29.89}  &  \multicolumn{1}{c}{\textbf{0.38}} \\

        \midrule
        \multicolumn{17}{c}{{COLMAP intrinsics, random poses}} \\
        \midrule

        \multicolumn{1}{c|}{\begin{tabular}[c|]{@{}c@{}}BARF\end{tabular}}          & 24.72   & \multicolumn{1}{c|}{0.53} & 24.89 &  \multicolumn{1}{c|}{0.54}  & 21.99  &  \multicolumn{1}{c|}{0.67}  & 22.87  &  \multicolumn{1}{c|}{0.61}  & 21.22  &  \multicolumn{1}{c|}{0.64}  & 23.88  &  \multicolumn{1}{c|}{0.48}  & 22.89  &  \multicolumn{1}{c|}{0.59}
        & 25.12  &  \multicolumn{1}{c}{0.65}
        \\     
        
        %\hline
        % \multicolumn{1}{c|}{\begin{tabular}[c]{@{}c@{}}Base B\end{tabular}} & 12.46 &  \multicolumn{1}{c|}{0.37} & 13.43 &   \multicolumn{1}{c|}{0.31} & 11.32  &  \multicolumn{1}{c|}{0.58}    &  14.26  &  \multicolumn{1}{c|}{ 0.29} & 13.71  &  \multicolumn{1}{c|}{0.42}  & 11.56  &  \multicolumn{1}{c|}{0.52}     & 12.22  &  \multicolumn{1}{c|}{0.47}& 15.87  &  \multicolumn{1}{c}{0.42}      \\ 
        \midrule
         % \multicolumn{1}{c|}{\begin{tabular}[c]{@{}c@{}}NeRF--\end{tabular}} & 16.89 &  \multicolumn{1}{c|}{0.094} &  19.89 &   \multicolumn{1}{c|}{0.118} & 15.67  &  \multicolumn{1}{c|}{0.074}    &  18.35  &  \multicolumn{1}{c|}{ 0.08}  & 20.22  &  \multicolumn{1}{c|}{0.098}  & 14.44  &  \multicolumn{1}{c|}{0.13} & 15.77  &  \multicolumn{1}{c|}{0.22} & 18.69  &  \multicolumn{1}{c}{0.20}         \\ %\hline
        
        \multicolumn{1}{c|}{\begin{tabular}[c]{@{}c@{}}NoPe-NeRF\end{tabular}} & 25.28 &  \multicolumn{1}{c|}{0.35} &  26.31 &   \multicolumn{1}{c|}{0.36} & 24.88  &  \multicolumn{1}{c|}{0.49}    &  25.93  &  \multicolumn{1}{c|}{ 0.44}  & 23.88  &  \multicolumn{1}{c|}{0.49}  & 27.28  &  \multicolumn{1}{c|}{0.28} & 26.68  &  \multicolumn{1}{c|}{0.43} & 29.28  &  \multicolumn{1}{c}{0.42}         \\

        \multicolumn{1}{c|}{\begin{tabular}[c]{@{}c@{}}RM-NeRF ({w/o pose})\end{tabular}} & \textbf{\textbf{25.81}} &  \multicolumn{1}{c|}{\textbf{0.33}} &  \textbf{26.71} &   \multicolumn{1}{c|}{\textbf{0.32}} & \textbf{24.88}  &  \multicolumn{1}{c|}{\textbf{0.49}}    &  \textbf{26.07}  &  \multicolumn{1}{c|}{\textbf{ 0.41}}  & \textbf{24.47}  &  \multicolumn{1}{c|}{\textbf{0.44}}  & \textbf{27.64}  &  \multicolumn{1}{c|}{\textbf{0.25}} & \textbf{26.89}  &  \multicolumn{1}{c|}{0.42} & \textbf{29.48}  &  \multicolumn{1}{c}{\textbf{0.40}}         \\
        \midrule
        \multicolumn{17}{c}{{w/o COLMAP poses and intrinsics}} \\
        \midrule 
        \multicolumn{1}{c|}{\begin{tabular}[c]{@{}c@{}}NeRF--\end{tabular}} & 24.98   & \multicolumn{1}{c|}{0.51} &  24.67 &  \multicolumn{1}{c|}{0.55} & 21.87  &  \multicolumn{1}{c|}{0.69}    & 22.98  & \multicolumn{1}{c|}{0.55}  & 21.03  &  \multicolumn{1}{c|}{0.67}  & 23.17 &  \multicolumn{1}{c|}{0.45} & 22.17  &  \multicolumn{1}{c|}{0.51}        & 25.62  &  \multicolumn{1}{c}{0.50}  \\ 
        
        \multicolumn{1}{c|}{\begin{tabular}[c]{@{}c@{}}RM-NeRF ({E2E})\end{tabular}} & \textbf{25.12} &  \multicolumn{1}{c|}{\textbf{0.36}} &  \textbf{25.88} &   \multicolumn{1}{c|}{\textbf{0.39}} & \textbf{24.07}  &  \multicolumn{1}{c|}{\textbf{0.52}}    &  \textbf{24.89}  &  \multicolumn{1}{c|}{ \textbf{0.47}}  & \textbf{23.28}  &  \multicolumn{1}{c|}{\textbf{0.57}}  & \textbf{26.02}  &  \multicolumn{1}{c|}{\textbf{0.34}} & \textbf{24.97}  &  \multicolumn{1}{c|}{\textbf{0.48}} & \textbf{27.97}  &  \multicolumn{1}{c}{\textbf{0.48}}         \\

        \midrule
        \multicolumn{17}{c}{{with same initialization as RM-NeRF}} \\
        \midrule 
        \multicolumn{1}{c|}{\begin{tabular}[c]{@{}c@{}}RM-NeRF(w/o pose) \end{tabular}} & \textbf{26.93} &  \multicolumn{1}{c|}{\textbf{0.28}} &  \textbf{28.12} &   \multicolumn{1}{c|}{\textbf{0.30}} & \textbf{26.46}  &  \multicolumn{1}{c|}{\textbf{0.34}}    &  \textbf{27.95}  &  \multicolumn{1}{c|}{ \textbf{0.24}}  & \textbf{26.34}  &  \multicolumn{1}{c|}{\textbf{0.28}}  & \textbf{28.34}  &  \multicolumn{1}{c|}{\textbf{0.20}} & \textbf{28.09}  &  \multicolumn{1}{c|}{\textbf{0.37}} & \textbf{30.43}  &  \multicolumn{1}{c}{\textbf{0.31}} \\
        
        \multicolumn{1}{c|}{\begin{tabular}[c]{@{}c@{}}RM-NeRF (E2E)\end{tabular}} & {26.57} &  \multicolumn{1}{c|}{{0.31}} &  {27.96} &   \multicolumn{1}{c|}{{0.31}} & {25.99}  &  \multicolumn{1}{c|}{{0.42}}    &  {27.89}  &  \multicolumn{1}{c|}{ {0.28}}  & {25.78}  &  \multicolumn{1}{c|}{{0.37}}  & {28.09}  &  \multicolumn{1}{c|}{{0.24}} & {27.78}  &  \multicolumn{1}{c|}{{0.39}} & {29.97}  &  \multicolumn{1}{c}{{0.37}} \\
        \bottomrule
       
    \end{tabular}
}
\vspace{1.0mm}
\caption{\footnotesize Comparison of our proposed methods with the baselines, on the Tanks and Temples dataset, under three different setups based on input required by the methods being evaluated. The table shows PSNR and LPIPS metric values for each of the methods on various scenes of the dataset.
% Performance comparison with other competing approaches. We used widely used PSNR and LPIPS performance metric to document the results.
Clearly, in all the three scenarios, our methods are able to surpass the baselines corresponding to that setup, thereby proving to be effective for real-world cases.
% Clearly, our method supersede the results of BARF \citep{lin2021barf}, Mip-NeRF \citep{barron2021mipnerf} and NeRF--   \citep{wang2021nerf} on Tanks and Temples dataset proposed by \cite{Knapitsch2017}. The results suggest that our approach can favorably deal with real-world scenes with a diverse set of camera trajectories. Also, it can be inferred from the above statistics that just relying on COLMAP poses for solving image based rendering on unconstrained sequence can demonstrably give inferior results. On the other hand, our joint optimization can handle outlier camera pose and provide favorable novel view rendering at the same time.
}\label{table:tankstemples_main}
\end{table*}

\noindent
\textbf{Results.} Table (\ref{table:1}) provides the quantitative comparison results on this dataset for RM-NeRF, RM-NeRF (w/o pose), and the baseline methods. We reported the results using the popular PSNR and LPIPS metrics averaged across all resolutions. For RM-NeRF (w/o pose), we initialized the orientation angles randomly and then converted them to rotation matrices.
%
% The reported results are average values of the metrics across all four resolutions.
% The row labelled with \textit{base} corresponds to the our RM-NeRF method and the row below it denote the RM-NeRF (w/o pose) method.
It can be observed that the baselines provides inferior view synthesis results for this setup.
The inferior results of baselines Base A, Base B, Base C as well Mip-NeRF, BARF and NeRF$-$$-$ shows that naively combining the multi-scale representation with existing pose refining methods is not an apt solution. Hence, showing the importance of our proposed RM-NeRF, RM-NeRF (w/o) appraoches. Note RM-NeRF (w/o pose) shows commendable results from randomly initialized rotations.
%as both of them show quite good metric values,  with RM-NeRF (w/o pose) reconstructing such a complex setup accurately from randomly initialized rotations.
% results show that our RM-NeRF method can jointly solve the multi-scale and pose error problems with NeRF and gives results significantly better than other baseline approaches in this setup.     
% Not only this, but our RM (w/o pose) method is also able to significantly outperform the baselines with a random initialization for camera rotations and intrinsics.

\begin{figure*}[!htb]
    \centering
    \includegraphics[width=\linewidth]{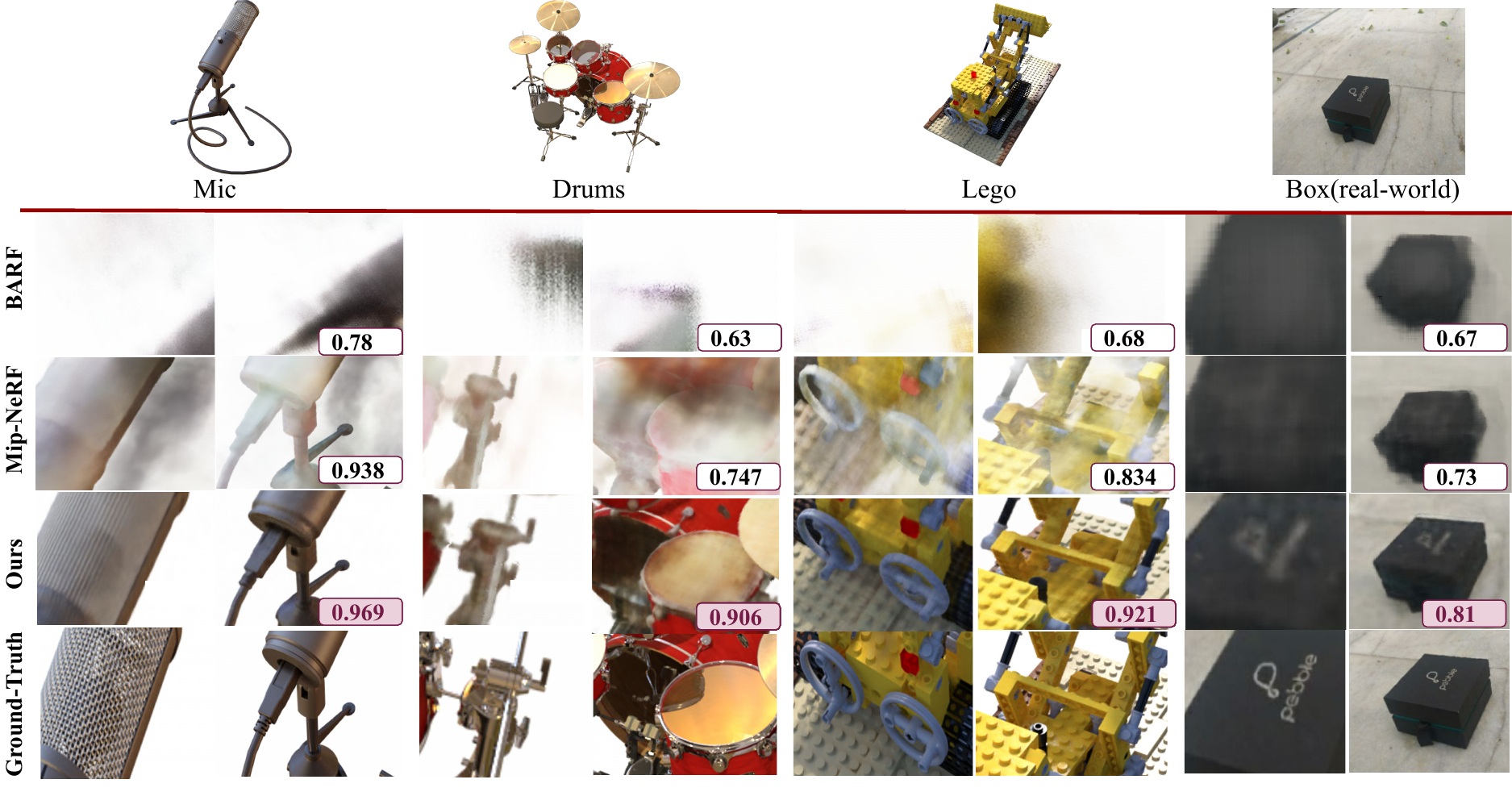}
    \caption{\footnotesize Qualitative Comparison of our method, and the baselines Mip-NeRF \citep{barron2021mipnerf}, BARF \citep{lin2021barf} on the multi-scaled version of the NeRF Blender dataset \citep{mildenhall2020nerf}, with noise added to camera poses, and our black-box sequence. We have used three scenes (Lego, Drums and Mic) from the Multi-Scale Blender dataset. Our method is able to localize and reconstruct the object much accurately as compared to the baselines for all the scenes including the black-box sequence.
    % We have visualized the synthesized images from all the approaches on 3 scenes namely Mic, Drums, Lego and our real sequence. Our method clearly provide better synthesized images. 
    %outperforms the competing approaches by rendering images comparable to the corresponding ground truths where other methods struggle to generate realistic images.
    %thus proving to be the suitable method for 3D reconstruction from real world datasets having multi-scale images and vulnerable to pose errors.
    %\kumar{Order: BARF, Mip-NERF, Ours, GT. Be creative.} 
    }
    \label{fig:vis}
\end{figure*}

Fig.(\ref{fig:vis}) provide the qualitative result comparison for the same. Both our approaches are able to achieve good quality renderings in this setup compared to the baselines.
%It can observed from the visual result that our approach robustly handle the multi-scale scene with pose error.

\noindent
\Rthree{\textbf{Inducing both rotational and translation errors.} Here, we analyze a more challenging setting where we introduce translation and rotation errors to the camera poses on the multi-scale blender dataset. A normal distribution with a standard deviation of 0.34  is used to sample the noisy translation, keeping the strategy for introducing rotational error the same as before. We keep our approach the same, i.e., use rotation averaging to refine the camera orientation solution and then use these refined solutions to estimate the translations. The results for this setup in Table \ref{table:2_rot_trans_blender} show that our method supersedes the other methods showing its effectiveness in a more realistic scenario.
}

\noindent
\Rone{\textbf{The need for  RM-NeRF (w/o pose).} We now analyze the effect of further increasing the perturbations to the Multi-Scale Blender dataset G.T. poses. We again sample the noise from a gaussian but this time the std is doubled. The results for RM-NeRF and RM-NeRF (w/o pose) are provided in table X for this setup. It can be observed that upon increasing noise, RM-NeRF might results in poor rendering and thus, for a very random scene trajectory, RM-NeRF (w/o pose) should be the chosen option. Also, we have added another row in tables \ref{table:1} and \ref{table:2_rot_trans_blender} where we use the initialization for RM-NeRF (w/o pose) to be same as RM-NeRF (noise added to G.T. poses). In this case, RM-NeRF (w/o pose) is able to outperform the RM-NeRF algorithm further proving its usefulness.}

\subsubsection{Tank and Temples}
\begin{figure}[!htb]
    \centering
    \includegraphics[width=.50\textwidth, height=0.50\textwidth]{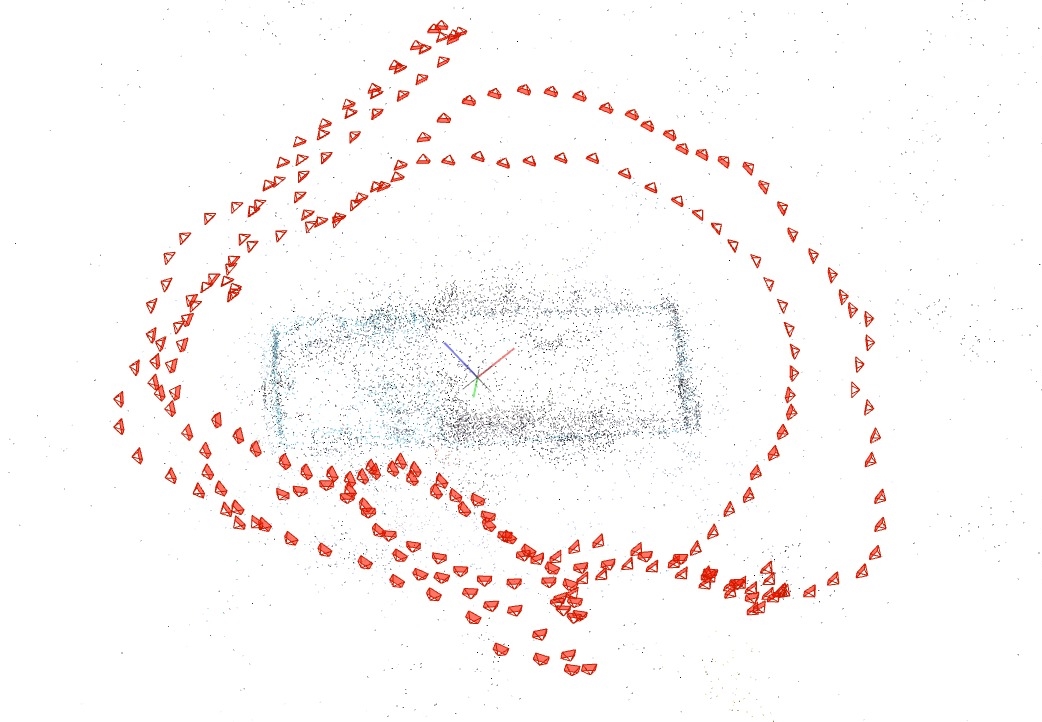}
    \caption{\footnotesize Camera poses recovered by COLMAP \citep{schoenberger2016sfm} corresponding to the Truck scene in Tanks and Temples dataset \citep{Knapitsch2017}, comprising a total of 251 images captured by gradually moving around the truck object. 
    % The camera pose trajectory recovered using COLMAP \citep{schoenberger2016sfm} on the truck sequence \citep{Knapitsch2017}. The video sequence consists of 251 poses taken freely with overall view of the truck.
    }
    \label{fig:truck_colmap}
\end{figure}

This dataset, proposed by \cite{Knapitsch2017}, comprises of challenging large scale real-world scenes. This dataset is widely used for bench-marking the 3D reconstruction algorithm. However, lately, this dataset has been popularly used in \Rone{evaluating} Novel View Synthesis methods designed to learn large scale scene representations with unconstrained camera pose trajectory.
%\sout{evalauting}
% It is widely used for evaluating 3D reconstruction methods  
% Tanks and Temples is a well-known challenging dataset containing real-world scenes, proposed by \cite{Knapitsch2017}. 
% It consists of images showing large scale scenes simulating realistic conditions and is largely used for evaluating 3D reconstruction methods. Further, this dataset can be very useful for testing unconstrained view-synthesis methods. 
Accordingly, we used some sequences of this dataset to evaluate our proposed methods against the popular baselines. Specifically, we have used 8 sequences from this dataset namely `{Truck}', `{M60}', `{Train}', `{Family}', `{Ignatius}', `{Horse}', `{Museum}' and `{Francis}', comprising both indoor and outdoor sequences with significant camera motion. For example, the {Truck} sequence consists of image set containing a 360\textdegree ~view of the subject captured freely at a varying distance from the object. Since there are no ground-truth poses available, COLMAP is used to estimate the initial poses and intrinsic camera matrix (see Fig.\ref{fig:truck_colmap} for the COLMAP result on {Truck} sequence). \textit{Note that our methods RM-NeRF (w/o pose) and RM-NeRF (E2E) are initialized with completely random camera poses for all the experiments performed on this dataset}.

\noindent
\textbf{Baselines.} We compared  view synthesis results of our approaches with the following baseline methods on this dataset: \emph{(a)} Mip-NeRF \cite{barron2021mipnerf}, Point-NeRF \citep{xu2022point} that uses COLMAP poses. \emph{(b)} NoPe-NeRF \citep{bian2022nope} that uses randomly initialized camera poses. \emph{(c)} NeRF $-$$-$ \citep{wang2021nerf} that do not use either intrinsic or extrinsic camera parameters.

% utilizing COLMAP poses and rendering good quality results (Mip-NeRF, Point-NeRF) on this dataset, the baselines learning correct scene representation for this dataset with randomly intialized poses (NoPe-NeRF \citep{bian2022nope}) and the baselines which can reconstruct scenes from this dataset without any requirement of intrinsics or camera poses (NeRF--).

\begin{figure*}[!htb]
    \centering
    \includegraphics[width=\linewidth]{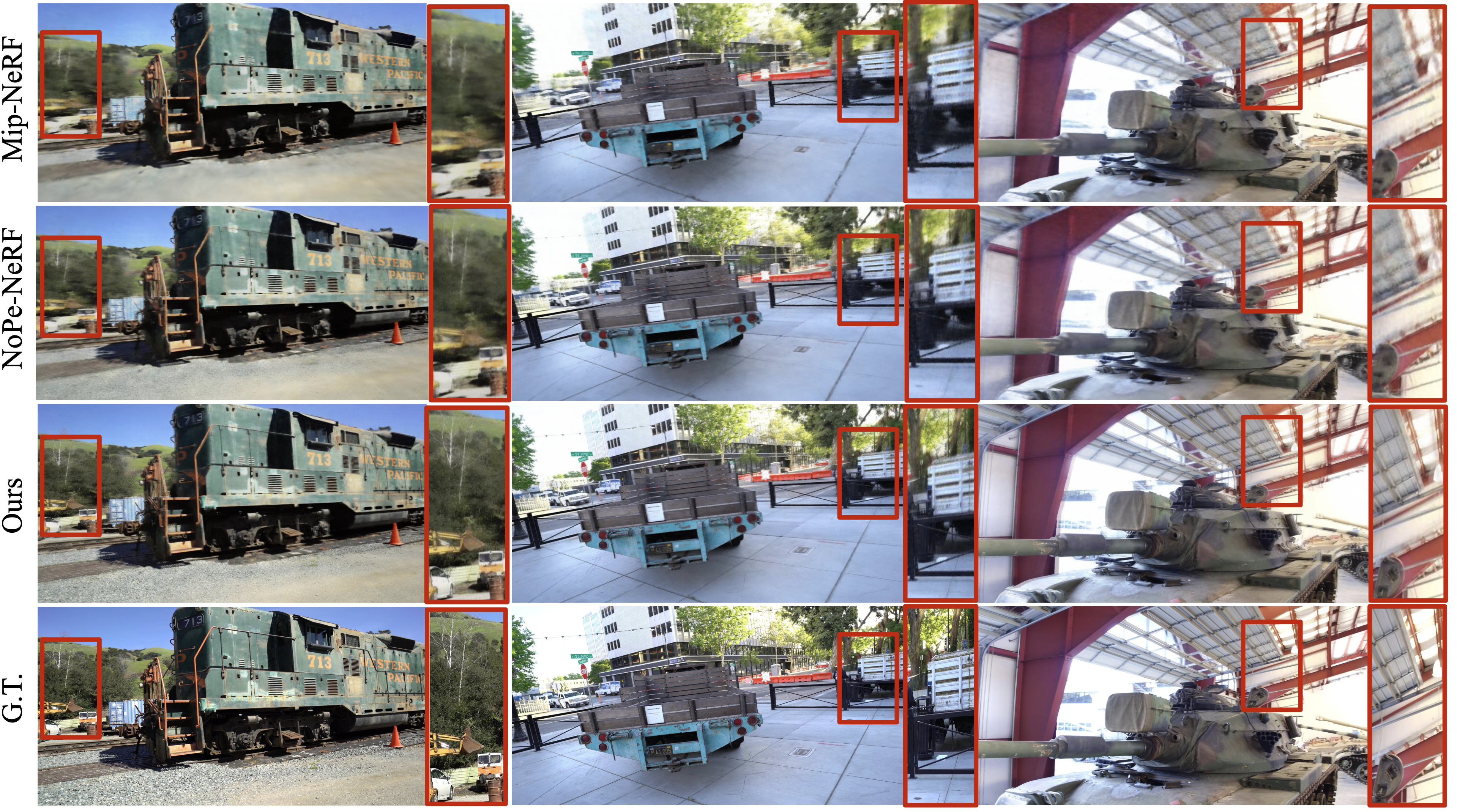}
    \caption{\footnotesize Qualitative Comparison of our RM-NeRF method, Mip-NeRF \citep{barron2021mipnerf} and NoPe-NeRF \citep{bian2022nope} on the Tanks and Temples dataset comprising real world scenes. We have visualized the synthesized images from all the approaches on 3 scenes namely Truck, M60 and Train. Our method clearly provide better synthesized images as emphasized by the red boxes. 
    %outperforms the competing approaches by rendering images comparable to the corresponding ground truths where other methods struggle to generate realistic images.
    %thus proving to be the suitable method for 3D reconstruction from real world datasets having multi-scale images and vulnerable to pose errors.
    %\kumar{Order: BARF, Mip-NERF, Ours, GT. Be creative.} 
    }
    \label{fig:vis_tanks}
\end{figure*}

\noindent
\textbf{Results.} Table (\ref{table:tankstemples_main}) shows the
quantitative comparison results using the PSNR and LPIPS metrics. It shows results under three different scenarios, the top set of results corresponds to the default scenario where COLMAP poses are used as input, the middle set of results corresponds to the scenario where poses are initialized randomly, and the bottom set of results corresponds to the scenario where both camera intrinsics and extrinsics are unknown.
For each setup, we use a different set of methods (baselines and our proposed method) for evaluation based on the input requirements.
\Rtwo{From the top setup, it can be observed that our RM-NeRF improves the camera pose accuracy once initialized using COLMAP camera pose results and provide improved image renderings when compared with the baselines}.  For the middle setup, it can be inferred that our approach RM-NeRF (w/o pose) surpasses the baselines, starting from random poses, thereby proving to be effective and robust in camera pose estimation than COLMAP and the Nope-NeRF \citep{bian2022nope} on this dataset. Finally, in the bottom setup, our method RM-NeRF (E2E) outperforms NeRF$-$$-$ and provides results comparable to our RM-NeRF (w/o pose) approach and baselines with random camera intrinsic parameters. We further show qualitative results for our RM-NeRF approach along with the Mip-NeRF (using COLMAP poses) and NoPe-NeRF (two-stage training \citep{bian2022nope}) in Figure \ref{fig:vis_tanks} for better insights. Clearly, our approach shows better image rendering results compared to the baselines.

%\Rone{\sout{other}}

The results obtained on this dataset demonstrate the potential of our approach in enhancing the view-synthesis framework to real-world scenes. Our joint optimization for modeling scene representation and camera pose estimation could provide favorable results for real-world scenes, including scenarios where no extra information other than the image set is given. Whereas relying on only COLMAP poses \citep{schoenberger2016sfm} with existing neural rendering approach may provide good results on a synthetic scene or a well-controlled setup. Yet, for a general real-world video sequence, such methods could lead to erroneous camera pose estimates leading to inferior view-synthesis results. Hence, our joint formulation provides robustness to the camera pose while giving better multi-scale image rendering.

% Thus, it can be concluded from the results on this dataset that our proposed methods have the potential to enhance the scope of multi-scale modelling to real-world scenes, based on a joint optimization of structure and camera, including scenario of no extra information other than the image set and the camera intrinsics or only the image set. 

% Also, relying on only COLMAP poses \citep{schoenberger2016sfm} may provide good results on synthetic scene or a well controlled sequence. But for a general real-world video sequence, it may provide erroneous camera poses thereby leading to inferior results. Hence, our joint formulation provide robustness to pose, at the same time giving better multi-scale image rendering.

%\Rone{\sout{the}}

\smallskip
\noindent
\Rone{\textbf{RM-NeRF (w/o pose, E2E) with COLMAP poses.} We use the Tanks and Temples dataset to conduct this study. We use COLMAP camera poses to initialize the  RM-NeRF (w/o poses) and RM-NeRF (E2E) methods and compare them against the RM-NeRF result. The results are provided in Table \ref{table:tankstemples_main}. It is easy to infer from the table that both RM-NeRF (E2E) and RM-NeRF (w/o pose) are able to outperform the base RM-NeRF method. This shows that the extension of RM-NeRF proposed in the article is better in a direct comparison setup, hence an encouraging take on the problem. The proposed extension could also work for a more realistic setting where the trajectory is sparse, and COLMAP might not be a very reliable pipeline, as we will also show in the later subsection with more realistic datasets.}

\subsubsection{ScanNet}
ScanNet is a widely used dataset to benchmark 3D reconstruction and semantic segmentation algorithm results for indoor scenes. Its train and validation sets contain 2.5M RGB-D images for 1512 scans acquired in 707 spaces. This dataset is collected using hardware-synchronized RGB and depth cameras of an iPad Air 2 at 30Hz exhibiting a realistic hardware setup. 

We studied the performance of our approaches on this dataset to observe view-synthesis result on room-scale indoor scene. Recently NoPe-NeRF \citep{bian2022nope} also performed such a study; therefore, we used their experimental setup for this study. This experiment involves subsampling the image sequences leading to 80-100 images per sequence. It is evaluated under two settings: \textbf{(I)} {with intrinsics}: where the intrinsic matrix is known, and we optimize for camera poses and \textbf{(II)} {w/o intrinsics}: where we optimize for both the camera intrinsic as well as extrinsic parameters. In setting \textbf{(I)}, we \Rone{compare} our RM-NeRF (w/o pose) method with the NoPe-NeRF baseline, which assumes given intrinsic parameters. For setting \textbf{(II)}, we compare our RM-NeRF (E2E) method with the NeRF$-$$-$ baseline, which optimizes both camera extrinsic and intrinsic parameters.

Table \ref{tab:scannet} shows both settings' experimental results using PSNR and LPIPS metrics. For each scenario, our proposed method outperforms the relevant baseline. This shows the effectiveness of our proposed joint optimization approaches in view-synthesis for indoor scene, thereby showcasing the benefit of motion averaging and multi-scale modeling.

%We further analyze the joint optimization capabilities of our proposed methods on four scenes from the ScanNet dataset \citep{dai2017scannet}, used in the NoPe-NeRF paper \citep{bian2022nope}. 

% We use the same experimental setup as the NoPe-NeRF, involving subsampling of the sequences, leading to 80-100 images per sequence and evaluate under two settings: (a) {with intrinsics}: where we use the known intrinsic matrices and optimize over the camera poses and (b) {w/o intrinsics}: where we optimize over all the camera parameters including both intrinsics and extrinsics. In setting (a), we \Rone{compare} our RM-NeRF (w/o pose) method with the NoPe-NeRF baseline which assumes given intrinsic parameters and for setting (b), we compare our RM-NeRF (E2E) method with the NeRF-- baseline which optimizes both camera poses and intrinsics.

%\sout{comapre}

%\noindent
% compare our methods (w/o pose and E2E) against it and the NeRF-- \cite{wang2021nerf} method which estimates the intrinsic matrix alongside dealing with randomly initialized poses.

\begin{table}[]
    \centering
    \scriptsize
    \begin{tabular}{c|cc|cc}
    \toprule
         & \multicolumn{2}{c|}{{with intrinsics}} & \multicolumn{2}{c}{{w/o intrinsics}} \\
        Scene & Nope-NeRF & Ours ({w/o pose}) & NeRF-- & Ours ({E2E}) \\
         \midrule
          0079\_00 & 32.47/0.41 &  \textbf{33.12/0.39} &30.59/0.49  & \textbf{31.88/0.47}\\
          0418\_00 & 31.33/0.34 &  \textbf{32.07/0.32}&30.00/0.40  & \textbf{31.23/0.46}\\
          0301\_00 & 30.83/0.36 & \textbf{30.83/0.35} & 27.84/0.45   & \textbf{29.14/0.42}\\
          0431\_00 & 33.83/0.39 & \textbf{34.09/0.38}  & 31.44/0.45  & \textbf{32.23/0.44}\\
          \bottomrule
    \end{tabular}
     \vspace{1.00mm}
    \caption{\small \textbf{ScanNet.} Performance comparison of our methods with NoPe-NeRF \citep{bian2022nope} and NeRF-- \cite{wang2021nerf} methods on 4 scenes of ScanNet dataset. The values in the table are of the format \Rtwo{PNSR/LPIPS}, respectively.}
    \label{tab:scannet}
\end{table}

\subsection{\Rone{NAVI Dataset}}
\Rone{NAVI is a recently proposed image collection dataset comprising scenes captured in the wild. It contains images of an object with various backgrounds and illumination conditions, which is an apt setting to test our proposed approaches. We evaluate the three proposed approaches on six complex scenes. Meanwhile, COLMAP performs poorly on 19 out of the 36 scenes from this dataset. For RM-NeRF, we use the COLMAP pose initialization whereas, for RM-NeRF (w/o pose) and RM-NeRF (E2E), camera parameters are initialized randomly. Table \ref{tab:navi} provides the results of our approaches on this dataset. It is easy to infer that RM-NeRF (w/o pose) can surpass RM-NeRF in this setup, demonstrating the advantage of our introduced extension in this article. Furthermore, the RM-NeRF (E2E) is better as compared to RM-NeRF in this case and is marginally inferior to the RM-NeRF (w/o pose), showing the possibility of a fully uncalibrated framework without sacrificing much on the rendering quality. Thus, RM-NeRF (E2E) method might be an excellent \textbf{self-contained framework} for view-synthesis in-the-wild or unstructured scenes instead of relying on third-party software such as COLMAP.}

%This can be attributed somewhat to COLMAP, which might have a somewhat higher error this time in the recovered poses and as a result RM-NeRF suffers since it is sensistive to pose initialization. 

% \Rone{We now analyze our newly proposed variations to the RM-NeRF on a recently proposed NAVI dataset \cite{jampani2023navi}, using its in-the-wild image-set collection. It comprises of images of an object invarious backgrounds and illumination conditions. The paper highlights that when using COLMAP, 19 out of the 36 scenes for this in-the-wild image set yields to poor reconstruction. We now evaluate our three methods on 6 most difficult out of these 19 scenes. For RM-NeRF, we again utilize the COLMAP initialization whereas for RM-NeRF (w/o pose) and E2E, poses are initialized randomly (intrinsics also randomly for E2E). We provide the results for this analysis in table \ref{tab:navi}. It can be observed that RM-NeRF (w/o pose) is able to surpass RM-NeRF in this setup. This can be attributed somewhat to COLMAP, which might have a somewhat higher error this time in the recovered poses and as a result RM-NeRF suffers since it is sensistive to pose initialization. The RM-NeRF E2E method is also slightly better as compared to RM-NeRF in this case and is worse than the RM-NeRF (w/o pose) by a small margin. Thus, this RM-NeRF (E2E) method might be a good framework for both intrinsics and extrinsic estimation for in-the-wild/unstructured image collections instead of relying on third party softwares like COLMAP.}

\subsection{Black-Box Dataset}
\subsubsection{Images taken from a freely moving camera}\label{subsec:ourdata}

\begin{figure*}[!htb]
    \centering
    \includegraphics[width=0.8\textwidth]{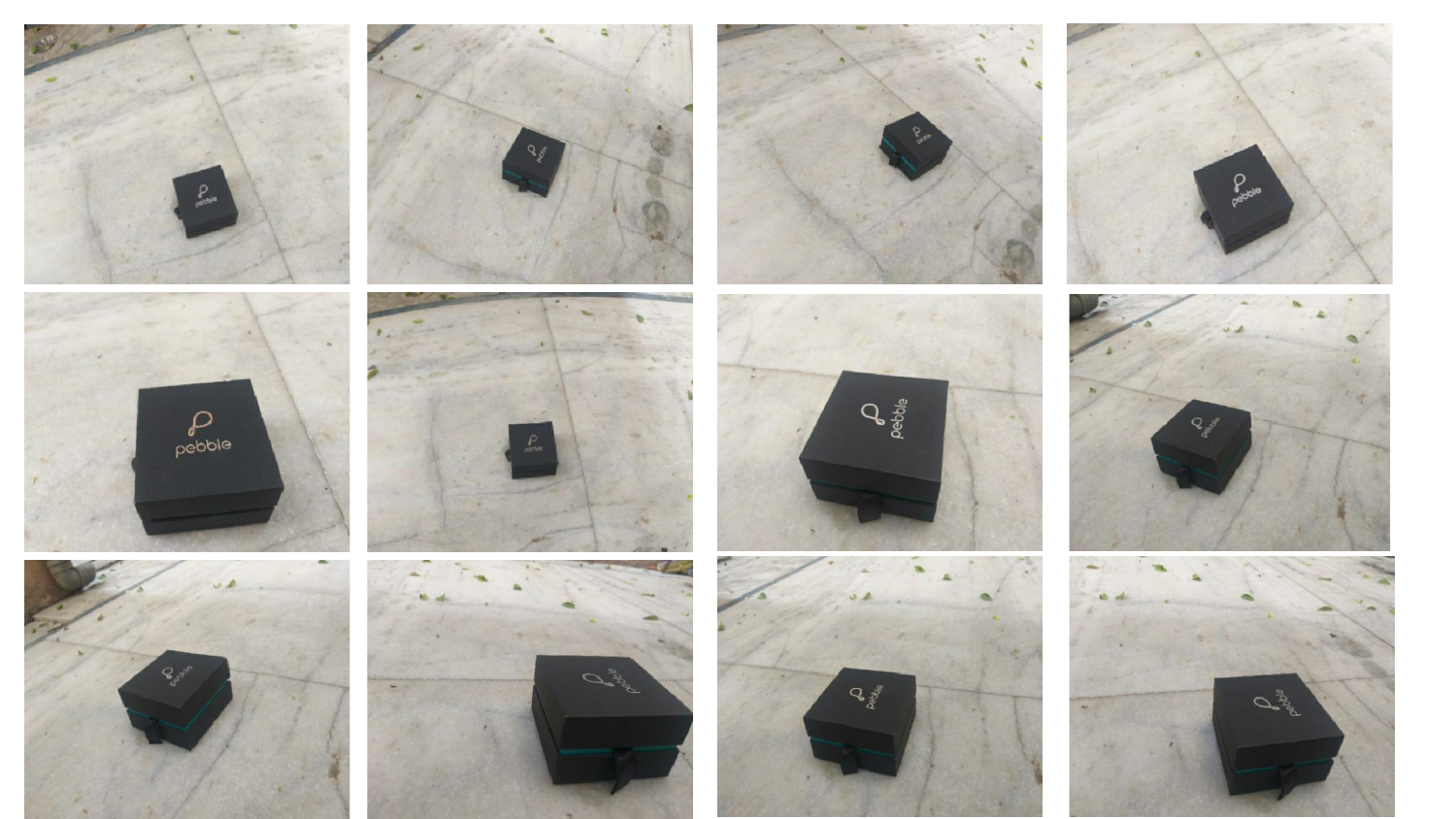}
    \vspace{1.00mm}
    \caption{\footnotesize  
    A subset of images corresponding to our black box dataset, captured using phone by randomly moving w.r.t. the box (Sec. \ref{subsec:ourdata}). 
    The aim behind this using scene is to mimic a very generic scenario representing day-to-day captured multi-view image sets.
    % Images corresponding to our black box dataset taken using freely moving camera in an unconstrained manner (discussed in Sec. \ref{subsec:ourdata}). We imitated a very general case of day-to-day captured multi-view images taken at different distance from the object with camera jittering effects as well.
    }
    \label{fig:custom_data}
\end{figure*}

\begin{figure}[!htb]
    \centering
    \includegraphics[width=0.77\linewidth]{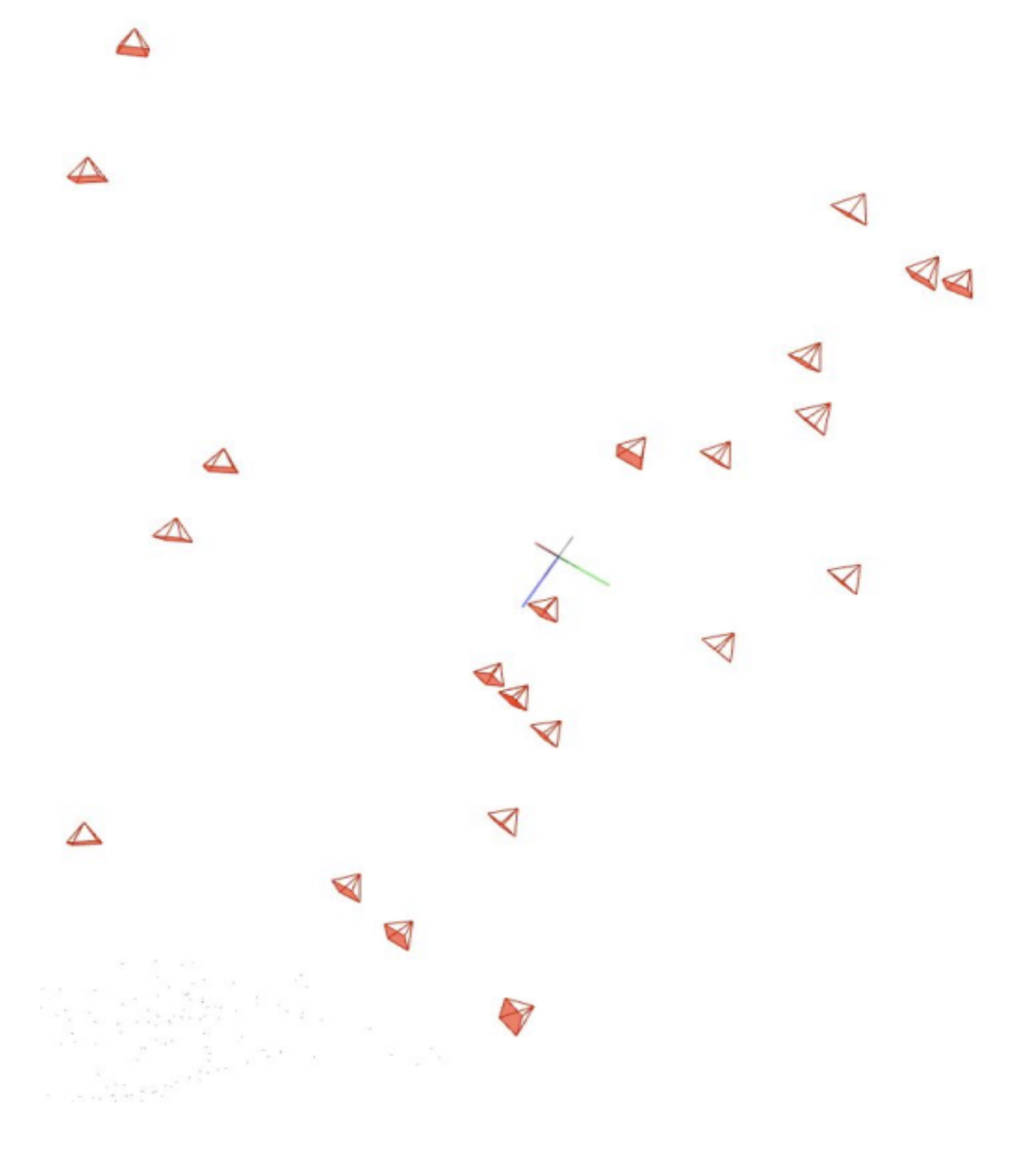}
    \vspace{1.00mm}
    \caption{\footnotesize Camera poses \Rthree{and a sparse collection of 3D points} corresponding to our black-box sequence (described in Sec. \ref{subsec:ourdata}) recovered using COLMAP \citep{schoenberger2016sfm}. 
    These estimated camera poses point verify the randomness in motion while capturing the back-box scene. 
    \Rthree{Also, the recovered set of 3D points (bottom-left) appear to be quite sparse further pointing to the failure of COLMAP in recovering the correct 3D and poses for this case.}
    % These camera poses are plotted to show the randomness in the camera motion estimated using COLMAP. Hence, using it blindly make the framework prone to  high pose errors.
    }
    \label{fig:camera_custom}
\end{figure}

To simulate a general real-world multi-view image capture setup for view-synthesis, we collect a dataset using a freely moving mobile phone. We captured 22 images of a simple black-box object (refer Fig.\ref{fig:custom_data}) using 16 out of them for training the model. The camera poses are shown in Figure \ref{fig:camera_custom} demonstrating that the images are captured at varying distance from the object.  The aim of this experiment is to show that for such real-world scene capture using COLMAP camera pose is not encouraging take for modeling view-synthesis problem. We must further refine the camera pose via mindful optimization. Accordingly, we compare our RM-NeRF approach result with Mip-NeRF by using COLMAP camera poses as initialization.

\begin{table}[]
  \scriptsize
  % \footnotesize
    \centering
    \resizebox{\columnwidth}{!}{
    \begin{tabular}{c|ccc|cc}
    \toprule
          Method & PSNR$\uparrow$ & LPIPS$\downarrow$ & SSIM$\downarrow$ & Rotation\textdegree$\downarrow$ & Translation$\downarrow$ \\
          \midrule
         RM-NeRF & 22.41 & 0.34 & 0.73& 26.54 & 24.91 \\
         RM-NeRF (w/o pose) & \textbf{23.12} & \textbf{0.29}  & \textbf{0.79} & \textbf{22.76} & \textbf{21.23} \\
         RM-NeRF (E2E) & 22.97 & 0.33  & 0.78 & 24.13 & 22.23  \\
         \bottomrule
    \end{tabular}
}
\caption{\small \textbf{Results on Navi Dataset.} Comparison of our proposed methods on navi dataset. For this, we have used 6 scenes from its in-the-wild image collection where COLMAP struggles to reconstruct accurate poses.}\label{tab:navi}
\end{table}

Figure \ref{fig:vis} shows the qualitative results for this scene in the last two columns alongside the Multi-Scale Blender dataset. For completeness, we included the results of the BARF method \citep{lin2021barf}. It is quite intuitive to assume that Mip-NeRF could have localized the object incorrectly, whereas our method can cast the apt cone in the scene space for object localization. And therefore, our method provides much better image rendering results. Additionally, it helps us deduce that assuming COLMAP poses as ground-truth poses can be misleading for real-world scenarios. This shows how a robust estimation approach on top of COLMAP camera poses initialization might effectively generalize the method for day-to-day captured multiview images.

% of the object, leading towards the conclusion that COLMAP poses might not be very accurate for this scene. This shows how a robust estimation scheme on top of COLMAP initialization might be an effective way to generalize the method for a randomly captured scene, instead of just assuming COLMAP as the ground-truth for real-world cases.

\begin{figure}
    \centering
    \includegraphics[width=0.5\textwidth]{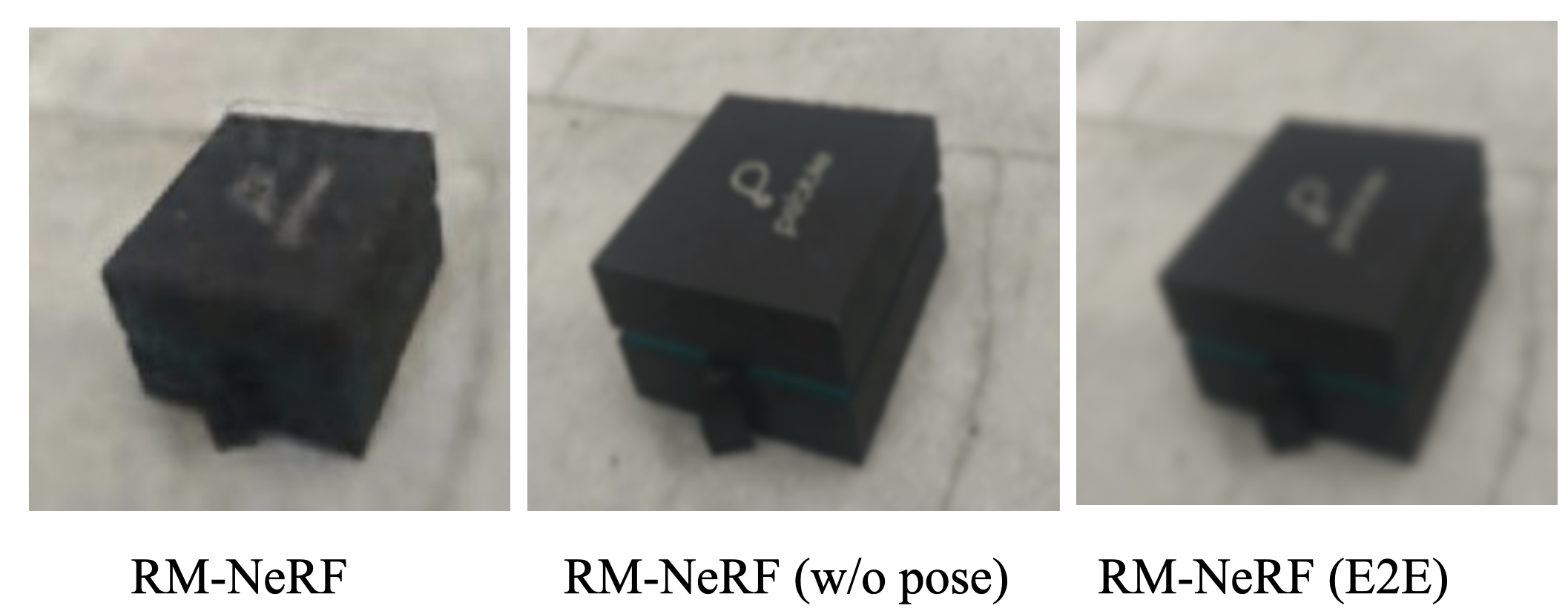}
    \caption{\small Results on the black box dataset for the proposed version. Here, RM-NeRF (w/o pose) provides more accurately rendering of the object, while RM-NeRF and RM-NeRF(E2E) approaches results are quite similar. Note that RM-NeRF (E2E) start from a randomly initialized pose without the knowledge of camera intrinsic parameters, hence demonstrating its suitability.}
    \label{fig:bb_results}
\end{figure}

\Rone{We further study the RM-NeRF (w/o pose) and RM-NeRF (E2E) performance on the black box dataset. The results for both these sequences are shown in Fig.\ref{fig:bb_results}. Clearly, the result shows the suitability of our introduced extension. The result demonstrates that it is quite possible to model the scene representation without access to COLMAP camera poses or camera intrinsic parameters without sacrificing much on the view-synthesis rendering quality.}
% \Rone{We further analyze our other variations namely RM-NeRF (w/o pose) and RM-NeRF (E2E) for this data. The results for both these sequences are present in Fig.\ref{fig:bb_results}. It can be observed that RM-NeRF (w/o) pose is almost similar (or slightly better) in terms of rendering to the RM-NeRF with poses initialized using COLMAP (Fig.\ref{fig:vis}). The RM-NeRF (E2E) version is somewhat poor in performance as compared to these both. Thus, for this scene, using COLMAP poses doesn't help the RM-NeRF much and instead our Rm-NeRF (w/o pose) version comes out to be the winner utilizing monocular priors and a well regularized pose optimization, leading us to a following conclusion: \textit{for randomly captured scenes, which depict the practical scenario, COLMAP might not be the best option even though it is quite frequently used as a ground Truth for popular real-world benchmarks. End-to-end optimization algorithms which are regularized using  geometric priors/constraints should serve the purpose for such cases.}}

\begin{figure}
    \centering
    \includegraphics[width=\linewidth]{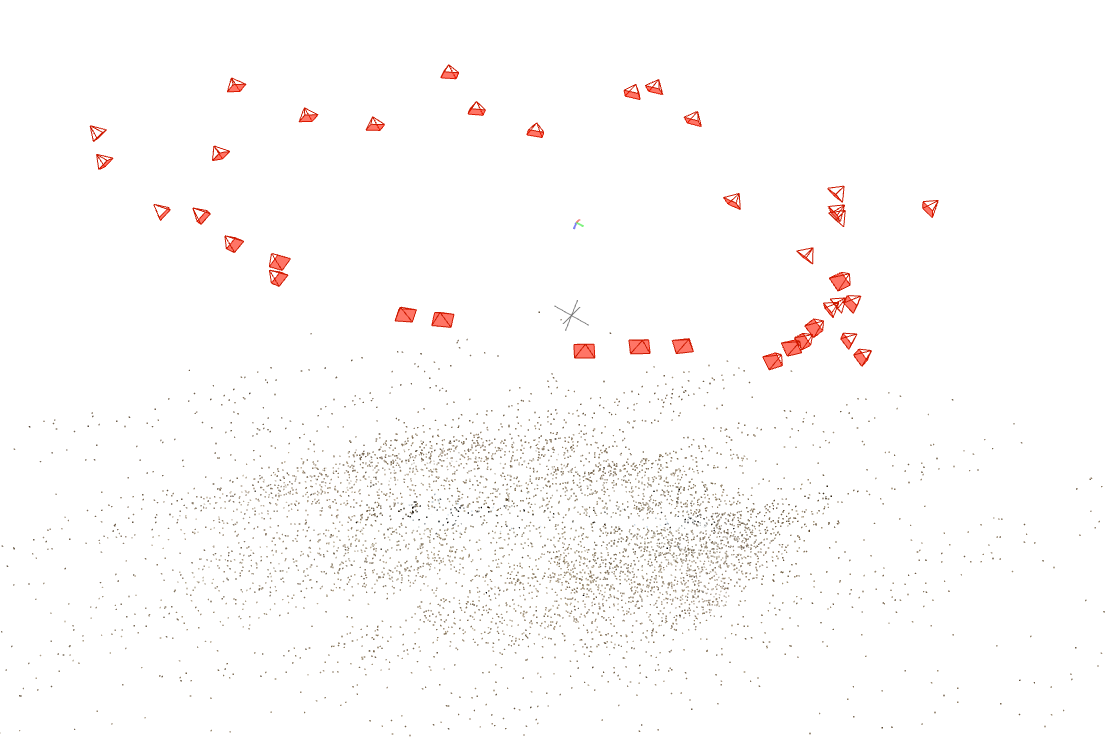}
    \caption{Approximate poses recovered by using COLMAP on the scene centered around a reflective fork object (Sec.\ref{sec:specular}), proposed in \cite{yen2022nerfsupervision}}
    \label{fig:fork_colmap}
\end{figure}

\subsubsection{Specular Objects}\label{sec:specular}
We studied the behavior of RM-NeRF (w/o pose) on objects with specular surfaces. As it is well-known that specular objects are often challenging to model for view synthesis, we test the limits of RM-NeRF (w/o pose) by conducting this study. Even recent works such as \cite{yen2022nerfsupervision} highlight this issue on their dataset comprising objects such as `spoon', `fork', etc. This dataset comprises eight object-centric scenes with  30-50 images of a reflective material object per scene. The scenes are captured using a slow-moving mobile phone around an object.  \cite{yen2022nerfsupervision}  estimates the camera poses and object's 3D using COLMAP. For testing, 8-10 images per scene are used.

\begin{figure}
    \centering
\includegraphics[width=0.85\linewidth]{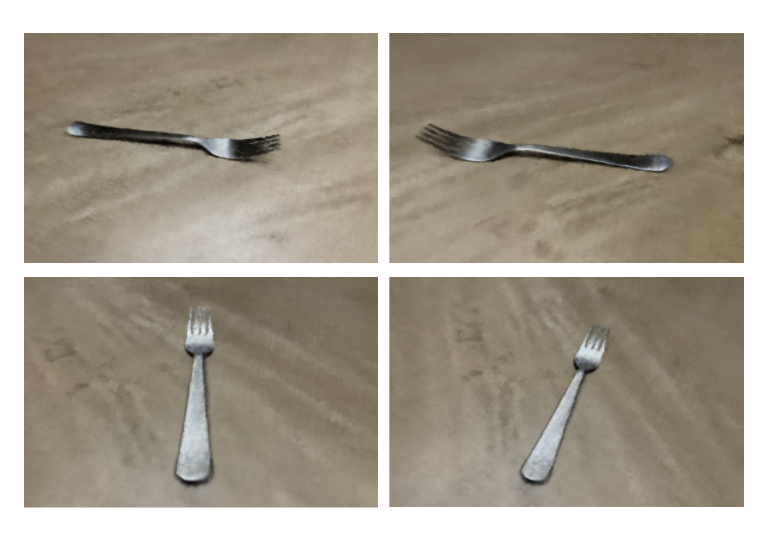}
\vspace{1.00mm}
    \caption{Qualitative evaluation of our RM-NeRF (w/o pose) method on a real-world scene (captured using phone) centered around a shiny fork object (Sec. \ref{sec:specular}), proposed in \cite{yen2022nerfsupervision}.}
    \label{fig:specular_result}
\end{figure}

For this experiment, we used the shiny fork object sequence comprising 38 training images. We followed the same setup as \cite{yen2022nerfsupervision} for evaluation. Figure \ref{fig:fork_colmap} shows the camera poses and sparse 3D points recovered for this scene. Figure \ref{fig:specular_result} shows our method's view synthesis result on this sequence. 
%We use the scene centered around a shiny fork object having 38 images for evaluating our method. 

Following the setup, we separate eight images for testing. Figure \ref{fig:fork_colmap} shows the camera poses and 3D points recovered via COLMAP on this scene, demonstrating the challenges in dealing with specular surfaces. Figure \ref{fig:specular_result} provides our method's view synthesis result on this sequence. \Rthree{Figure \ref{fig:specular_result} result shows 4 rendered images obtained using our method. The model hasn't seen the object from this viewpoint at train time. Despite favorable results in modeling view-synthesis for such an object, it is observed that our method has clear limitations in modeling it.}

\subsection{Ablations}\label{sec:ablations}
%\vspace{-1.0mm}
% We further analysed the performance of our approach under different input settings. For that, we first examine the performance behaviour of our method by introducing pose error and multi-scale error independently. Later in the section, we also provide a study on the optimization behaviour of Eq.\eqref{eq:weighted} w.r.t $\lambda$.

\subsubsection{Synthetic Datasets}
Here, we analyse our method's performance on the original Blender data \citep{mildenhall2020nerf} with camera pose error, Multi-Scaled Blender data \citep{barron2021mipnerf} with ground truth poses and a variation of our RM-NeRF method with unbiased weighting of the rotation averaging and rendering losses for updating camera poses. 
% We first begin by analyzing the gains provided by our method on the single-scale case in presence of pose errors. Then, we analze the performance of our method and the baselines on the Multi-scale Blender data with ground truth poses.

\noindent
\textbf{(a) Same Scale Images with Camera Pose Error.} 
We analyze our method's image-rendering results with the Mip-NeRF and BARF on the original NeRF Blender dataset in the presence of noisy rotations. Here, all images have the same resolution and are captured at a fixed distance from the object.
The results for this setup are shown in Table (\ref{table:3}), using the PSNR, LPIPS, and SSIM metrics for the four scenes. The statistical comparison show that our method supersedes the Mip-NeRF baseline. Also, it is comparable to the BARF method performance, which was designed to handle pose errors for single-scale datasets where all images are taken at approximately the same distance from the object.

\begin{table*}[!htb]
\scriptsize
\centering
\resizebox{\textwidth}{!}
{
\begin{tabular}{ccccccccccccc}
        \toprule
        \multicolumn{1}{c|}{}                                        & \multicolumn{3}{c|}{Lego}                         &  \multicolumn{3}{c|}{Ship}     & \multicolumn{3}{c|}{Drums}   & \multicolumn{3}{c}{Mic}                     \\ 
        \multicolumn{1}{c|}{}                                                                  & \multicolumn{1}{c}{{\fontsize{6.5}{4}\selectfont PSNR$\uparrow$}} & \multicolumn{1}{c}{{\fontsize{6.5}{4}\selectfont LPIPS$\downarrow$}} & \multicolumn{1}{c|}{{\fontsize{6.5}{4}\selectfont SSIM$\uparrow$}}                                                                                                                          & \multicolumn{1}{c}{{\fontsize{6.5}{4}\selectfont PSNR$\uparrow$}} & \multicolumn{1}{c}{{\fontsize{6.5}{4}\selectfont LPIPS$\downarrow$}} & \multicolumn{1}{c|}{{\fontsize{6.5}{4}\selectfont SSIM$\uparrow$}}                           & \multicolumn{1}{c}{{\fontsize{6.5}{4}\selectfont PSNR$\uparrow$}} & \multicolumn{1}{c}{{\fontsize{6.5}{4}\selectfont LPIPS$\downarrow$}} & \multicolumn{1}{c|}{{\fontsize{6.5}{4}\selectfont SSIM$\uparrow$}}    & \multicolumn{1}{c}{{\fontsize{6.5}{4}\selectfont PSNR$\uparrow$}} & \multicolumn{1}{c}{{\fontsize{6.5}{4}\selectfont LPIPS$\downarrow$}} & \multicolumn{1}{c}{{\fontsize{6.5}{4}\selectfont SSIM$\uparrow$}}      
        
        \\ \midrule
       
        \multicolumn{1}{c|}{\begin{tabular}[c]{@{}c@{}}Mip-NeRF\end{tabular}} & 17.90 & 0.089   & \multicolumn{1}{c|}{0.82} & 22.90 & 0.107 & \multicolumn{1}{c|}{0.71} & 14.07  & 0.110 & \multicolumn{1}{c|}{0.799}    & 21.90  & 0.064 & \multicolumn{1}{c}{0.930}    \\ %\hline
        \multicolumn{1}{c|}{\begin{tabular}[c|]{@{}c@{}}BARF\end{tabular}}          &   \textbf{27.61}  &  0.05 & \multicolumn{1}{c|}{\textbf{0.92}} &   \textbf{26.18} &  0.121 & \multicolumn{1}{c|}{\textbf{0.74}}  & 23.68  &  0.095 & \multicolumn{1}{c|}{ 0.880}  &  27.03  &  0.060 & \multicolumn{1}{c}{0.960}                   \\

        \multicolumn{1}{c|}{\begin{tabular}[c]{@{}c@{}}RM-NeRF(ours)\end{tabular}} &  27.10 &  \textbf{0.048}   & \multicolumn{1}{c|}{\textbf{0.92}} &  25.45 &  \textbf{0.069} & \multicolumn{1}{c|}{{0.73}} &   \textbf{24.98}  &  \textbf{0.072} & \multicolumn{1}{c|}{\textbf{0.907}} &   \textbf{30.03}  &  \textbf{0.027} & \multicolumn{1}{c}{\textbf{0.963}} \\ 
        \bottomrule
    \end{tabular}
    }
    \vspace{1.00mm}
\caption{\footnotesize  Performance comparison of the  proposed RM-NeRF method and Mip-NeRF, BARF baselines using 4 scenes from the NeRF Blender dataset \citep{mildenhall2020nerf}, with noise added to the poses. The evaluation is carried out using PSNR, LPIPS and SSIM metrics.}
\label{table:3}
\end{table*}

\noindent
\textbf{(b) Multi Scale Images with Ground-Truth Pose.}
We further study the multi-scale case on the multi-scale Blender dataset, but this time without perturbing the ground truth poses. Table (\ref{table:ms_gt}) compares the PSNR, LPIPS and SSIM values for this scenario. Once again, our method performance is similar to the current baselines, and the difference to the best method \citep{barron2021mipnerf} is minor, thereby showing the effectiveness of our approach. The point to note is, using our method, we don't have to rely on a separate module for estimating accurate pose and it is recovered jointly with the object's neural representation.

\emph{Looking at the rendered image quality results of both the tables, i.e., Table (\ref{table:3}) and Table (\ref{table:ms_gt}), our method performs well on both settings showing a clear advantage.}

\begin{table*}[!htb]
\centering
\scriptsize
\resizebox{\textwidth}{!}
{
    \begin{tabular}{ccccccccccccc}
        %\hline
        \toprule
        \multicolumn{1}{c|}{}                                        & \multicolumn{3}{c|}{Lego}                         &  \multicolumn{3}{c|}{Ship}     & \multicolumn{3}{c|}{Drums}   & \multicolumn{3}{c}{Mic}                      \\ 
        \multicolumn{1}{c|}{}                                                                  & \multicolumn{1}{c}{{\fontsize{6.5}{4}\selectfont PSNR$\uparrow$}} & \multicolumn{1}{c}{{\fontsize{6.5}{4}\selectfont LPIPS$\downarrow$}} & \multicolumn{1}{c|}{{\fontsize{6.5}{4}\selectfont SSIM$\uparrow$}}                                                                                                                          & \multicolumn{1}{c}{{\fontsize{6.5}{4}\selectfont PSNR$\uparrow$}} & \multicolumn{1}{c}{{\fontsize{6.5}{4}\selectfont LPIPS$\downarrow$}} & \multicolumn{1}{c|}{{\fontsize{6.5}{4}\selectfont SSIM$\uparrow$}}                           
        
        & \multicolumn{1}{c}{{\fontsize{6.5}{4}\selectfont PSNR$\uparrow$}} & \multicolumn{1}{c}{{\fontsize{6.5}{4}\selectfont LPIPS$\downarrow$}} & \multicolumn{1}{c|}{{\fontsize{6.5}{4}\selectfont SSIM$\uparrow$}}    & \multicolumn{1}{c}{{\fontsize{6.5}{4}\selectfont PSNR$\uparrow$}} & \multicolumn{1}{c}{{\fontsize{6.5}{4}\selectfont LPIPS$\downarrow$}} & \multicolumn{1}{c}{{\fontsize{6.5}{4}\selectfont SSIM$\uparrow$}}    
        
        \\ \midrule
        % \multicolumn{1}{c|}{\begin{tabular}[c|]{@{}c@{}}NeRF\end{tabular}}          & Passed  & Passed & \multicolumn{1}{c|}{100\%} & Passed & Passed & \multicolumn{1}{c|}{100\%}  & Passed  & Passed & \multicolumn{1}{c|}{100\%}   & Passed  & Passed & \multicolumn{1}{c}{100\%}                      \\ %\hline
        \multicolumn{1}{c|}{\begin{tabular}[c]{@{}c@{}}Mip-NeRF\end{tabular}} &  \textbf{29.34} &  \textbf{0.045}   & \multicolumn{1}{c|}{ \textbf{0.938}} &  \textbf{28.64} &  \textbf{0.0651} & \multicolumn{1}{c|}{\textbf{0.778}} &  \textbf{26.9}  &  \textbf{0.0452} & \multicolumn{1}{c|}{ 0.92}    &  \textbf{34.70}  &  0.0088 &  \multicolumn{1}{c}{\textbf{0.978}}    \\ %\hline
        \multicolumn{1}{c|}{\begin{tabular}[c|]{@{}c@{}}BARF\end{tabular}}          & 11.18  & 0.520 & \multicolumn{1}{c|}{0.700} & 9.230 & 0.760 & \multicolumn{1}{c|}{0.480}  & 11.2  & 0.680 & \multicolumn{1}{c|}{0.66}  & 12.18  & 0.530 & \multicolumn{1}{c}{0.740}                   \\    
        
        \multicolumn{1}{c|}{\begin{tabular}[c]{@{}c@{}}RM-NeRF (ours)\end{tabular}} & 29.30 &  \textbf{0.045}   & \multicolumn{1}{c|}{0.929} &  28.57 &  0.0653 & \multicolumn{1}{c|}{\textbf{0.778}} &  26.7  &  0.0455 & \multicolumn{1}{c|}{ \textbf{0.93}} &  34.30  &  \textbf{0.0082} & \multicolumn{1}{c}{0.969} \\ %
        %\multicolumn{1}{c|}{\begin{tabular}[c]{@{}c@{}}RM-NeRF+pose opt.\end{tabular}} & 25.6 & \cellcolor[HTML]{ffff80} 0.098 & \multicolumn{1}{c|}{0.81} & 23.18  & 0.13 & \multicolumn{1}{c|}{\cellcolor[HTML]{ffff80} 0.71} & 21.7  & 0.087 & \multicolumn{1}{c|}{0.83}  & 26.3  & 0.062 & \multicolumn{1}{c}{0.88} \\ %\hline
        %\hline
        \bottomrule
    \end{tabular}
}
\vspace{1.00mm}
\caption{\footnotesize Performance comparison of the  proposed RM-NeRF method and Mip-NeRF, BARF baselines using 4 scenes from the multi-scale Blender dataset \citep{barron2021mipnerf}, utilizing ground truth poses. The evaluation is carried out using PSNR, LPIPS and SSIM metrics.
% Comparison of the 3 metrics: PSNR, LPIPS and SSIM, resulting from the evaluation of our proposed RM-NeRF, BARF \citep{lin2021barf} and Mip-NeRF \citep{barron2021mipnerf} on 4 scenes of the multi-scale blender dataset with ground truth poses.
}\label{table:ms_gt}
\end{table*}

\noindent
\textbf{(c) Unbiased Optimization of Eq. \eqref{eq:weighted}}. 
%We further analyse a variant of our optimization method for the case without the warm starting distentangled optimization by fixing  $\lambda=0.5$ from the beginning.
We performed this study to provide a better insight into our weighted loss optimization strategy for the RM-NeRF method. For this, we initialized $\lambda=0.5$ in Eq.\eqref{eq:weighted} in the overall optimization. Table \ref{tab:poseopt} provides the results for this study using the PNSR, LIPIPS metric on the Multi-Scale blender dataset with noisy camera poses. The unbiased optimization variant result is denoted as Ours$^\dagger$ in Table \ref{tab:poseopt}. The results clearly show the benefit of biased optimization in Sec. \S \ref{sec:biased_opt}. It can be observed that this unbiased optimization results in inferior performance due to the complex optimizing landscape. Hence, the proposed biased optimization is suited for such loss function optimization. Initially, the bias is built toward estimating correct poses, which is gradually decreased with time. This leads to adequate minima after convergence.

\begin{table}
\scriptsize
% \footnotesize
\centering
\begin{tabular}{ c|cc|cc|cc|  }
\toprule
 \multicolumn{1}{c|}{} &\multicolumn{2}{c|}{Mean Error (\textdegree)}&\multicolumn{2}{c|}{Single-Scale Dataset}&\multicolumn{2}{c}{Multi-scale Dataset} \\
 \multicolumn{1}{c|}{}& \multicolumn{1}{c}{\fontsize{6.5}{4}\selectfont Ours\textsuperscript{$\dagger$} }&\multicolumn{1}{c|}{\fontsize{6.5}{4}\selectfont Ours}&\multicolumn{1}{c}{\fontsize{6.5}{4}\selectfont PSNR$\uparrow$}& 
 \multicolumn{1}{c|}{\fontsize{6.5}{4}\selectfont LPIPS$\downarrow$}& %\multicolumn{1}{c|}{\fontsize{6.5}{4}\selectfont SSIM$\uparrow$}& 
 \multicolumn{1}{c}{\fontsize{6.5}{4}\selectfont PSNR$\uparrow$}& \multicolumn{1}{c}{\fontsize{6.5}{4}\selectfont LPIPS$\downarrow$} %\multicolumn{1}{c|}{\fontsize{6.5}{4}\selectfont SSIM$\uparrow$}
  \\
 \midrule
 \multicolumn{1}{c|}{Lego}& \multicolumn{1}{c}{0.86}&\multicolumn{1}{c|}{\textbf{0.03}}&\multicolumn{1}{c}{24.0(\textbf{27.1})}&\multicolumn{1}{c|}{0.8(\textbf{0.5})}&\multicolumn{1}{c}{22.2(\textbf{27.0})}&\multicolumn{1}{c}{0.6(\textbf{0.4})} \\
 \multicolumn{1}{c|}{Ship}& \multicolumn{1}{c}{0.74}& \multicolumn{1}{c|}{\textbf{0.05}}&\multicolumn{1}{c}{23.7(\textbf{25.4})}&\multicolumn{1}{c|}{1.0(\textbf{0.7})}&\multicolumn{1}{c}{23.3(\textbf{26.5})}&\multicolumn{1}{c}{0.7(\textbf{0.7})} \\
 \multicolumn{1}{c|}{Drum}& \multicolumn{1}{c}{0.83}&\multicolumn{1}{c|}{\textbf{0.03}}&\multicolumn{1}{c}{20.2(\textbf{24.9})}&\multicolumn{1}{c|}{0.8(\textbf{0.7})}&\multicolumn{1}{c}{15.6(\textbf{26.0})}&\multicolumn{1}{c}{0.7(\textbf{0.4})} \\
 \multicolumn{1}{c|}{Mic}& \multicolumn{1}{c}{0.42}&\multicolumn{1}{c|}{\textbf{0.07}}&\multicolumn{1}{c}{24.2(\textbf{30.0})}&\multicolumn{1}{c|}{0.6(\textbf{0.3})}&\multicolumn{1}{c}{22.6(\textbf{30.0})}&\multicolumn{1}{c}{0.4(\textbf{0.1})} \\
  \midrule
  \multicolumn{1}{c|}{Chair}& \multicolumn{1}{c}{0.47}&\multicolumn{1}{c|}{\textbf{0.06}}&\multicolumn{1}{c}{27.4(\textbf{33.7})}&\multicolumn{1}{c|}{0.3(\textbf{0.5})}&\multicolumn{1}{c}{25.7(\textbf{35.2})}&\multicolumn{1}{c}{0.7(\textbf{0.3})} \\
   \multicolumn{1}{c|}{Ficus}& \multicolumn{1}{c}{0.78}&\multicolumn{1}{c|}{\textbf{0.07}}&\multicolumn{1}{c}{24.2(\textbf{28.6})}&\multicolumn{1}{c|}{0.7(\textbf{0.4})}&\multicolumn{1}{c}{22.7(\textbf{29.2})}&\multicolumn{1}{c}{0.7(\textbf{0.3})} \\
    \multicolumn{1}{c|}{Mat.}& \multicolumn{1}{c}{1.12}&\multicolumn{1}{c|}{\textbf{0.05}}&\multicolumn{1}{c}{20.2(\textbf{25.1})}&\multicolumn{1}{c|}{0.7(\textbf{0.7})}&\multicolumn{1}{c}{18.8(\textbf{24.8})}&\multicolumn{1}{c}{0.9(\textbf{0.6})} \\
     \multicolumn{1}{c|}{H.D.}& \multicolumn{1}{c}{0.62}&\multicolumn{1}{c|}{\textbf{0.05}}&\multicolumn{1}{c}{26.5(\textbf{33.1})}&\multicolumn{1}{c|}{0.6(\textbf{0.3})}&\multicolumn{1}{c}{23.3(\textbf{32.5})}&\multicolumn{1}{c}{0.8(\textbf{0.3})} \\
%  Hotdog & 1.8 & 0 & 4.4 & 0.03& 0.04 & 0.04\\
%  Materials & 1.8 & 0 & 4.4 & 0.03& 0.04 & 0.04\\
%  Lego & 1.8 & 0 & 4.4 & 0.03& 0.04 & 0.04\\
%  Lego & 1.8 & 0 & 4.4 & 0.03& 0.04 & 0.04\\
 \bottomrule
\end{tabular}
\vspace{1.00mm}
\caption{\footnotesize Performance analysis of our method with unbiased joint optimization ($\lambda=0.5$ in Eq.\eqref{eq:weighted}). This ablation is denoted as Ours\textsuperscript{$\dagger$}. We analyze both pose estimation and learned scene representations of this ablation. The first two columns compare the mean error in estimating poses using this ablation and our method which uses a biased optimization strategy in a single scale scale case. The remaining columns analyze the PSNR and LPIPS metric values for this ablation on the original and multi-scale versions of Blender dataset. \textbf{ For reference, we have provided the PSNR and LPIPS values for our RM-NeRF method, from Table \ref{table:1}, in the brackets.
% The bracket values for the PSNR and LPIPS denote our RM-NeRF method's result from Table \ref{table:1}.
}
% Performance of the joint optimization with $\lambda=0.5$ in Eq.\ref{eq:weighted} of the main paper. We recorded the pose estimation error and image rendering accuracy on both single scale and multi-scale Blender dataset. For pose-error mean angular error (in degrees) across all the rotations, are tabulated. As a reference to show the degradation of the performance, we have added the mean angular error in rotations using our approach for joint optimization, for the single scale case. The table also provides PSNR and LPIPS image rendering accuracy metric results for better understanding. \textbf{The bracket values for the PSNR and LPIPS denote our RM-NeRF method's result from Table \ref{table:1}.}
}
\label{tab:poseopt}
\end{table}

\subsubsection{Real-world Datasets}
Here, we study our approach under two circumstances often observed in any imaging data used for practical purposes. The first study corresponds to noisy monocular depth maps, and the second corresponds to the noise in input images. To study both cases, we use the Tanks and Temples dataset \citep{Knapitsch2017} comprising real-world images. We add some noise to the predicted monocular depth maps to simulate a general scene, which is usually valid for complex and cluttered scenes. Similarly, we also analyze the performance of our proposed methods in the presence of noise in the input images, accounting for the case with unclear or occluded regions in the day-to-day collected images.

%although the images have been selected with care

\smallskip
\noindent
\textbf{(a) Noisy 3D prior.}
Noise in the estimated monocular depths ultimately leads to noisy 3D prior for: (a) updating initial poses $\hat{\mathcal{P}}$ for our RM-NeRF (w/o pose), RM-NeRF (E2E) methods via point cloud alignment and (b) aligning depths to a global frame using rendered depth and parameters $\alpha$, $\beta$. To simulate this, we add noise $\delta \sim \mathcal{N}(\text{\textbf{0}}, 1e^{-1}\text{\textbf{I}})$ to a fraction of pixels in the normalized estimated depth maps, sampled uniformly throughout the image.

Table \ref{tab:3d_prior_noise} shows our method's results with noisy poses and the other relevant baselines in this setup on three sequences from the Tanks and Temples dataset, namely `M60', `Family', and `Ignatius'. For comparison, it shows results with and without (\textit{w/o}) this noise. We compared the results with the recent NoPe-NeRF \citep{bian2022nope} and Point-NeRF \citep{xu2022point} method. 
For Point-NeRF, we add the noise $\delta$ to the MVS estimated depth maps. All the methods' performance degrades with noise, yet our method performs better than others.

%We study the case when there is noise in estimated monocular depths, which ultimately leads to noisy 3D prior for: 
% (including ours) suffer in presence of this noise, showing significant drop in the PSNR values, although our method still performs the best. This shows the high sensitivity of these methods towards noise in the 3D prior, thereby enforcing a requirement of accurately estimated 3D structure. 
%both utilizing 3D prior either based on monocular or MVS based estimation
% Note, here no noise is added in the poses when estimating depth maps using MVS for Point-NeRF prior.
%this analysis (PSNR values) for our method (\textit{w/o poses})

\begin{table}[]
  % \scriptsize
  \footnotesize
    \centering
    \begin{tabular}{c|cc|cc|cc}
    \toprule
          & \multicolumn{2}{c|}{NoPe-NeRF} & \multicolumn{2}{c|}{Point-NeRF} & \multicolumn{2}{c}{Ours}  \\
         & \textit{noise} & \textit{clean} & \textit{noise} & \textit{clean} & \textit{noise} & \textit{clean} \\
         \midrule
         M60 & 24.47 & 26.31 & 24.23& 26.54 & \textbf{24.91} & \textbf{26.71}\\
         Family & 23.12 & 25.93  & 22.97 & 25.76 & \textbf{23.78} & \textbf{26.07}  \\
         Ignatius & 21.67 & 23.88  & 21.79& 24.13 & \textbf{22.23} & \textbf{24.47} \\
         \bottomrule
    \end{tabular}
    \vspace{1.00mm}
    \caption{\small Performance analysis of our RM-NeRF (w/o pose) method and the NoPe-NeRF \citep{bian2022nope}, PointNeRF \citep{xu2022point} baselines on scenes from Tanks and Temples dataset, when noise is added to the 3D prior (monocular depth maps in case of RM-NeRF (w/o pose), NoPe-NeRF and MVS depths in case of PointNeRF).}
    \label{tab:3d_prior_noise}
\end{table}

\noindent
\textbf{(b) Image quality and noise.} Day-to-day collected images can have noise due to low-quality imaging sensors or bad physical condition of the scene, which could effect the scene representation using images. We performed this study to observe the sensitivity of the proposed approaches to such noisy images. \Rthree{For this, we add holes to the given scene images, at randomly selected locations. Alongside this, we also add a small gaussian blur (std=0.1) to certain randomly sampled locations in the image. This is done to simulate the occlusion effect.}

%Day-to-day collected images can sometimes possess significant amounts of noise in form of occlusions or unclear portions, which can also affect the 3D reconstruction achieved using these images. We now analyze the effect of such noisy images on the reconstruction achieved by our proposed methods.
%We simulate this real-world noisy image scenario by adding synthetic noise to the input images and then feeding these corrupted images to our methods.

%\noindent
Table \ref{tab:image_noisy} shows the results for this experiment on three Tanks and Temples dataset namely `M60', `Family', and `Ignatius'. All the three proposed approaches have observable drop in their PSNR values, with E2E version having the maximum, when using these noisy images. It shows the  vulnerability of our approach to noisy images.

% This shows their vulnerability (although relatively different impact on each) to this scenario and this point towards an end-to-end differentiable scheme comprising a joint prior refinement along with correcting poses and learning the scene radiance field. Note, here only training images have noise induced in them.

% \noindent
% Table Y shows the results for this setup for our methods and baselines on the \textit{Truck}, \textit{Family} and \textit{Ignatius} sequence from the Tanks and Temples data along with the results for clean image setup. 

\begin{table}[]
    \scriptsize
    % \footnotesize
    \centering
    \begin{tabular}{c|cc|cc|cc}
    \toprule
          & \multicolumn{2}{c|}
          {\textit{RM-NeRF}} & \multicolumn{2}{c|}{RM-NeRF (w/o pose)} & \multicolumn{2}{c}{RM-NeRF (E2E)}  \\
         & \textit{noise} & \textit{clean} & \textit{noise} & \textit{clean} & \textit{noise} & \textit{clean} \\
         \midrule
         M60 & 26.91 & 27.60 & 26.12 & 26.71 &24.79 & 25.88\\
         Family & 26.41 & 27.12  & 25.43 & 26.07 & 23.93 &  24.89\\
         Ignatius & 24.71 & 25.29  & 23.71& 24.47 & 22.07 & 23.28 \\
         \bottomrule
    \end{tabular}
    \vspace{1.00mm}
    \caption{\small Performance analysis of our proposed methods on three scenes from the Tanks and Temples dataset \citep{Knapitsch2017}, with noise added to the input images.}
    \label{tab:image_noisy}
\end{table}

\subsection{Analyzing Estimated Camera Intrinsic Parameters}
Here, we provide the statistical results obtained using our RM-NeRF (E2E) approach in estimating intrinsic camera parameters. We compared our approach's result with popular NeRF$-$$-$ \citep{wang2021nerf} and recently proposed SC-NeRF \citep{jeong2021self}, which estimates both camera intrinsic and extrinsic parameters. For the experimental evaluation, we use three scenes from Tanks and Temples and ScanNet datasets. For Tanks and Temples, the authors have provided COLMAP parameters as the pseudo ground truth, which we used as it is for evaluation. Likewise, ScanNet's camera poses recovered using Bundle Fusion are treated as the pseudo ground truth.

Table \ref{tab:intrinsic} shows the difference in focal lengths, $\Delta f$ (in pixels) estimation by our approach compared to relevant baselines. Our approach performs better than the baselines, hence can be a good step towards differentiable intrinsic estimation leading to a complete end-to-end pipeline for joint estimation camera parameters--- intrinsic, extrinsic, and scene representation, using images.

\begin{table}[]
    \centering
    \begin{tabular}{c|ccc|ccc}
    \toprule
    & \multicolumn{3}{c|}{Tanks and Temples} & \multicolumn{3}{c}{ScanNet} \\
         & M60 & Family & Ignatius & 0079 & 0418 & 0301 \\
         \midrule
        SC-NeRF & 2.8 & 26.9 & 2.3 & 6.3 & 4.9 & 8.7\\
        NeRF$-$$-$ & 3.8 & 3.7 & 2.2 &  6.1 & 4.7 & 8.9\\
        Ours & \textbf{1.6} & \textbf{1.3} & \textbf{1.1} & \textbf{2.5} & \textbf{3.3} & \textbf{3.1} \\
        \bottomrule
    \end{tabular}
    \vspace{1.00mm}
    \caption{\small Performance comparison (\Rthree{in percentage error}) of our RM-NeRF (E2E) method and the baselines for focal length estimation on the Tanks and Temples dataset \citep{Knapitsch2017}. The table shows focal error $\Delta$f results of our method compared to NeRF$-$$-$ \citep{wang2021nerf}, and SC-NeRF \citep{jeong2021self}. The results are compiled \Rone{w.r.t. assumed ground-truths for these datasets, i.e., COLMAP results for Tanks and Temples and BundleFusion for ScanNet}
    }
    \label{tab:intrinsic}
\end{table}

% \subsubsection{Synthetic View Graphs} %Sec.\Sec(\textcolor{red}{4}) of 
% As discussed, our pose network is pre-trained by using supervision from 1200 synthetic viewgraphs proposed by \cite{purkait2020neurora}. This dataset consists of both outliers (fraction varying between 0 and 0.3) and noisy rotations where noise is sampled from a normal distribution with std. uniformly varying between 5\textdegree and 30\textdegree. We train our model on 80\% of the viewgraphs and remaining are used for testing.
% The mean and median angular error values resulting from evaluating our trained pose robustifying network on this dataset are $2.09^\circ$ and $1.1^\circ$ respectively. 
% % \smallskip
% \vspace{5.00pt}
% \begin{center}
% \begin{tabular}{ |c|c|c| } 
% \hline
%  no. of examples & mean error & median error \\ 
%  \hline
%  120 & 2.09\textdegree & 1.1\textdegree \\ 
%  \hline
% \end{tabular}
% \end{center}
% \vspace{5.00pt}
% Such statistics shows that our approach is effective in reducing the overall  camera motion error, which otherwise could directly affect the structure estimation leading to inferior 3D reconstruction result. Next, we study
% %
% %
% %To further make our case concrete, we 
% %
% the performance of our method on the pose estimation when tested on Blender Synthetic Dataset.

%in the subsequent section, thus providing further experimental reasoning to its robust performance in presence of noisy rotations, as presented in the paper.

\begin{table}
\footnotesize
\centering
\begin{tabular}{ c|cc|cc  }
 \toprule
 \multicolumn{1}{c|}{} &\multicolumn{2}{c|}{Noisy (\textdegree)}&\multicolumn{2}{c}{Improved (\textdegree)} \\
  \multicolumn{1}{c|}{}& \multicolumn{1}{c}{mean}&  \multicolumn{1}{c|}{rms}& 
 \multicolumn{1}{c}{mean}& \multicolumn{1}{c}{rms}
  \\
  \midrule
 \multicolumn{1}{c|}{Lego}& \multicolumn{1}{c}{1.78}& \multicolumn{1}{c|}{4.24}&\multicolumn{1}{c}{\textbf{0.031}}&\multicolumn{1}{c}{\textbf{0.041}} \\
 \multicolumn{1}{c|}{Ship}& \multicolumn{1}{c}{2.12} &\multicolumn{1}{c|}{4.02}&\multicolumn{1}{c}{\textbf{0.052}}&\multicolumn{1}{c}{\textbf{0.063}} \\
 \multicolumn{1}{c|}{Drums}& \multicolumn{1}{c}{1.98}& \multicolumn{1}{c|}{3.65}&\multicolumn{1}{c}{\textbf{0.038}}&\multicolumn{1}{c}{\textbf{0.052}} \\
 \multicolumn{1}{c|}{Mic}& \multicolumn{1}{c}{2.36}&\multicolumn{1}{c|}{5.12}&\multicolumn{1}{c}{\textbf{0.073}}&\multicolumn{1}{c}{\textbf{0.091}} \\
 \midrule
  \multicolumn{1}{c|}{Chair}& \multicolumn{1}{c}{1.76}&\multicolumn{1}{c|}{3.45}&\multicolumn{1}{c}{\textbf{0.056}}&\multicolumn{1}{c}{\textbf{0.071}} \\
   \multicolumn{1}{c|}{Ficus}& \multicolumn{1}{c}{2.46}&\multicolumn{1}{c|}{5.34}&\multicolumn{1}{c}{\textbf{0.065}}&\multicolumn{1}{c}{\textbf{0.096}} \\
    \multicolumn{1}{c|}{Materials}& \multicolumn{1}{c}{1.58}&\multicolumn{1}{c|}{3.55}&\multicolumn{1}{c}{\textbf{0.046}}&\multicolumn{1}{c}{\textbf{0.074}} \\
     \multicolumn{1}{c|}{Hotdog}& \multicolumn{1}{c}{2.23}&\multicolumn{1}{c|}{4.78}&\multicolumn{1}{c}{\textbf{0.052}}&\multicolumn{1}{c}{\textbf{0.071}} \\

%  Hotdog & 1.8 & 0 & 4.4 & 0.03& 0.04 & 0.04\\
%  Materials & 1.8 & 0 & 4.4 & 0.03& 0.04 & 0.04\\
%  Lego & 1.8 & 0 & 4.4 & 0.03& 0.04 & 0.04\\
%  Lego & 1.8 & 0 & 4.4 & 0.03& 0.04 & 0.04\\
\bottomrule
\end{tabular}
\vspace{2pt}
\caption{\small Our pose-refinement GNN results on noisy camera poses synthesized from Blender dataset \citep{mildenhall2020nerf}. The columns \Rone{corresponding} to noisy and improved denote the errors before and \Rone{after} applying our method, respectively.
%\sout{correpsonding} 
% The performance of our MRA pose-network pipeline on the view-graphs generated from the poses available in the Blender Synthetic Dataset \citep{mildenhall2020nerf}. For this experiment, we perturb the poses using the method described in the paper. The results show the performance for both the cases \textit{i.e.}, before and after applying our MRA pose network.  
% The results indicate that our method is able to reduce the errors in rotations by substantial amount and also provide a strong evidence regarding the improved performance of our method when compared with Mip-NeRF in presence of errors in estimating poses.
}
\label{tab:blender_view}
\end{table}

\subsection{Motion Averaging Analysis}\label{sec:exp_gnn}
A critical idea from multi-view geometry used in this article is motion averaging. In this section, we provide insights into the usefulness of motion averaging in providing robustness to the camera pose estimation pipeline.

\subsubsection{View graphs Analysis}

We conducted a simple test using the Blender dataset to show the benefits of view-graph modeling in motion averaging for robust camera pose estimation. For this, we perturbed one-fifth of the camera poses corresponding to each scene and constructed a view graph. We analyze camera pose errors before and after applying our pose-refining network to the view graph constructed using perturbed poses. Table (\ref{tab:blender_view}) result shows the robustness of our method in camera pose estimation. It can be observed that our pose-refining network handles noise efficiently and significantly reduces the overall camera pose error.

\subsubsection{Camera pose error analysis in presence of noisy feature key-point correspondence}

\begin{figure}
    \begin{center}
      \begin{subfigure}
  \centering
  \includegraphics[width=.23\textwidth, height=0.19\textwidth]{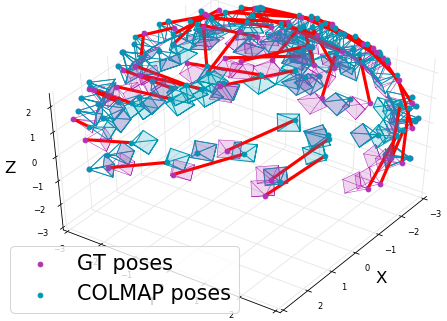}
  %\caption{A subfigure}
  \label{fig:sub1}
\end{subfigure}%
\begin{subfigure}
  \centering
  \includegraphics[width=.23\textwidth, height=0.19\textwidth]{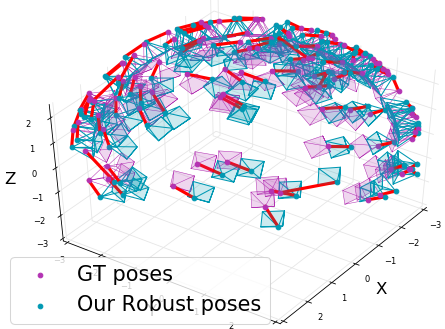}
  %\caption{A subfigure}
  \label{fig:sub2}
\end{subfigure}
\end{center}
\caption{\footnotesize Performance comparison of COLMAP and our pose-refining GNN w.r.t. ground truth poses on the Lego scene from the Blender dataset \citep{mildenhall2020nerf}, in presence of noise to the matched keypoint correspondences. The read line corresponds to the translation error in the estimated poses. 
\textbf{Left:} Camera pose output from COLMAP \citep{schoenberger2016sfm} along with the ground truth poses provided by the dataset. \textbf{Right:} Camera pose output from our poses refining GNN alongside the ground truth poses.
It can be observed that our estimated poses are much closer to the ground truth as compared to COLMAP.}
% (Red line shows pose error).
% Comparison of camera pose predicted by COLMAP  \citep{schoenberger2016sfm} and our methods on Lego dataset, when noise is added to the pairwise matched correspondences. 
% \textbf{Left:} Estimated camera pose from COLMAP \citep{schoenberger2016sfm} and the ground truth pose trajectory with the red line representing the pose difference. \textbf{Right:} Our recovered camera pose plotted against the ground-truth poses. It is easy to infer that our approach has better pose estimate.
% (Red line shows pose error).}
\label{fig:correspondences}
\end{figure}

Images captured in a real-world setting often contain noises leading to misleading keypoint matching between frames. This can affect the camera pose estimation results using the popular COLMAP framework. And therefore, to analyze this effect of noisy correspondences on our pose-refining method and COLMAP, we perform this simple experiment on the Lego scene from the Blender dataset. We first generate pairwise keypoint correspondences for images. Then, we add noise to these correspondences and estimate the relative pose between every image pair using these noisy correspondences. We then estimate absolute rotations from these noisy relative pose estimates using both approaches, i.e., COLMAP and our pose-refining GNN. Using these absolute camera rotation estimates, we compute corresponding camera translations.
Fig.(\ref{fig:correspondences}) shows the difference in predicted poses by each of these methods w.r.t. the ground truth poses provided by the dataset. Our approach offers robustness to such noise and attains significantly lesser error (consistent for all the images) compared to COLMAP in this scenario. The difference in the recovered camera pose is shown using lines. The results clearly show the effectiveness of our approach. Our approach gives better results than COLMAP, and the recovered camera pose is consistently better across images.

\Rone{The quantitative results for these experiments are provided in Table \ref{tab:noisy_corres}. For clarity, the rotation and translation error are provided separately.}

\begin{table}
\footnotesize
\centering
\begin{tabular}{ c|cc|cc  }
 \toprule
 \multicolumn{1}{c|}{} &\multicolumn{2}{c|}{Rot. error (\textdegree)}&\multicolumn{2}{c}{Trans. error (cm)} \\
  \multicolumn{1}{c|}{}& \multicolumn{1}{c}{COLMAP}&  \multicolumn{1}{c|}{Ours}& 
 \multicolumn{1}{c}{COLMAP}& \multicolumn{1}{c}{Ours}
  \\
  \midrule
 \multicolumn{1}{c|}{Lego}& \multicolumn{1}{c}{18.3\textdegree}& \multicolumn{1}{c|}{\textbf{12.4}\textdegree}&\multicolumn{1}{c}{{10.2}}&\multicolumn{1}{c}{\textbf{7.1}} \\
%  Hotdog & 1.8 & 0 & 4.4 & 0.03& 0.04 & 0.04\\
%  Materials & 1.8 & 0 & 4.4 & 0.03& 0.04 & 0.04\\
%  Lego & 1.8 & 0 & 4.4 & 0.03& 0.04 & 0.04\\
%  Lego & 1.8 & 0 & 4.4 & 0.03& 0.04 & 0.04\\
\bottomrule
\end{tabular}
\vspace{2pt}
\caption{\footnotesize Quantitative results for the noisy image keypoint correspondence case.
}
\label{tab:noisy_corres}
\end{table}

\noindent
\Rtwo{\textbf{Limitations.} Our proposed approaches could perform poorly on scenes containing specular and highly reflective surfaces. Moreover, further improvements could be made to our RM-NeRF (E2E) approach to apply to images collected from the internet, where each image of the same object is captured from a different camera.}
% As discussed in Sec.\ref{sec:specular} for specular surfaces, there is a significant scope of improvement for w/o pose and E2E variants which currently perform worse than the RM-NeRF initialized with COLMAP. Thus, it might be currently tricky to optimize starting from unknown poses for such shiny surfaces. Secondly, another limitation is to further modify the intrinsic estimation pipeline as the current E2E version performs worse than w/o pose method. Also, for random scenes (also include case of a sparse image set), the current approaches might struggle. }

\section{Conclusion and Future Directions}
In this paper, we introduce two extensions of our published work that allows the NeRF based scene representation for continuous view synthesis to work well for daily captured multi-view images. Specifically, the proposed approaches addresses the practical view synthesis issues around multi-scale images and the unavailability of camera parameters at train time. These issues are addressed using concepts from multi-view geometry, NeRF representation, and existing robust camera pose estimation literature. Although the proposed approaches may not be perfect, they open up the scope for modeling a randomly captured image set using continuous neural volumetric rendering without relying on third-party software such as COLMAP, hence self-contained framework. One interesting future direction is to extend our RM-NeRF (E2E) method to a scenario where different cameras are used to capture the multi-view images, leading to continuously varying intrinsic parameters. This can further broaden the scope of these NeRF-based methods to randomly collect images of a scene from the internet uploaded by different users. Few other interesting direction is to extend our RM-NeRF (w/o pose) and RM-NeRF (E2E) methods to scenes containing specular objects exhibiting interreflection \citep{kaya2021uncalibrated} or dynamic objects exhibiting non-rigid deformation \citep{kumar1885-164278, kumar2022organic}.

\smallskip
\noindent
\textbf{Dataset availability statement.}
The datasets used in this paper are publicly available. Their names and links are as follows: 

\begin{enumerate}
    \item \href{https://drive.google.com/file/d/18JxhpWD-4ZmuFKLzKlAw-w5PpzZxXOcG/view?usp=share\_link}{NeRF Blender Dataset}
    
    \item \href{https://github.com/google/mipnerf/blob/main/scripts/convert\_blender\_data.py}{Multi-Scale Blender Dataset}

    \item \href{https://www.tanksandtemples.org/download/}{Tanks and Temples Dataset} 

    \item \href{http://www.scan-net.org}{ScanNet Dataset}

    \item \href{https://drive.google.com/file/d/16\_y\_Nnh19Qhml0bg9RYR-hav0YOpWKuw/view?usp=share\_link}{Fork Scene}

    \item \textbf{Black Box example}: This is not public yet, but we will put online after the reviews. 
\end{enumerate}

\noindent
Other underlying data related to paper such as authors Orcid-ID and institution affiliation are publicly available online.

\bibliographystyle{spbasic}
\bibliography{review_file.bib}     % name your BibTeX data base

\appendix

\end{document}